\documentclass[10pt,twocolumn,letterpaper]{article}

\usepackage{cvpr}              

\usepackage[dvipsnames]{xcolor}

\definecolor{cvprblue}{rgb}{0.21,0.49,0.74}

\usepackage[pagebackref,breaklinks,colorlinks,citecolor=cvprblue]{hyperref}
\usepackage[utf8]{inputenc} 
\usepackage[T1]{fontenc}    
\usepackage{hyperref}       
\usepackage{url}            
\usepackage{booktabs, arydshln}       
\usepackage{amsfonts}       
\usepackage{nicefrac}       
\usepackage{microtype}      
\usepackage{amsmath}
\usepackage{graphicx}
\usepackage{amsthm}
\usepackage{subcaption}
\usepackage{pifont}
\newcommand{\cmark}{\ding{51}}%
\newcommand{\xmark}{\ding{55}}%
\newcommand{\blockcomment}[1]{}
\usepackage{multirow}
\usepackage{hhline}
\usepackage{paralist, tabularx}
\usepackage{floatrow}
\usepackage{color}
\usepackage{enumitem}
\definecolor{forestgreen}{rgb}{0.13, 0.55, 0.13}
\definecolor{bubblegum}{rgb}{0.99, 0.76, 0.8}
\definecolor{lightskyblue}{rgb}{0.53, 0.81, 0.98}
\definecolor{inchworm}{rgb}{0.7, 0.93, 0.36}
\newcommand{\imp}[1]{\textcolor{forestgreen}{\tiny{+#1}}}
\newcommand{\nimp}[1]{\textcolor{red}{\tiny{-#1}}}
\usepackage[dvipsnames]{xcolor,colortbl}
\usepackage{listings}

\usepackage{soul}

\newtheorem{theorem}{Theorem}
\newtheorem{lemma}{Lemma}

\newtheorem{definition}{Definition}

\makeatletter
\def\adl@drawiv#1#2#3{%
        \hskip.5\tabcolsep
        \xleaders#3{#2.5\@tempdimb #1{1}#2.5\@tempdimb}%
                #2\z@ plus1fil minus1fil\relax
        \hskip.5\tabcolsep}
\newcommand{\cdashlinelr}[1]{%
  \noalign{\vskip\aboverulesep
           \global\let\@dashdrawstore\adl@draw
           \global\let\adl@draw\adl@drawiv}
  \cdashline{#1}
  \noalign{\global\let\adl@draw\@dashdrawstore
           \vskip\belowrulesep}}
\makeatother

\usepackage[acronym,nowarn,section,nogroupskip,nonumberlist]{glossaries}

\glsdisablehyper{}
\newacronym{MoE}{MoE}{Mixture of Experts}
\newacronym{CC}{M\textsc{o}C\textsc{a}E}{Mixture of Calibrated Experts}
\newacronym{DE}{DE}{Deep Ensemble}
\newacronym{LR}{LR}{Linear Regression}
\newacronym{IR}{IR}{Isotonic Regression}
\newacronym{CA}{CA}{Class-agnostic}
\newacronym{CW}{CW}{Class-wise}
\newacronym{LaECE}{L\textsc{a}ECE}{Localisation-aware Expected Calibration Error}
\newacronym{LaACE}{L\textsc{a}ACE}{Localisation-aware Average Calibration Error}
\newacronym{LaMCE}{L\textsc{a}MCE}{Localisation-aware Maximum Calibration Error}

\newacronym{AP}{AP}{Average Precision}
\newacronym{AR}{AR}{Average Recall}
\newacronym{SOTA}{SOTA}{state-of-the-art}

\newacronym{ID}{ID}{in-distribution}
\newacronym{BA}{BA}{Balanced Accuracy}
\newacronym{OOD}{OOD}{out-of-distribution}
\newacronym{SAOD}{SAOD}{Self-aware Object Detection}
\newacronym{OVOD}{OVOD}{Open Vocabulary Object Detection}
\newacronym{SAODet}{SAOD\textsc{et}}{Self-aware Object Detector}

\newacronym{DAQ}{DAQ}{Detection Awareness Quality}
\newacronym{IDQ}{IDQ}{In-Distribution Quality}
\newacronym{ECE}{ECE}{Expected Calibration Error}
\newacronym{AUC}{AUC}{area-under-curve}
\newacronym{TN}{TN}{true-negative}
\newacronym{TP}{TP}{true-positive}
\newacronym{FP}{FP}{false-positive}
\newacronym{FN}{FN}{false-negative}
\newacronym{IoU}{IoU}{Intersection-over-Union}
\newacronym{LRP}{LRP}{Localisation-Recall-Precision Error}
\newacronym{AV}{AV}{Autonomous Vehicles}
\newacronym{AUROC}{AUROC}{Area-under ROC Curve}
\newacronym{NMS}{NMS}{Non-Maximum Suppression}
\newfloatcommand{capbtabbox}{table}[][\FBwidth]
\usepackage{titlecaps}

\usepackage{blindtext}
\usepackage[capitalize]{cleveref}
\crefname{section}{Sec.}{Secs.}
\Crefname{section}{Section}{Sections}
\Crefname{table}{Table}{Tables}
\crefname{table}{Tab.}{Tabs.}

\def\paperID{5059} 
\def\confName{CVPR}
\def\confYear{2024}
 
\title{MoCaE: Mixture of Calibrated Experts Significantly Improves Object Detection}

\author{Kemal Oksuz, \; Selim Kuzucu, \; Tom Joy, \; Puneet K. Dokania \\
Five AI Ltd., United Kingdom\\
{\tt\small \{kemal.oksuz, selim.kuzucu2, tom.joy, puneet.dokania\}@five.ai}
}

\begin{document}
\maketitle
\begin{abstract}
    Combining the strengths of many existing predictors to obtain a Mixture of Experts which is superior to its individual components is an effective way to improve the performance without having to develop new architectures or train a model from scratch. 
    %
    %
    However, surprisingly, we find that na\"ively combining expert object detectors in a similar way to Deep Ensembles, can often lead to degraded performance.
    We identify that the primary cause of this issue is that the predictions of the experts do not match their performance, a term referred to as miscalibration.
    Consequently, the most confident detector dominates the final predictions, preventing the mixture from leveraging all the predictions from the experts appropriately.
    To address this, when constructing the Mixture of Experts, we propose to combine their predictions in a manner which reflects the individual performance of the experts; an objective we achieve by first calibrating the predictions before filtering and refining them.
    We term this approach the Mixture of Calibrated Experts and demonstrate its effectiveness through extensive experiments on 5 different detection tasks using a variety of detectors, showing that it: (i) improves object detectors on COCO and instance segmentation methods on LVIS by up to $\sim 2.5$ AP; (ii) reaches state-of-the-art on COCO test-dev with $65.1$ AP and on DOTA with $82.62$ $\mathrm{AP_{50}}$; (iii) outperforms single models consistently on recent detection tasks such as Open Vocabulary Object Detection.
\end{abstract}

\blockcomment{
\begin{abstract}
We propose an extremely simple and highly effective approach to faithfully combine different object detectors to obtain a \gls{MoE} that has a superior accuracy to the individual experts in the mixture. 
We find that na\"ively combining these experts in a similar way to the well-known \glspl{DE}, does not result in an effective \gls{MoE}. 
We identify the incompatibility between the confidence score distribution of different detectors to be the primary reason for such failure cases.
Therefore, to construct the \gls{MoE}, our proposal is to first calibrate each individual detector against a target calibration function. Then, filter and refine all the predictions from different detectors in the mixture. %
We term this approach as \gls{CC} and demonstrate its effectiveness through extensive experiments on 5 different detection tasks using 15 different object detectors.
%
For example, \gls{CC} (i) improves object detectors on COCO and instance segmentation methods on LVIS up to $\sim 2.5$ AP; (ii) reaches \gls{SOTA} on COCO \textit{test-dev} with $65.1$ AP and on DOTA for rotated object detection with $82.62$ $\mathrm{AP_{50}}$; (iii) outperforms single models consistently on recent detection tasks such as \gls{OVOD} and \gls{SAOD}.
%
Code will be made public.
\end{abstract}
}

\blockcomment{
\paragraph{TODO Theory/Narrative}

\begin{enumerate}
    \item Think about calibration objective, is it appropriate to set precision = 1? 
    \item Link to posterior predictive and BNNs, we are now sampling over the function space and not the parameter space. Would be really nice to make an argument about bias, uncertainty etc...
    \item Mechanistically can link this to diversity of the experts, i.e. architectures or predictive distributions? Is it even appropriate to mix different predictive distributions? 
    \item re-write sections of the paper
    \item Paper is quite verbose currently, should try to improve the clarity
\end{enumerate}

\paragraph{TODO Experiments}
\begin{enumerate}
    \item \st{Add robustness experiment results as per SAOD task, aka follow Table 6 in SAOD paper. High priroty.} This is in~\cref{tab:evaluation}.
    \begin{itemize}
        \item Why is MoCaE worse for OOD?
        \item \st{Is calibration of experts the issue? Test OOD detection on each MoCaE expert. This is done on each expert}
        \item \st{Test on MoCaE, check OOD when aggregating, is it NMS+?}
        \item \st{Check the final calibration, possible that we have confounding of predicted image level uncertainties. This is done on MoCaE with calibration}
        \item \st{Keep boxes, use raw scores for OOD}
        \item \st{Is it how we are aggregating the detection level? Play for top k}
        \item \st{Reliability diagrams for before/after MoCaE-level calibration for single models}
        \item Analyze the False Positives of MoCaE: How much of them is due to the mistakes of one and only one single model?
        \item Create pie charts at OOD + corruptions
        \item Analyse the calibration function, is it still relevant under domain shift. Make sure MoCaE calibration on each expert is similar to each detector in SAOD
        \item Do we use uncalibrated probabilites or not? A B test     
        \item 
    \end{itemize}
    \item histogram pre and post calibration
    \item \st{MoCaE with different corruption levels (AP), as is standard in practice. High priority.}~\cref{tab:corr} still need to add \gls{LaECE}
    \item Add FLOPs + param size. High priority, results are in rebuttal. High priority.
    \item Diversity of predictions. Create histogram of mAP for each class (think about clustering of classes on x axis). One graph small/medium/large objects etc
    \item A couple of figures and/or ablations to show that if the performance gap between the detectors is sufficiently high, then MoCaE won't improve much over the better performing one. I.e. what if all of the TPs are a subset of the TPs of a better detector. I.e. explore the limits of calibration. Medium priority.
    \item Fine grained analysis of detections which then become TPs in MoCaE, are there any which are only TPs in MoCaE? Or do you have to ensure that there is a TP in one of the experts, can it be greater than the sum of it's parts? Medium priority
    \item Additional experiments to show what would happen if more than one (e.g 3) of the same detector is present in the mixture. Can link this to theory. Low priority
    \item Ensure that for the same compute budget, MoCaE is superior to other models. Medium priority
    \item Can we use other models? I.e. let's not restrict ourselves to ATSS, PAA, and RS R-CNN. Can link this to experiment on TPs being a subset of other detectors, are some detectors redundant? Low/Medium priority
    \item Training time calibration experiments
\end{enumerate}
}

\begin{figure}[t!]
    \centering
              \begin{subfigure}[b]{0.4\textwidth}
                \includegraphics[width=\textwidth]{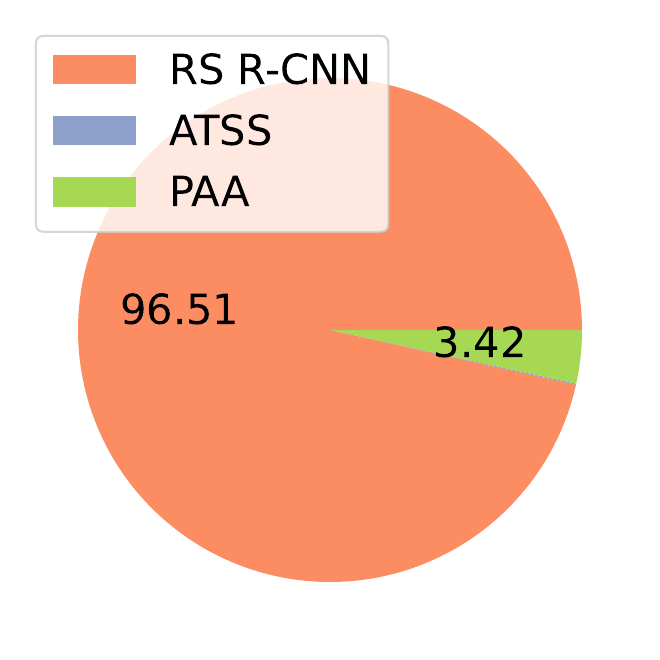}
                \caption{Mixture of \\uncalibrated Experts}
            \end{subfigure}
            \begin{subfigure}[b]{0.4\textwidth}
                \includegraphics[width=\textwidth]{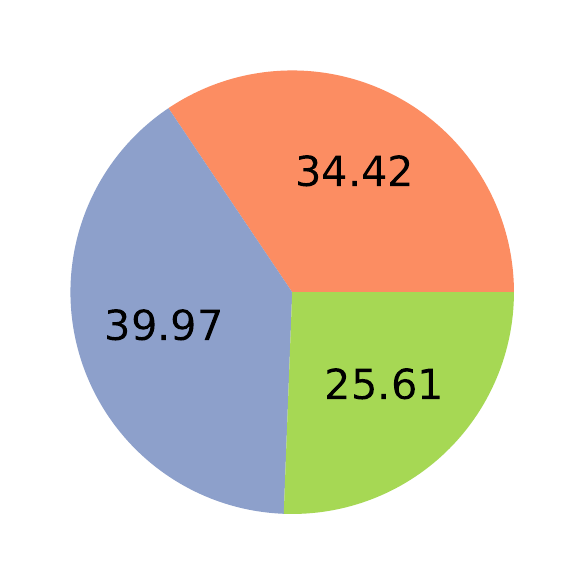}
                \caption{Mixture of \\calibrated Experts}
            \end{subfigure}
            \vspace{-2.ex}
        \begin{subfigure}[b]{0.70\textwidth}
             \includegraphics[width=\textwidth]{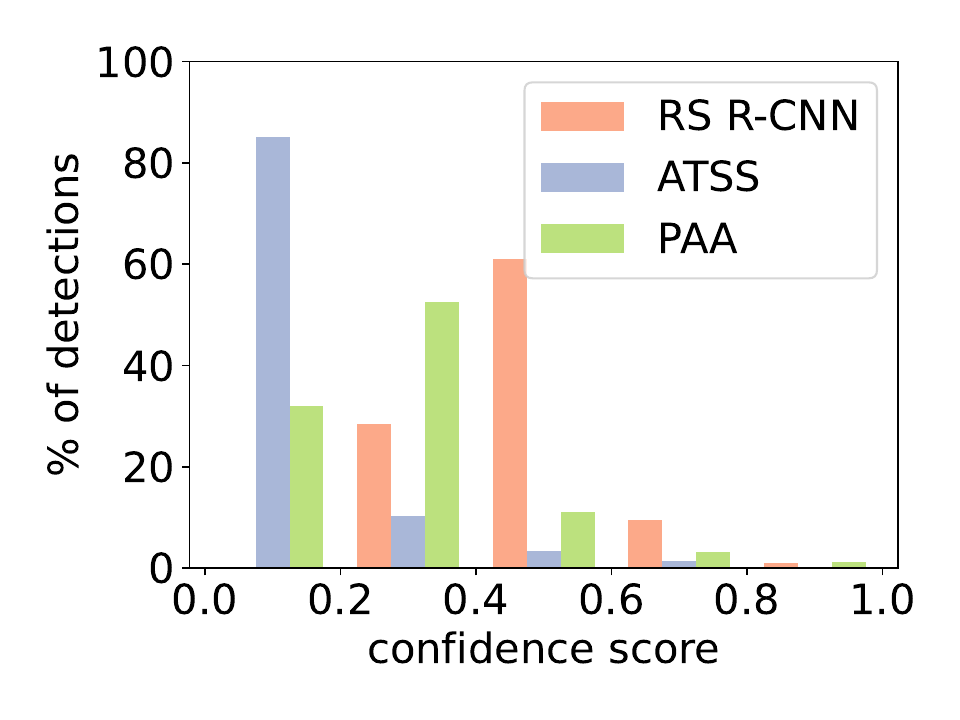}
             \caption{Uncalibrated confidence scores}
    \end{subfigure}
  \caption{Piecharts showing  \% of detections from three similarly performing  detectors in their resulting \gls{MoE}s on COCO dataset. \textbf{(a)} \gls{MoE} of uncalibrated detectors, \textbf{(b)} \gls{MoE} of calibrated detectors, and \textbf{(c)} histogram of confidence scores.}%
  \label{fig:teaser_method}
\end{figure}

\section{Introduction}\label{sec:introduction}
\begin{figure*}[t]
        \captionsetup[subfigure]{}
        \centering
        

        \begin{subfigure}[b]{0.3\textwidth}
            \includegraphics[width=\textwidth]{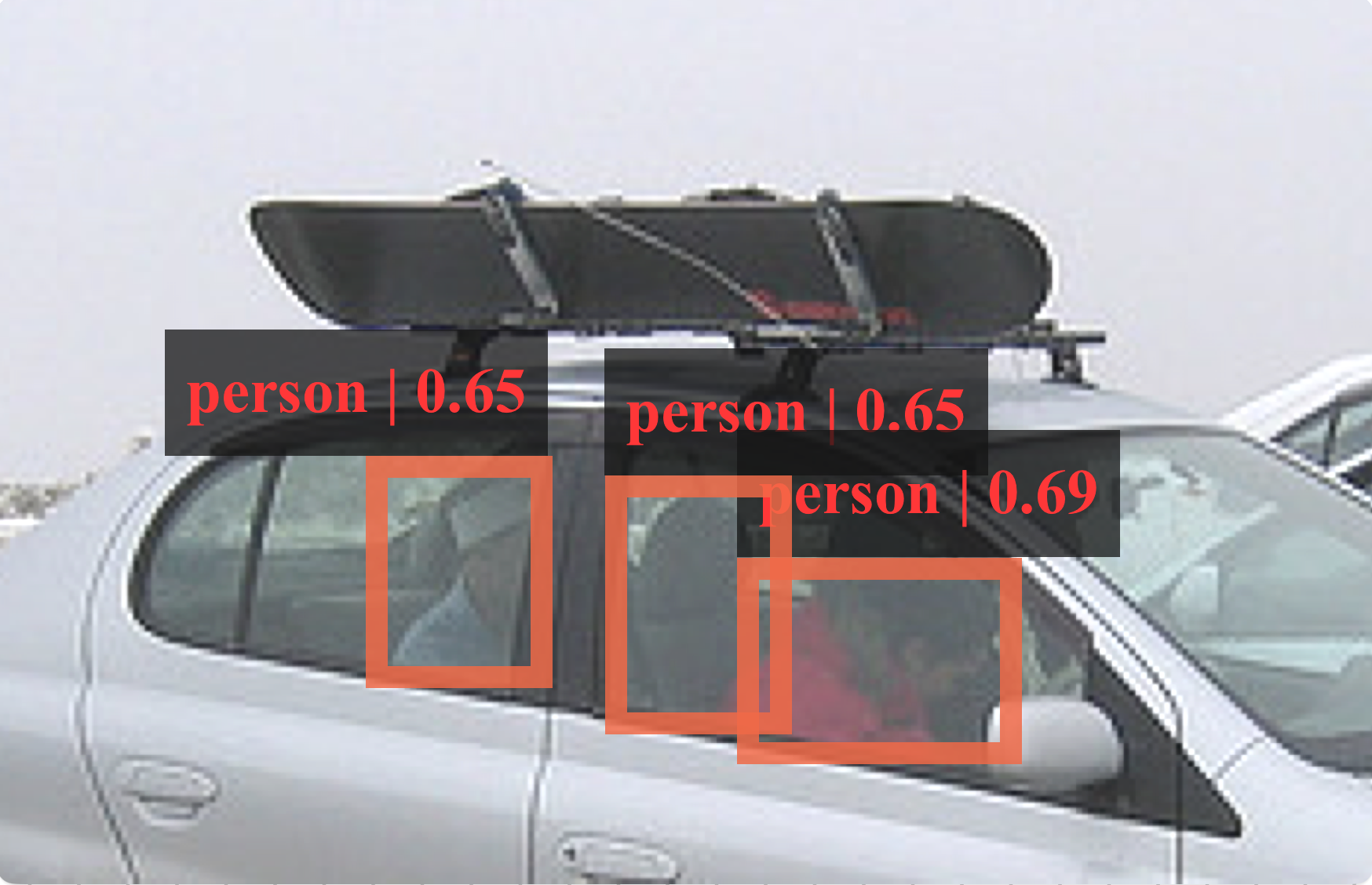}
            \caption{Expert 1: RS R-CNN}
        \end{subfigure}
        \begin{subfigure}[b]{0.3\textwidth}
            \includegraphics[width=\textwidth]{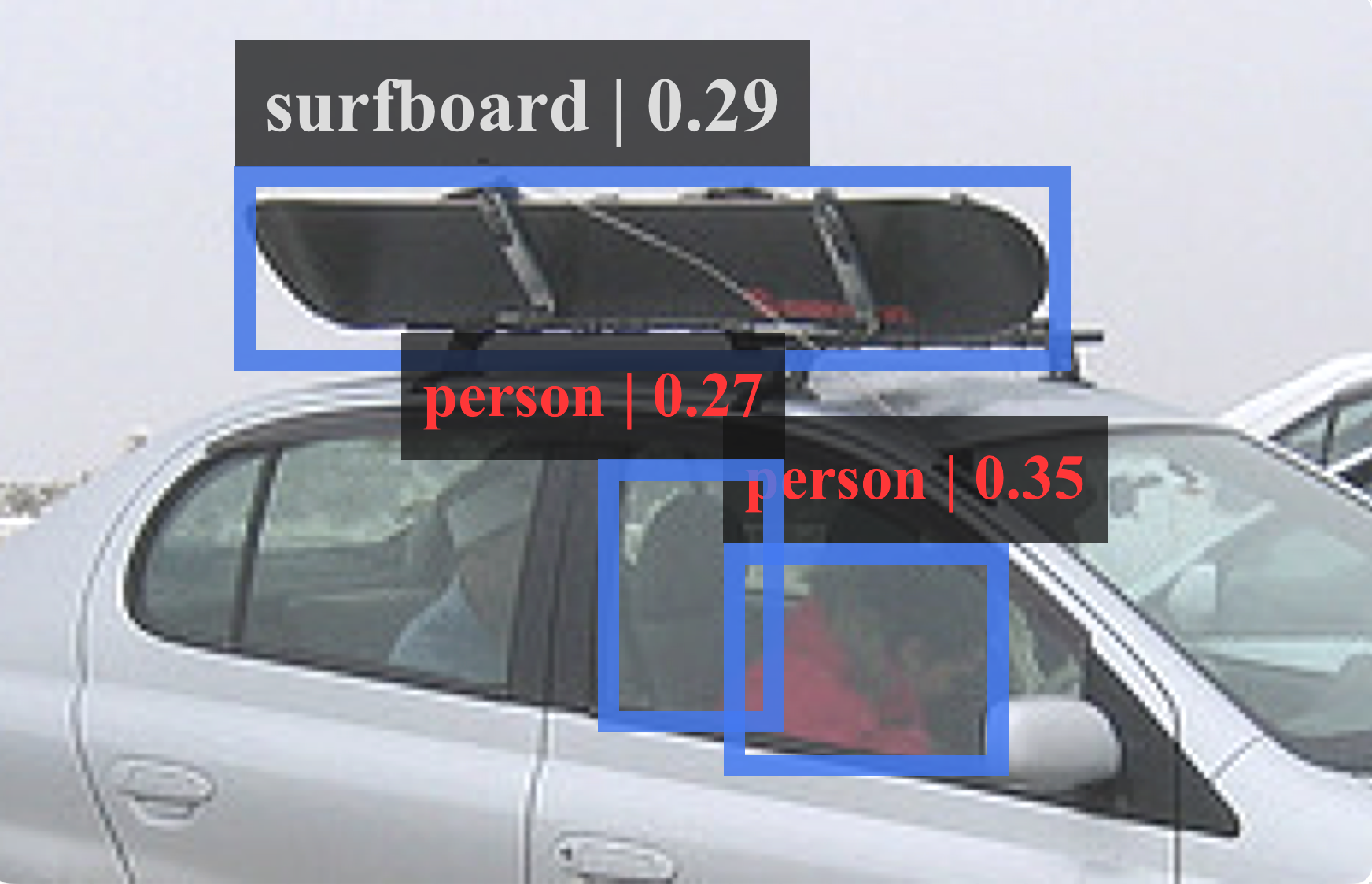}
            \caption{Expert 2: ATSS}
        \end{subfigure}
        \begin{subfigure}[b]{0.3\textwidth}
            \includegraphics[width=\textwidth]{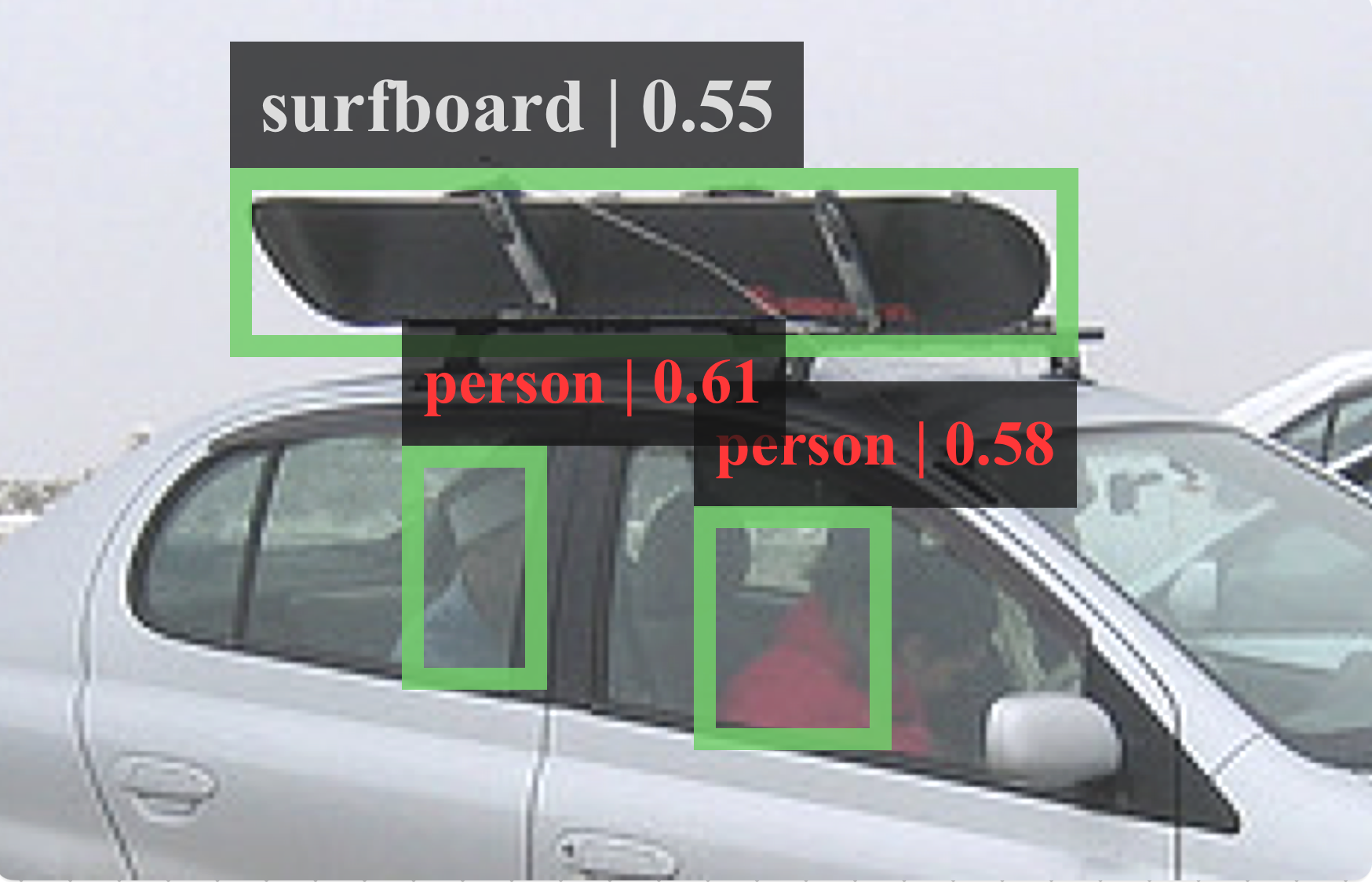}
            \caption{Expert 3: PAA}
        \end{subfigure}
        
        \begin{subfigure}[b]{0.3\textwidth}
            \includegraphics[width=\textwidth]{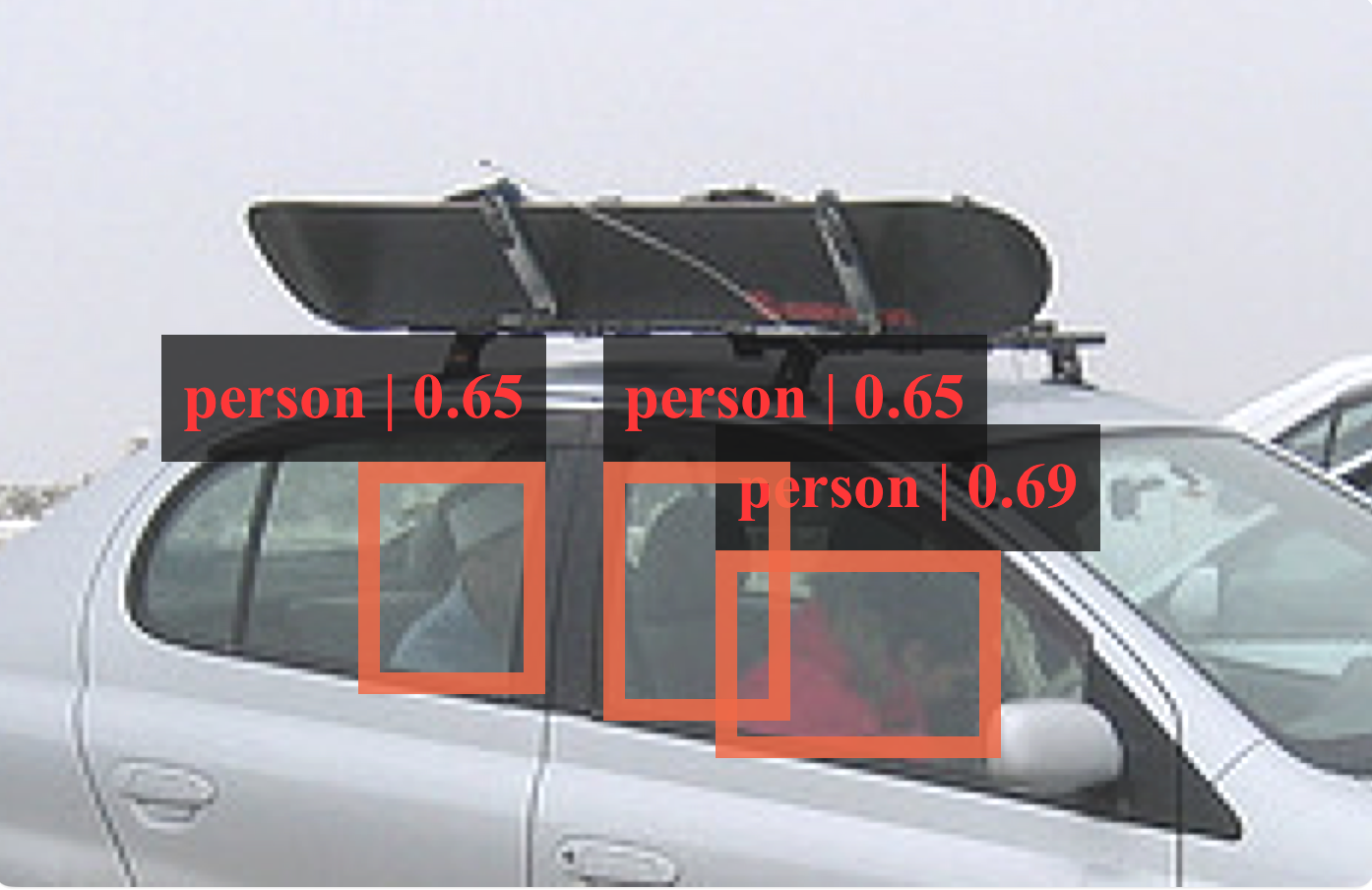}
            \caption{Mixture of Uncalibrated Experts}
        \end{subfigure}
        \begin{subfigure}[b]{0.3\textwidth}
            \includegraphics[width=\textwidth]{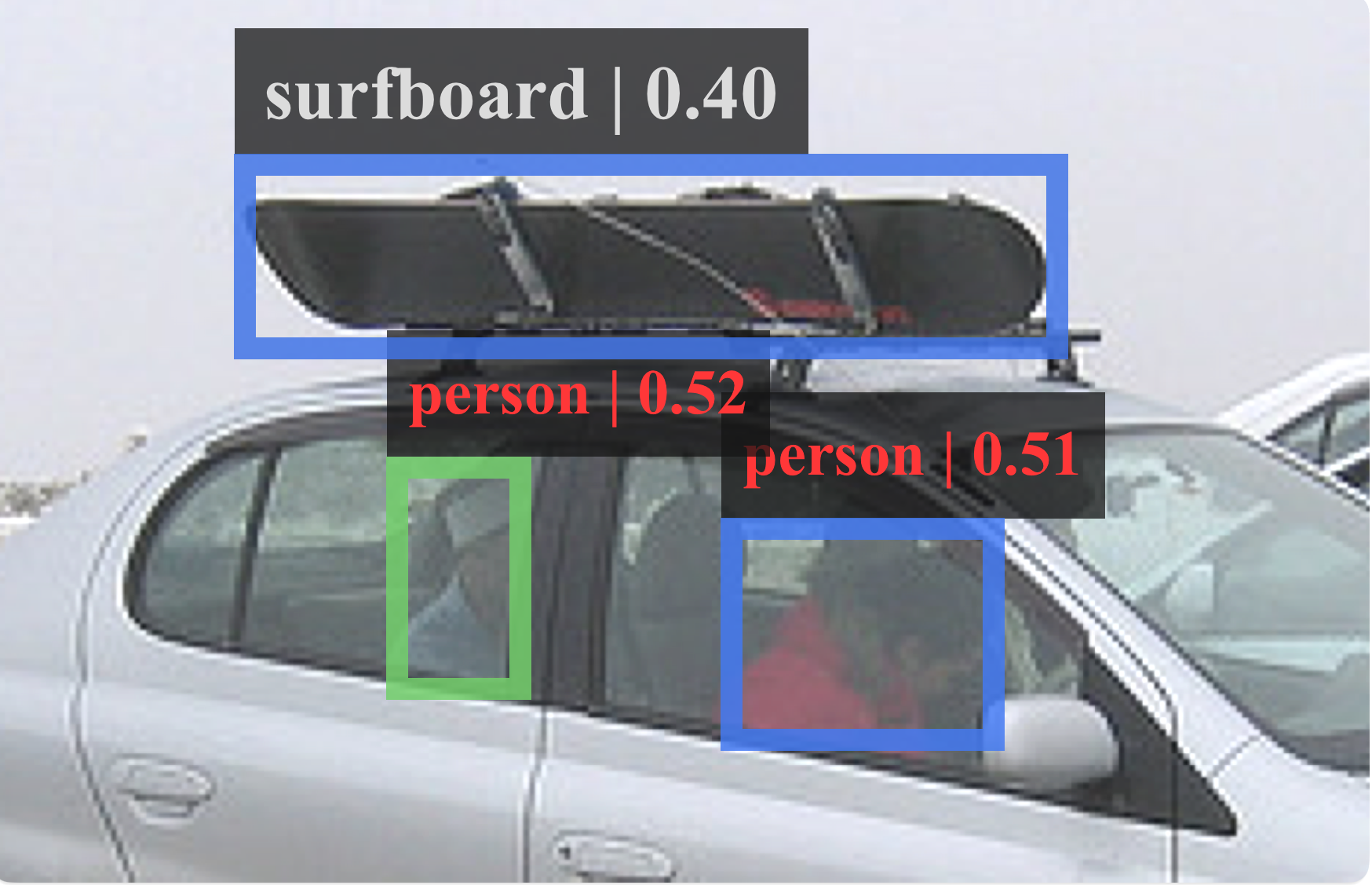}
            \caption{Mixture of Calibrated Experts (Ours)}
        \end{subfigure}
        \begin{subfigure}[b]{0.3\textwidth}
            \includegraphics[width=\textwidth]{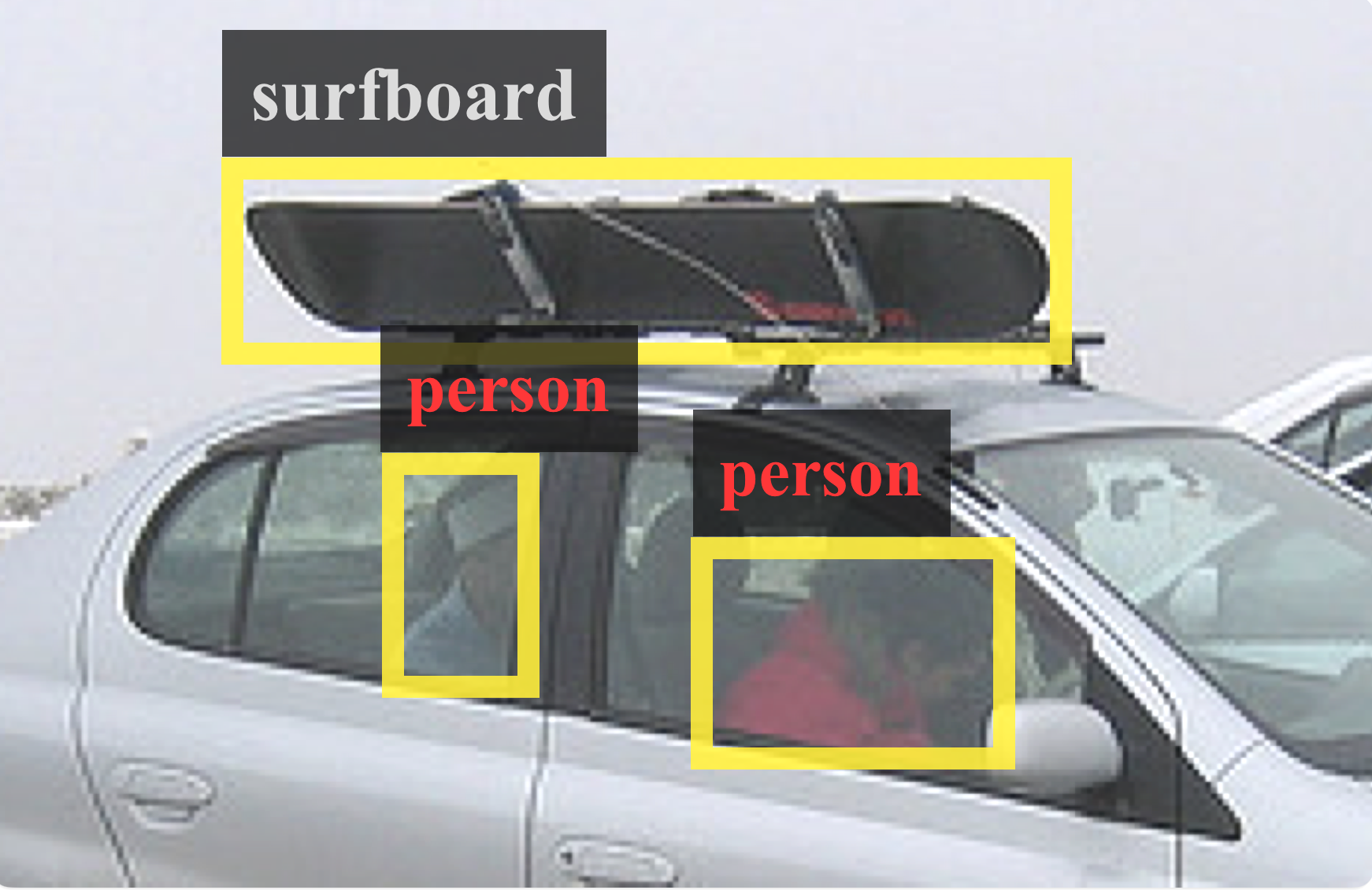}
            \caption{Ground Truth Objects}
        \end{subfigure}
        \caption{Detections are color-coded.  \textcolor{Red}{red}: RS R-CNN, \textcolor{NavyBlue}{blue}: ATSS, \textcolor{Green}{green}: PAA.  
        \textbf{(a-c)} Outputs of the detectors on an example image. RS R-CNN misses the ``surfboard'', ATSS misses a ``person'', PAA has a notable localisation error for the ``person'' in front seat. \textbf{(d-f)} The detections from \gls{MoE} of uncalibrated detectors; \gls{MoE} of calibrated detectors; and the ground truth. (d) is dominated by the most confident RS R-CNN and misses the ``surfboard''. After calibration in (e), all objects are detected accurately by improving each expert. \vspace{-2.ex}
        }
        \vspace{-2.ex}
        \label{fig:teaser}
\end{figure*}

\glspl{DE}~\cite{ensembles} is an effective method for obtaining improved performance by simply training multiple models before combining their predictions at inference time.
%
%
Providing that compute is accessible, and inference time is not a significant issue, this approach provides a significant boost in performance at minimal cost.
Another variant of this approach, the \gls{MoE} \cite{jacobs1991adaptive, jordan1994hierarchical, xu1994alternative, yuksel2012twenty}, which rather than sampling in the parameter space, samples in the function space, which in practice is achieved by combining different predictors.
Given that these experts will typically behave differently for different data samples, one would thus expect that the model is able to leverage the benefits of one whilst ignoring the contributions of the other poorer models.
Interestingly, when considering object detectors, we observe that na\"ively combining experts in the standard way often leads to a degradation in performance, resulting in an ~\gls{MoE} that is completely unable to leverage the strengths of the individual experts in certain situations.

We identify that the primary reason for this is due to a failure when combining the predictions, such that the final output \emph{does not} respect the individual performance of the experts, an issue which arises when the predicted confidences do not match the accuracy.
This inconsistency results in the most confident detector dominating the final predictions, regardless of its accuracy, as can be seen in~\cref{fig:teaser_method}(a), where, we see RS R-CNN dominating the predictions due to its high levels of confidence, which are shown in~\cref{fig:teaser_method}(c).

It is natural to ask why is this specific to \gls{MoE}? and not present in \gls{DE}?
For \gls{DE}, the main source of variation stems from the initialisation and other stochastic processes present in the optimisation, leading to similar histograms of predictive confidences.
However, for an \gls{MoE}, despite the fact that the experts perform similarly, there is a vast diversity in the mechanisms to arrive at the predictions: such as the use of an additional auxiliary localisation head~\cite{FCOS,ATSS,IoUNet,maskscoring,paa} or the choice of classifier, which commonly vary between a softmax \cite{FasterRCNN,DETR,RFCN,yolact} or sigmoid classifiers for each class \cite{ATSS,FocalLoss,paa,DDETR}.
Furthermore, different backbones \cite{impartial}, loss functions \cite{FocalLoss_Calibration} and the training length \cite{FocalLoss_Calibration,saod} can drastically affect the confidence of the model.

Consequently, given the vast diversity of methods, and the corresponding differences in their associated confidences for the prediction, it is imperative that for an effective \gls{MoE} to be constructed their confidences must match their performance; that is, they are said to be calibrated~\cite{calibration, saod}.
To address this, we propose \gls{CC}, which first calibrates the individual experts before combining the predictions using our refinement strategy, an approach we term as Refining \gls{NMS}.
The effects of our method can be seen in~\cref{fig:teaser_method}(b) and in~\cref{fig:teaser}, which shows that \gls{CC} is able to detect all objects in the scene with good localisation quality and avoid the \gls{FP} of the third person picked up by RS R-CNN and ATSS, but not by PAA.
Importantly, \gls{CC} is extremely simple to implement and leverage, requiring only a few parameters to be learnt in the calibration stage when using off the shelf detectors.
Overall, our contributions can be summarized as follows:
\begin{compactitem}
    \item We show that due to the diversity in training regimes for different detectors they consequently become miscalibrated in vastly different ways, resulting in \glspl{MoE} where the most confident expert dominates.
    \item To address this, we propose \gls{CC} which first calibrates the experts and then combines them through our refinement mechanism to make the prediction.
    \item We show that \gls{CC} yields significant gain over the single models and \glspl{DE} on different real world challenging detection tasks: such as (i) improving object detectors \textbf{by up to $\sim 2.5$ AP}; (ii) \textbf{reaching \gls{SOTA}} on COCO \textit{test-dev} with $65.1$ AP and on DOTA for rotated object detection with $82.62$ AP; (iii) outperforming single models consistently on recent detection tasks such as \gls{OVOD}.
\end{compactitem}

\section{Background and Notation} \label{sec:relatedwork}

Given that the set of $M$ objects in an image $X$ is represented by $\{b_i, c_i\}^M$ where $b_i \in \mathbb{R}^{4}$ is a bounding box and $c_i \in  \{1,\dots, K\}$ its class; the goal of an object detector is to predict the bounding boxes and the class labels for the objects in $X$, $f(X) = \{\hat{c}_i, \hat{b}_i, \hat{p}_i\}^N$, where $\hat{c}_i, \hat{b}_i, \hat{p}_i$ represent the class, bounding box and confidence score of the $i$th detection respectively and $N$ is the number of predictions.
In general, the detections are obtained in two steps, $f(X) = (h \circ g)(X)$~\cite{FasterRCNN,FocalLoss,DETR,sparsercnn}: 
where $g(X) = \{\hat{b}^{raw}_i, \hat{p}^{raw}_i \}^{N^{raw}}$ is a deep neural network predicting raw detections with bounding boxes $\hat{b}^{raw}_i$ and predicted class distribution $\hat{p}^{raw}_i$.
Then, in the second step, $h(\cdot)$ applies post-processing to raw-detections and the final detections are obtained. 
In general, $h(\cdot)$ consists of discarding the detections predicted as background; \gls{NMS} to remove the duplicates; and keeping useful detections, normally achieved via top-$k$ survival, where typically $k=100$ for COCO. 
Further discussion on background are provided in App. \ref{app:relatedwork}.

\section{Enabling Accurate \glspl{MoE} via \gls{CC}} \label{sec:mocae}

We seek to further examine the inconsistency of calibration errors for different detectors before proceeding to propose \gls{CC}. 
Specifically, in~\cref{subsec:diff_cal_errors}, we highlight the many reasons why detectors differ significantly in their confidence, and the consequences of this when constructing an \gls{MoE}.
To address this, \cref{subsec:mocae} proposes \gls{CC}, which calibrates the individual detectors before refining their predictions.

\label{sec:method}
\blockcomment{
\begin{figure}[t]
        \centering
        \includegraphics[width=0.98\textwidth]{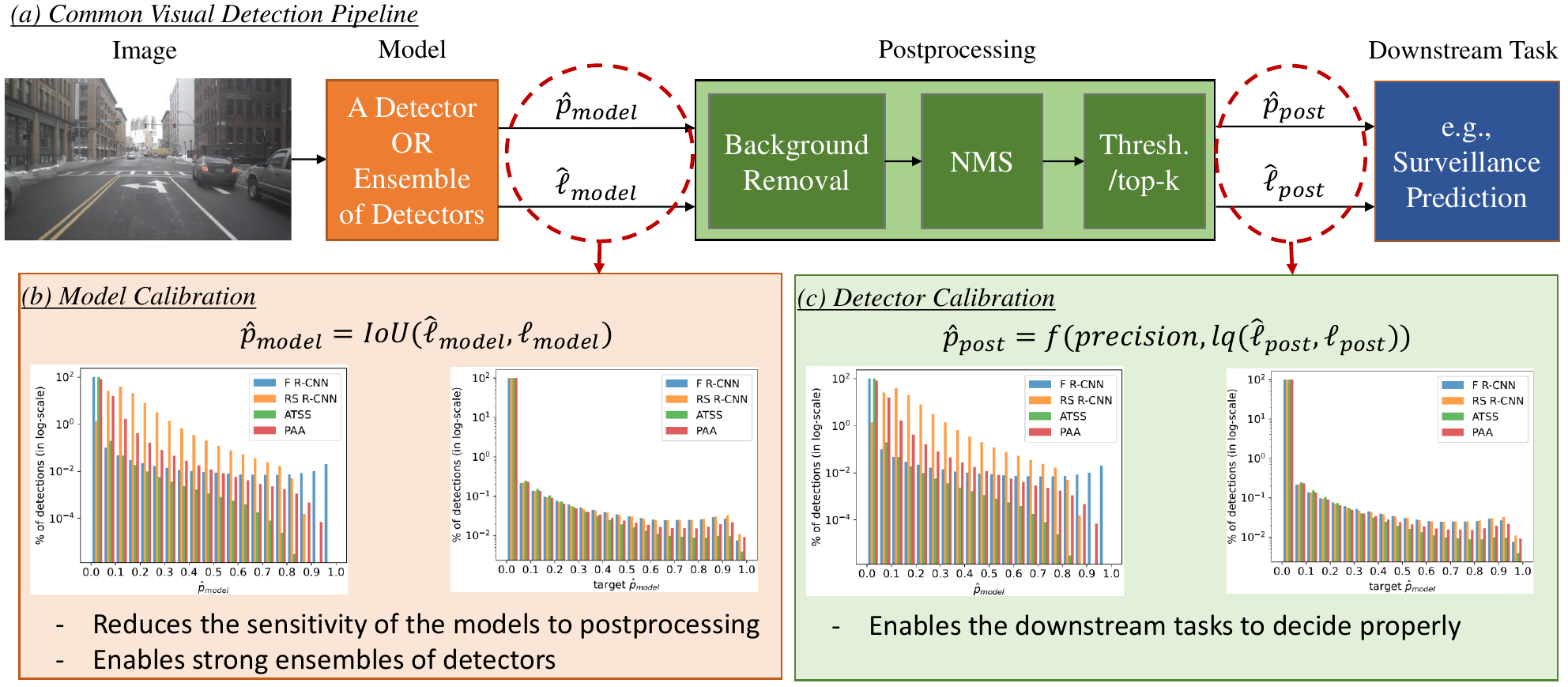}
        \caption{Detection pipeline and what we calibrate.
        }
        \label{fig:teaser2}
\end{figure} 
}

\subsection{Why do different detectors produce vastly different confidences?}\label{subsec:diff_cal_errors}
%
%
As already eluded to in~\cref{sec:introduction} among different factors causing this difference, one major factor is related to the parameterisation of the predictive function.
For example, some recent detectors employ an additional auxiliary head to predict localisation confidence \cite{FCOS,ATSS,IoUNet,maskscoring,paa}.
Consequently, the auxiliary head choice such as centerness \cite{FCOS, ATSS} or IoU \cite{IoUNet,paa} as well as the aggregation function such as multiplication \cite{FCOS,ATSS} or geometric mean \cite{paa} provides significant variation in the confidence scores.
Architectural difference can also manifest itself in the type of the detector, which can be fully convolutional one-stage~\cite{ATSS,FCOS}, two-stage~\cite{FasterRCNN,CascadeRCNN}, bottom-up~\cite{CornerNet,CenterNet} as well as transformer-based ~\cite{DETR,DDETR}.
%
Another factor causing the confidence incompatibility across the detectors is the used classifier, which commonly vary between a softmax \cite{FasterRCNN,DETR,RFCN,yolact} or sigmoid classifiers for each class \cite{ATSS,FocalLoss,paa,DDETR}.
Besides, different backbones \cite{impartial}, training objectives \cite{FocalLoss_Calibration,APLoss,aLRPLoss,seesawloss,FocalLoss,FocalLossConf,GFL,GFLv2,CorrLoss} and the training length \cite{FocalLoss_Calibration,saod} affect the confidence of the model.

\begin{table}[t]
    \small
    \centering
    \begin{tabular}{c||c|c|c} 
    \toprule
    \midrule
    Calibration&RS R-CNN&ATSS&PAA \\ \midrule
    \xmark&$36.45$&$5.01$&$11.23$\\
    \cmark&$\mathbf{3.15}$&$\mathbf{4.51}$&$\mathbf{1.62}$\\ 
    \midrule 
    \midrule
    Method&R@0.50&R@0.75&$\mathrm{AR}$ \\ \midrule
    RS R-CNN (1) &$83.2$&$62.7$&$58.3$\\
    ATSS (2) &$83.1$&$65.9$&$60.8$\\
    PAA (3) &$83.4$&$65.8$&$61.1$\\  
    Uncal. \gls{MoE} (1+2+3) &$83.6$&$64.1$&$59.7$\\
    Cal. MoE (1+2+3) &$\mathbf{85.1}$&$\mathbf{67.7}$&$\mathbf{62.3}$\\ \midrule 
    \bottomrule
    \end{tabular}
    \vspace{-1.5ex}
    \caption{\textbf{(Top)} \gls{LaECE} of detectors before and after calibration (cal.). \textbf{(Bottom)} Recall@\gls{IoU} (R@IoU) and \gls{AR} using 100 detections. \vspace{-2.5ex}} \label{tab:teaser_method}
\end{table}

\begin{figure}[t!]
    \centering
    \parbox{\linewidth}{
    \ffigbox{%
              \begin{subfigure}[b]{0.49\textwidth}
                \includegraphics[width=\textwidth]{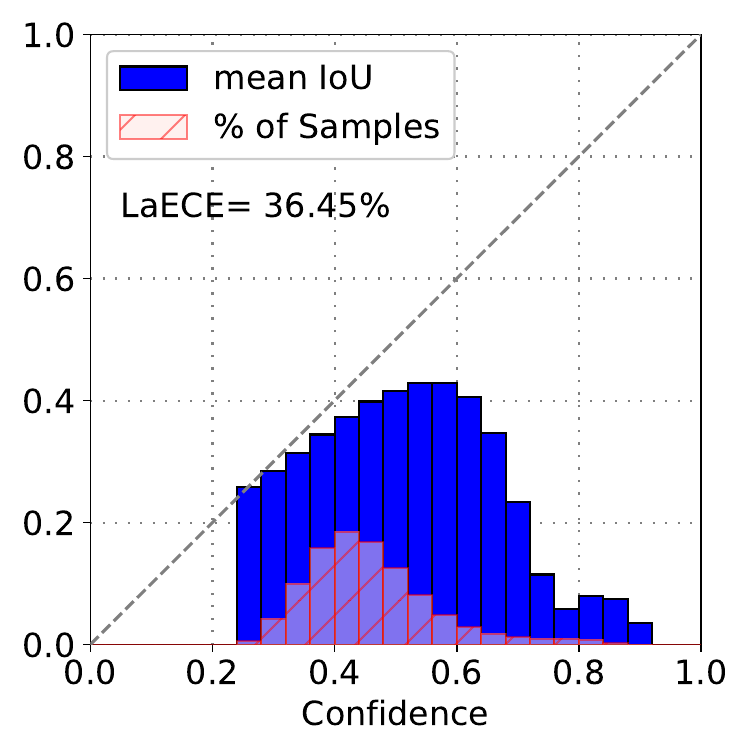}
                \caption{RS R-CNN}
            \end{subfigure}
            \begin{subfigure}[b]{0.49\textwidth}
                \includegraphics[width=\textwidth]{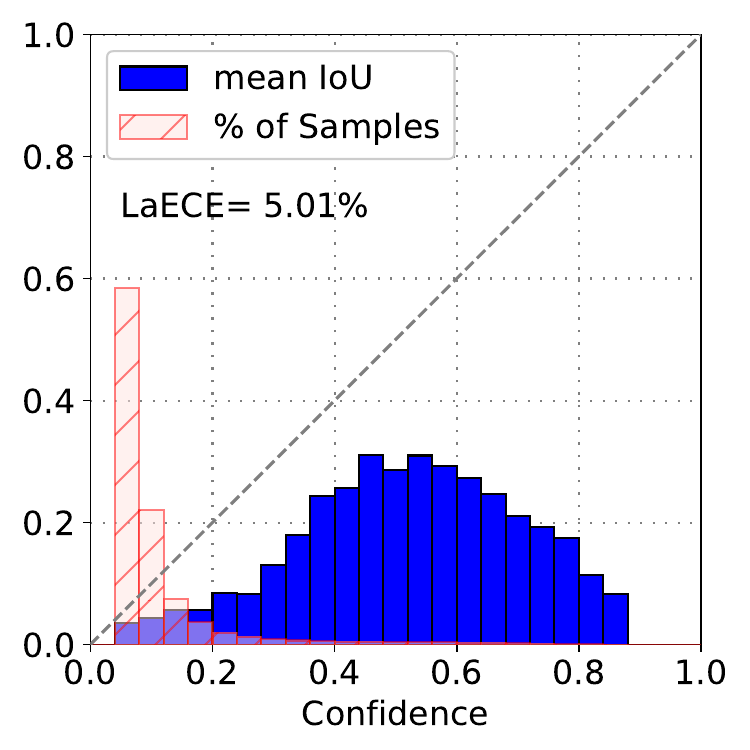}
                \caption{ATSS}
            \end{subfigure}
    }{%
\vspace{-1.5ex}
  \caption{Reliability diagrams \cite{calibration} of \textbf{(a)} RS R-CNN and \textbf{(b)} ATSS. \vspace{-2.5ex}}%
  \label{fig:reliability_plots}
    }
    }
\end{figure}

\paragraph{How different are the predicted confidences?}
To evaluate this, we leverage the recently proposed \gls{LaECE}~\cite{saod}, which returns the absolute difference between the accuracy and the assosiated confidence, for each class it is defined as
\begin{align}
\label{eq:laece}
   \text{\small\gls{LaECE}$^c 
    = \sum_{j=1}^{J} \frac{|\hat{\mathcal{D}}^{c}_j|}{|\hat{\mathcal{D}}^{c}|} \left\lvert \bar{p}^{c}_{j} - \mathrm{precision}^{c}(j) \times \bar{\mathrm{IoU}}^{c}(j)  \right\rvert$}.
\end{align}
with $\hat{\mathcal{D}}^{c}$ denoting the set of detections and those in the $j$th bin by $\hat{\mathcal{D}}^{c}_j$ as well as the average confidence, precision and average \gls{IoU} of $\hat{\mathcal{D}}^{c}_j$ by $\bar{p}^{c}_{j}$, $\mathrm{precision}^{c}(j)$ and $\bar{\mathrm{IoU}}^{c}(j)$ respectively\footnote{We use \gls{LaECE} by measuring the difference of the confidence from the \gls{IoU} of the prediction following our calibration objective in Eq. \eqref{eq:gen_calibration_}, which will be introduced shortly. Formal definition is presented in App. \ref{app:calibration}}.
%
%
We report \gls{LaECE} in \cref{tab:teaser_method} (top) where we see that RS R-CNN, PAA and ATSS, all perform similarly in terms of AP, but have vastly different \gls{LaECE}, implying different confidence predictions.
Moreover, in~\cref{fig:reliability_plots}, we display the reliability plots for RS R-CNN and ATSS, which shows that RS R-CNN is significantly more confident than ATSS.

\paragraph{An uncalibrated Mixture of Experts}
%
For the purposes of exposition, we first show that na\"ively combining these uncalibrated detectors results in a poor \gls{MoE}, where the most confident detector dominates the \gls{MoE} regardless of its accuracy.
%
%
To show this, similar to~\cite{ensembleod}, we construct a ``Vanilla \gls{MoE}'' which aggregates the \emph{uncalibrated} predictions from RS R-CNN \cite{RSLoss}, ATSS \cite{ATSS} and PAA \cite{paa} using \gls{NMS}.
\cref{fig:teaser_method}(a) shows that Vanilla \gls{MoE} is dominated by the most confident RS R-CNN with a small contribution from less confident PAA and almost no detections from the least confident ATSS.
Furthermore, we see in~\cref{fig:teaser}(d), that it is unable to identify the \texttt{surfboard} and produces a \gls{FP} for a person.
%
%
As a result, while one would expect an \gls{MoE} to detect more objects than individual detectors and obtain a better recall, \cref{tab:teaser_method}(bottom) shows the opposite; \textit{Vanilla \gls{MoE} yields a lower Average Recall (AR) compared to ATSS and PAA.}
This clearly indicates that na\"ively obtaining \gls{MoE} will normally be biased and lead to an ineffective mixture.
%

We have now highlighted that a fundamental issue with constructing an \gls{MoE}, is that a situation can often arise when one of the detector dominates the predictions.
However, this is not necessarily a deficiency, as one would expect an accurate detector to dominate the predictions when combined with inaccurate ones.
%
%
Conceptually, we want the \gls{MoE} to combine predictions based on their performance, which can be inferred through the confidence estimates provided at test time.
However, as shown above, it is imperative that these predictions are calibrated.
Therefore, to appropriately construct the \gls{MoE}, we calibrate the experts individually, before filtering the predictions in our refinement strategy.
As we show in~\cref{sec:exp}, this enables reliable contributions from each detector and an effective \gls{MoE}.

%
%

\begin{figure}[t]
\centering
\includegraphics[width=\textwidth]{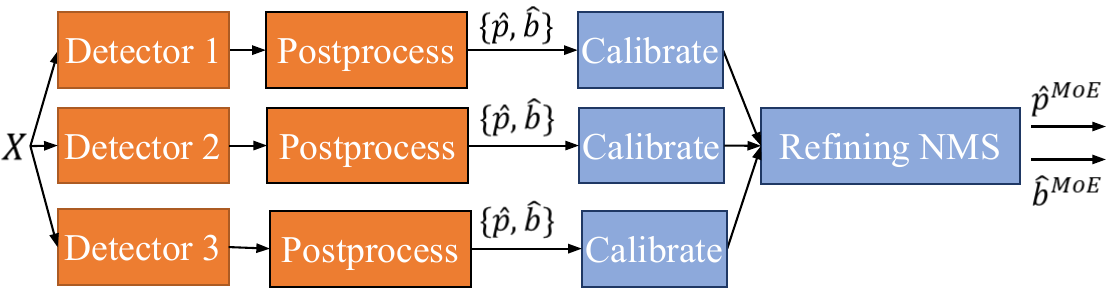}
\caption{\gls{CC} pipeline. Given an image $X$, each detector follows its own pipeline including postprocessing (in orange) and outputs $\{\hat{p},\hat{b}\}$. Without any modification to the pipeline of each detector, we introduce two modules in blue: (i) calibrate the confidence scores of each detector; (ii) aggregate them via Refining \gls{NMS} providing the detections of \gls{CC} $\{\hat{p}^{MoE},\hat{b}^{MoE}\}$. \vspace{-2.5ex}}
\label{fig:cc}
\end{figure}

%
%
%
%
%
%
%

\subsection{Constructing an Effective Mixture of Experts}\label{subsec:mocae}
Here we highlight the two main components for constructing an effective~\gls{MoE}, obtaining similarly performing calibrated experts; and aggregating their detections in the best way possible. 
%
Overall pipeline of \gls{CC} is presented in~\cref{fig:cc}.

\subsubsection{Calibrating Individual Experts}
\label{subsec:calibrate}
Having identified the issue with the Vanilla \gls{MoE}, the question naturally arises as to how we calibrate the single detectors to address this deficiency.
As opposed to the standard classification task, object detection jointly solves both classification and regression tasks; and also involves post-processing steps that can influence the accuracy of the detector. 
Therefore, it is not straightforward as to what objective the final calibrator should have and at which stage of the pipeline it should be applied.
A natural choice would be to calibrate the scores such that it helps the most crucial aggregation stage (e.g., \gls{NMS}). 
This stage does not require training and has significant impact on the accuracy of a detector.

For simplicity, let's consider the standard \gls{NMS}, which groups the detections that have an \gls{IoU} with the maximum-scoring detection larger than a predefined \gls{IoU} threshold.
Then, within that group, \gls{NMS} survives the detection with the largest score and removes the remaining detections from the detection set.
%
%
In such a setting, as also discussed by the recent works \cite{GFL,GFLv2,CorrLoss,IoUNet,ATSS}, the ideal confidence that should be transferred to the \gls{NMS} is the \gls{IoU} of the detection with the object.
This will guide \gls{NMS} to pick \textit{accurately-localised detections} for the objects detected by multiple detectors.
%
%
Furthermore, if an object is detected by a single less confident detector, aligning the confidence with \gls{IoU} implies that the scores of the \gls{TP}s are to be promoted.
Thus, the \glspl{TP} of a less confident detector will not be dominated by the \glspl{FP} of more confident ones unlike the case in \cref{fig:teaser_method}(a).
Following this intuition, we call a detector calibrated if it yields a confidence that matches the \gls{IoU}, implying
%
%
\begin{align}\label{eq:gen_calibration_}
    \mathbb{E}_{\hat{b}_i \in B_i(\hat{p}_i)}[ \mathrm{IoU}(\hat{b}_i, b_{\psi(i)})] = \hat{p}_i, \forall \hat{p}_i \in [0,1],
\end{align}
where $B_i(\hat{p}_i)$ is the set of detection boxes with the confidence score of $\hat{p}_i$ and $b_{\psi(i)}$ is the ground-truth box that $\hat{b}_i$ has the highest \gls{IoU} with.
%
%

From an optimization perspective, calibrating each expert to meet the criterion in Eq.~\eqref{eq:gen_calibration_} requires us to design an objective that maps the output confidence of each bounding box to a calibrated one. 
Though there can be several ways to design such an objective, we take a rather simple approach where we learn a post-hoc calibrator $\zeta_\theta:[0,1] \rightarrow [0,1]$ using the input-target pairs ($\{\hat{p}_i, \mathrm{IoU}(\hat{b}_i, b_{\psi(i)})\}$) obtained on a held-out validation set.
Specifically, we parameterise $\zeta_\theta(\cdot)$ as a simple \gls{LR} model, containing only two learnable parameters or an \gls{IR} model \cite{zadrozny2002transforming,saod}.
Thereby being easily applicable to any off-the-shelf detector without adding any notable overhead.

\cref{tab:teaser_method} shows an example case using \gls{IR}, where the \gls{LaECE} improves and the resulting \gls{MoE} has a higher recall than the uncalibrated \gls{MoE} and single models. 
%
%
Please refer to App. \ref{app:calibration} and App. \ref{app:exp} for more detail.
Furthermore, \cref{theorem:mocae} (provided in App.~\ref{app:proof}) indicates that Eq. \eqref{eq:gen_calibration_} is the optimal choice in terms of \gls{AP} in obtaining an \gls{MoE} for object detection.
%

%
%

\blockcomment{
\begin{theorem}
\label{theorem:mocae}
For the sake of simplicity in \gls{NMS} algorithm, we ensure that the detections targeting different ground truths do not overlap. Specifically, denoting the threshold to validate \glspl{TP} by $\tau$, we assume that while aggregating detections from different experts \gls{NMS} operates as follows: (i) for the detections with $\mathrm{IoU} \geq \tau$ with a ground truth box $b_i$, \gls{NMS} survives the maximum scoring detection and (ii) for the rest of the detections, that is the ones with $\mathrm{IoU} < \tau$ for all ground truths, the conventional NMS algorithm is followed. Then, the \gls{MoE} using perfectly calibrated experts in terms of Eq. \eqref{eq:gen_calibration_} yields the upper bound AP for IoU threshold $\tau$ of $X$ over any possible \glspl{MoE}.
\begin{proof}
    Appendix~\ref{app:proof} provides the proof and  discussions.
\end{proof}
\end{theorem}{}
}

\blockcomment{
\begin{theorem}
\label{theorem:mocae}
While aggregating the detections from multiple experts, we assume that the standard \gls{NMS} is used from an \gls{IoU} threshold of $\mathrm{IoU_{NMS}}$. Furthermore, for the sake of simplicity, we also assume that \gls{NMS} keeps the maximum scoring detection among the ones that target each ground truth. Specifically, any two ground truth bounding boxes $b_i$ and $b_j$ in an image $X$ are sufficiently far apart such that $\mathrm{IoU}(\hat{b}_i,\hat{b}_j) < \mathrm{IoU_{NMS}}$ if $\hat{b}_i$ is the maximum scoring detection among all detections from different experts with $\mathrm{IoU}(\hat{b}_i,b_i) \geq \tau$ and $\hat{b}_j$ satisfies the same for $b_j$ where $\tau$ is the \gls{IoU} threshold for validating \glspl{TP}. Then, the \gls{MoE} using perfectly calibrated experts in terms of Eq. \eqref{eq:gen_calibration_} yields the upper bound AP for any $\tau$ of $X$ over any possible \glspl{MoE}.
\begin{proof}
    Appendix~\ref{app:proof} provides the proof and  discussions.
\end{proof}
\end{theorem}{}
}
\subsubsection{Refining \gls{NMS} for Aggregating Detections}

\label{subsec:cluster}
%
Another critical component of the \gls{MoE} is aggregating the combined detections.
There is a very high chance that more than one detector produces the same detection; therefore we aim to suppress these duplicate detections targeting the same object and obtain detections with high localisation quality.
As aforementioned, \gls{NMS} is a method that fits for this purpose and, as we observe experimentally, does provide highly competitive results.
However, it is rigid in nature when removing overlapping detections, thereby not utilizing the rich information provided by multiple \glspl{MoE}.
To address this, we present \textit{Refining NMS} that simply combines Soft NMS~\cite{SoftNMS} with Score Voting~\cite{paa}. 
Soft NMS, by design, instead decreases the scores of overlapping boxes that naturally leads to improved recall. 
Score Voting combines multiple overlapping detections (using their confidences and \gls{IoU}s with each other) and obtains a refined detection with better localisation. 
Thus, combining these two approaches leads to a much effective aggregator. 
Please see App. \ref{app:cc} for details.

\blockcomment{
\begin{table}[t]
\RawFloats
\parbox{.44\linewidth}{
    \small
    \setlength{\tabcolsep}{0.5em}
    \centering
    \caption{Object detection performance on COCO test-dev split \cite{COCO}. \underline{Underlined}: Best single model, \textbf{Bold}: Best overall}
    \label{tab:teaser_od}
    \scalebox{1.00}{
    \begin{tabular}{c||c|c|c} 
    \toprule
    \midrule
    Method&$\mathrm{AP}$&$\mathrm{AP_{50}}$&$\mathrm{AP_{75}}$ \\ \midrule
    YOLOv7 \cite{yolov7}&$55.5$&$73.0$&$60.6$\\
    QueryInst \cite{queryinst}&$55.7$&$\underline{75.7}$&$61.4$\\
    DyHead \cite{dyhead}&$\underline{56.6}$&$75.5$&$\underline{61.8}$\\ \midrule 
    Vanilla MoE &$57.6$&$76.6$&$63.2$\\
    &$\imp{1.0}$&$\imp{1.1}$&$\imp{1.4}$ \\ \midrule 
    Late C\&C &$\mathbf{59.0}$&$\mathbf{77.2}$&$\mathbf{64.7}$ \\
    (Ours)&$\imp{2.4}$&$\imp{1.5}$&$\imp{2.9}$\\ \midrule 
    \bottomrule
    \end{tabular}
    }
}
\hfill
\parbox{.53\linewidth}{
    \small
    \setlength{\tabcolsep}{0.5em}
    \centering
    \caption{Instance segmentation performance on LVIS validation set \cite{LVIS}. \underline{Underlined}: Best single model, \textbf{Bold}: Best overall}
    \label{tab:teaser_segm}
    \scalebox{1.00}{
    \begin{tabular}{c||c|c|c} 
    \toprule
    \midrule
    Method&$\mathrm{AP}$&$\mathrm{AP_{50}}$&$\mathrm{AP_{75}}$ \\ \midrule
    Seesaw Mask R-CNN \cite{seesawloss}&$\underline{25.4}$&$\underline{39.5}$&$26.9$\\
    RS Mask R-CNN \cite{RSLoss}&$25.1$&$38.2$&$26.8$\\
    Mask R-CNN \cite{MaskRCNN}&$\underline{25.4}$&$39.2$&$\underline{27.3}$\\ \midrule 
    Vanilla MoE &$25.2$&$38.3$&$26.8$\\
    &$\nimp{0.2}$&$\nimp{1.2}$&$\nimp{0.5}$ \\ \midrule 
    Late C\&C &$\mathbf{27.7}$&$\mathbf{42.8}$&$\mathbf{29.4}$ \\
    (Ours)&$\imp{2.3}$&$\imp{3.3}$&$\imp{2.1}$\\ \midrule 
    \bottomrule
    \end{tabular}
    }
}
\end{table}

\paragraph{\gls{CC} results in significant performance gain even for challenging cases.}
While our approach is extremely simple, we observe that it addresses an important gap in the literature that can allow the practitioners to easily boost the performance of their detectors.
To demonstrate this, here we present the initial results of our Late \gls{CC} approach on two challenging scenarios.
In the first scenario, we train calibrators on COCO validation set for YOLOv7 \cite{yolov7}, QueryInst \cite{queryinst} and ATSS with Dynamic Head \cite{dyhead} as very strong detectors shown in \cref{tab:teaser_od}(a).
We observe that our Late \gls{CC} approach improves the best single method by $2.4$ AP, which is a significant boost especially for such strong detectors.
Note that these three detectors are among the most strong detectors that are publicly available for COCO dataset \cite{COCO} and most of the best performing detectors are not available publicly.
Considering the performance boost here, we would easily expect that our improvement generalizes to the best set of detectors as well.
In the second case, we investigate whether Late \gls{CC} generalizes to instance segmentation on the challenging long-tailed LVIS dataset \cite{LVIS} with 1.2K classes.
In \cref{tab:teaser_od}(b), while an uncalibrated MoE degrades the performance, Late \gls{CC} achieves $2.3$ mask AP improvement again on this challenging scenario.
Having shown the initial results, we next present further empirical evidence for our claims and insights on our \gls{CC}.

One obvious approach is to apply \gls{CC} on the raw probabilities and then utilize the standard post-processing of the visual detectors for all of the detectors. 
This approach is shown in \cref{fig:cc}(a) in which we only introduce calibration and concatenation operations into the detection pipeline.
Finally, an \gls{NMS}-based algorithm within the post-processing steps can handle clustering the detection and removing the duplicate detections as we will discuss very soon.

The calibration targets should be useful for NMS to operate properly. 
For this reason, we expect the raw probabilities to be aligned with their \gls{IoU} as the localisation quality.
This will enable \gls{NMS} to choose the detection with the largest \gls{IoU} as the detection with the highest confidence is survived by \gls{NMS} within each cluster.
Formally speaking, denoting the set of detection boxes with the confidence score of $\hat{p}_i$ by $B_i(\hat{p}_i)$, and $b_{\psi(i)}$ is the ground-truth box that $\hat{b}^{raw}_i$ has the largest overlap, the calibration criterion for Early \gls{CC} is:

\begin{align}\label{eq:gen_calibration_}
    \mathbb{E}_{\hat{b}^{raw}_i \in B_i(\hat{p}^{raw}_i)}[ \mathrm{IoU}(\hat{b}^{raw}_i, b_{\psi(i)})] = \hat{p}^{raw}_i, \forall \hat{p}^{raw}_i \in [0,1].
\end{align}

\paragraph{Late Calibrate \& Cluster} Another alternative to incorporate \gls{CC} is after the detection pipeline on the final probabilities.
As \cref{fig:cc} suggests this includes detector-specific calibrators applied to the final detections, concatenating them and jointly clustering by an additional \gls{NMS}-based algorithm.
Similar to Early \gls{CC}, the calibration criterion can formally be defined as:
\begin{align}\label{eq:gen_calibration_2}
    \mathbb{E}_{\hat{b}_i \in B_i(\hat{p}_i)}[ \mathrm{IoU}(\hat{b}_i, b_{\psi(i)})] = \hat{p}_i, \forall \hat{p}_i \in [0,1],
\end{align}
where the only difference from Eq. \eqref{eq:gen_calibration_} is that Eq. \eqref{eq:gen_calibration_2} operates on final detections, that are the detections obtained after the post-processing step of individual detectors.
}

\blockcomment{
\paragraph{Measuring Calibration Error} We use Localisation-aware calibration error variants to measure the miscalibration in terms of Early \gls{CC} in Eq. \eqref{eq:gen_calibration_} and Late \gls{CC} in Eq. \eqref{eq:gen_calibration_2}.
Specifically, matching the detection boxes with the ground truth box that it has the highest overlap implies that all detection boxes are matched and hence $\mathrm{precision}^{c}(j)=1$ in Eq. \eqref{eq:laece} due to this matching.
Therefore, in our case, \gls{LaECE} for class $c$ reduces to:
\begin{align}
\label{eq:laece_nms}
   \mathrm{LaECE}^c 
    = \sum_{j=1}^{J} \frac{|\hat{\mathcal{D}}^{c}_j|}{|\hat{\mathcal{D}}^{c}|} \left\lvert \bar{p}^{c}_{j} - \bar{\mathrm{IoU}}^{c}(j)  \right\rvert,
\end{align}
Furthermore, note that we use a very large number of detections for calibration and measuring the error especially in the case of Early \gls{CC} relying on raw probabilities.
Aligned with the observation of Oksuz et al., we observed that the large number low-scoring calibration targets decreases the sensitivity of \gls{LaECE}.
Alternatively, we also define Localisation-aware versions of Average Calibration Error and Maximum Calibration Error, which are commonly used in the calibration literature.
Specifically, \gls{LaACE} is defined as the average of calibration errors of the bins
\begin{align}
\label{eq:laace_nms}
   \mathrm{LaACE}^c 
    = \sum_{j=1}^{J} \frac{1}{|J|} \left\lvert \bar{p}^{c}_{j} - \bar{\mathrm{IoU}}^{c}(j)  \right\rvert,
\end{align}
and \gls{LaMCE} is their maximum:
\begin{align}
\label{eq:lamce_nms}
   \mathrm{LaMCE}^c 
    = \max_{j\in J} \left\lvert \bar{p}^{c}_{j} - \bar{\mathrm{IoU}}^{c}(j)  \right\rvert.
\end{align}
While we stick to these definitions for our purpose, note that these calibration errors can easily extended to include the precision term as in the case of Eq. \eqref{eq:laece}.
Specifically, Soft NMS decreases the confidence of a detection by considering its overlap with $\dot{b}^{\mathrm{NMS}}_i$.
For example, using a Gaussian penalty function, the updated score for a detection is:
\begin{align}
    \label{eq:softnms_g}
    p_i^{\mathrm{NMS}} = p_i^{\mathrm{NMS}} \exp\left( \frac{-\mathrm{lq}(b_i^{\mathrm{NMS}}, \dot{b}^{\mathrm{NMS}}_i)^2}{\sigma^{\mathrm{NMS}}} \right)
\end{align}
Furthermore, we also update the bounding box following 

\begin{align}
    \label{eq:softnms_g}
    p_i^{\mathrm{NMS}} = p_i^{\mathrm{NMS}} \exp\left( \frac{-\mathrm{lq}(b_i^{\mathrm{NMS}}, \dot{b}^{\mathrm{NMS}}_i)^2}{\sigma^{\mathrm{SV}}} \right)
\end{align}

considering the stronger alternatives of NMS \cite{SoftNMS,solov2,paa,KLLoss,yolact}, we design a

{\small
\begin{align}
\label{eq:modelcalibrationerror}
 \sum_{c=1}^{C} \frac{1}{C} \left( \sum_{j=1}^{J} \frac{|\hat{\mathcal{D}}_{c,j}|}{|\hat{\mathcal{D}}_{c}|} \left\lvert \bar{p}_{c,j} -  \bar{\mathrm{IoU}}(\hat{\mathcal{D}}_{c,j}) \right\rvert \right) ,
\end{align}}

\section{Postprocessing Calibration}

In addition to the model-calibration that targets NMS, we also expect the final scores of the detections to match the accuracy of the downstream task. 
More specifically, this level of calibration is necessary for two main reasons: the difference of the targets of NMS and the downstream task, as well as the imperfect accuracy of the detector as proven by the following theorem.
\begin{theorem}
A model-calibrated and recall-accurate detector is postprocessing-calibrated and achieves perfect down-stream accuracy in terms of any measure combining precision, recall and localisation errors once thresholded from $\bar{v}$ using any threshold $\bar{v}$ satisfying $1 > \bar{v} > \tau^{\mathrm{post}} >\mathrm{max}(\mathrm{IoU}(b_i, b_j))$ if the $\mathrm{lq}(\cdot, \cdot)$ of the post-processing calibration is equal to $\mathrm{IoU}(\cdot, \cdot)$ and $f(\mathrm{precision}(\hat{\mathcal{D}}_{c,j}), \bar{\mathrm{lq}}(\hat{\mathcal{D}}_{c,j}))=\bar{\mathrm{lq}}(\hat{\mathcal{D}}_{c,j})$ when $\mathrm{precision}(\hat{\mathcal{D}}_{c,j})=1$.
\end{theorem}
\begin{proof}[Just proof idea]

\end{proof}

While achieving a model that is both model-calibrated and recall-accurate seems to alleviate the necessity on postprocessing calibration, either of these objectives is neither achieved nor nontrivial.
Besides, $\bar{\mathrm{lq}}(\hat{\mathcal{D}}_{c,j})$ might be different from IoU as in the case of instance segmentation.
Accordingly, we introduce postprocessing calibration as an additional step to improve the calibration performance towards the needs of the downstream task.

Broadly speaking, postprocessing calibration involves a three step process:
\begin{itemize}
    \item Design an accuracy measure based on the requirement of the downstream task. This accuracy measure is also the thresholding criterion while cross-validating thresholds
    \item Design a specific case of GLaECE based on accuracy
\end{itemize}
In some cases, the accuracy measure may not be associated directly with GLaECE. What to do now?

\section{Post-hoc Calibration of Object Detectors}

\paragraph{Model Calibration}

\paragraph{Postprocessing Calibration}

\subsection{Model Calibration for NMS}
NMS is a critical component on most of the state-of-the-art object detectors by providing a mechanism to remove duplicate detections; thereby increasing the precision of the detector. While doing that NMS relies on the raw probabilities while prioritizing the detections and selecting the one that survives. As a result, for the sake of both (i) the efficiency of the NMS; and (ii) the accuracy of the surviving detections, it is critical that:
\begin{itemize}
    \item The background removal step passes a proper set of detections to the NMS such that NMS operation is not inefficient
    \item The raw probabilities encodes a useful information such that NMS operates with high accuracy
\end{itemize}

Previous works \cite{GFL,CorrLoss} show that the positive correlation of the raw probabilities and the localisation qualities of the detections is critical to achieve a better detection performance. Here, we take a step further and require the probabilities to represent the IoU of the prediction. Also by considering that NMS generally operates class-wise, we define model calibration as:
\begin{definition}[Model Calibration of Object Detectors]
An object detector is model-calibrated if the model probability of the detection ($\hat{p}_i^{\mathrm{model}}$) matches with the IoU between the detection box $\hat{b}_i$ and the ground truth box $b_{\psi(i)}$ where $\psi(i)$ is the ground truth box that   (Question: What should be the assignment - global like Hungarian or greedy like standard NMS): 
\begin{align}\label{eq:model_calibration}
    \mathbb{E}_{\hat{b}_i}[ \mathrm{IoU}(\hat{b}_i, b_{\psi(i)})] = \hat{p}_i^{\mathrm{model}}, \forall \hat{p}_i^{\mathrm{model}} \in [0,1]
\end{align}
\end{definition}

A model-calibrated detector has certain provable advantages that enables the efficiency of NMS without losing from the accuracy and abstracting away the model from the postprocessing. To demonstrate those, we investigate the background removal threshold $p^{\mathrm{post}}$ and the IoU threshold of NMS $\tau^{\mathrm{post}}$ as two critical hyper-parameters in the postprocessing.

\paragraph{The background removal threshold $p^{\mathrm{post}}$} Following theorem shows how to set $p^{\mathrm{post}}$ for a model-calibrated detector.

\begin{theorem}
Let us partition the raw detection set into three disjoint sets: set of TPs (IoU>0.50 and matched to a gt with a matching algorithm), set of duplicates (IoU > 0.50 but not matched) and set of FPs (IoU < 0.50), i.e., $\hat{\mathcal{D}}_{raw}=\hat{\mathcal{D}}_{TP} \cup \hat{\mathcal{D}}_{dup} \cup \hat{\mathcal{D}}_{FP}$. Then, if a detector is perfectly model-calibrated, setting background removal threshold to $\tau^{\mathrm{post}}$ (downstream task TP validation threshold) removes all FPs and keeps all potential TPs (TPs and duplicates), that is $\hat{\mathcal{D}}_{TP} \cup \hat{\mathcal{D}}_{dup}$. Thereby obtaining the smallest set of detections and enabling the most efficient NMS without a loss in the downstream task accuracy.
\end{theorem}
\begin{proof}[Just proof idea]
As perfectly calibrated then the scores correspond to the IoU and the detections with scores lower than $\tau^{\mathrm{down}}$ are simply false-positives. Removing the low-scoring false-positives does not decrease the performance. The remaining detections are then only TPs and duplicates, which will be then split by NMS.
\end{proof}

\paragraph{The IoU threshold of NMS $\tau^{\mathrm{post}}$}

As it is nontrivial to conduct an analysis for very different potential box combinations, we assume a recall-accurate detector and show that, even it is not trivial to output a perfectly accurate set of detections from such an accurate detector. Formally, we define the recall-accurate detector as follows:

\begin{definition}[A Recall-accurate detector]
We define a detector as recall-accurate if the raw detections include ground truths with perfect localisation. 
\end{definition}
In the following, we show that there are two main criteria for a recall-accurate detector to yield perfect detections:
\begin{theorem}
A recall-accurate detector achieves $AP=1$ for any $\tau > \tau^{\mathrm{post}}$ if it is model-calibrated and $\tau^{\mathrm{post}} >\mathrm{max}(\mathrm{IoU}(b_i, b_j))$ where $b_i$ and $b_j$ are the ground truth bounding boxes.
\end{theorem}
\begin{proof}[Just proof idea]
If calibrated, then scores represent IoUs. Also, if $\tau^{\mathrm{post}} >\mathrm{max}(\mathrm{IoU}(b_i, b_j))$, then NMS will not remove the perfect detection with score of $1.00$ corresponding to a different object. As a result, there will be a detection with a score of $1.00$ for each ground truth in the final detection set. Furthermore, there will not be any other detection with a score of $1.00$ as these lower-scoring detections were already removed by NMS as the model is calibrated.
\end{proof}

This theorem has two important implications, not investigated in the literature:
\begin{itemize}
    \item Dataset statistics have an important part in setting NMS threshold. 1.00 -> NMS does not operate; 0.00-> NMS removes correct predictions due to occlusion
    \item An improper IoU threshold results in removing perfect predictions for a recall-accurate detector.
    \item Class specific setting of IoU threshold based on dataset statistics can help model abstraction
\end{itemize}

\begin{figure}[t]
        \captionsetup[subfigure]{}
        \centering
        \begin{subfigure}[b]{0.6\textwidth}
        \includegraphics[width=\textwidth]{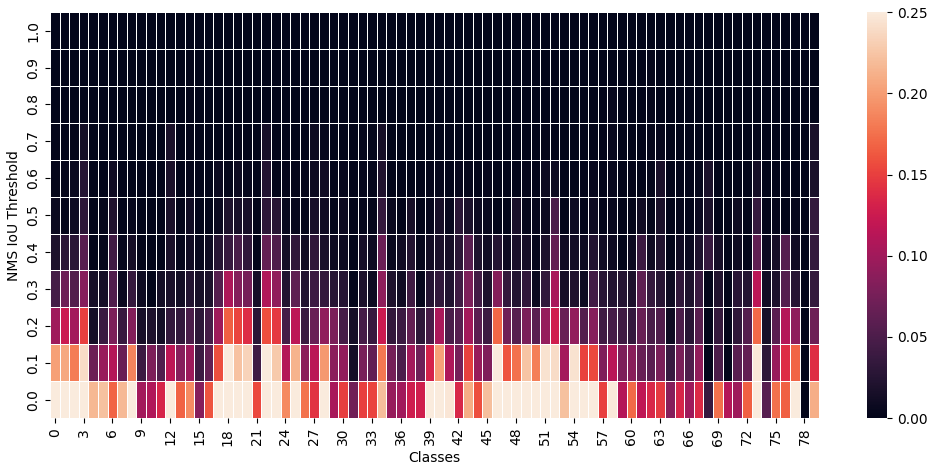}
        \caption{COCO val}
        \end{subfigure}
        \begin{subfigure}[b]{0.32\textwidth}
        \includegraphics[width=\textwidth]{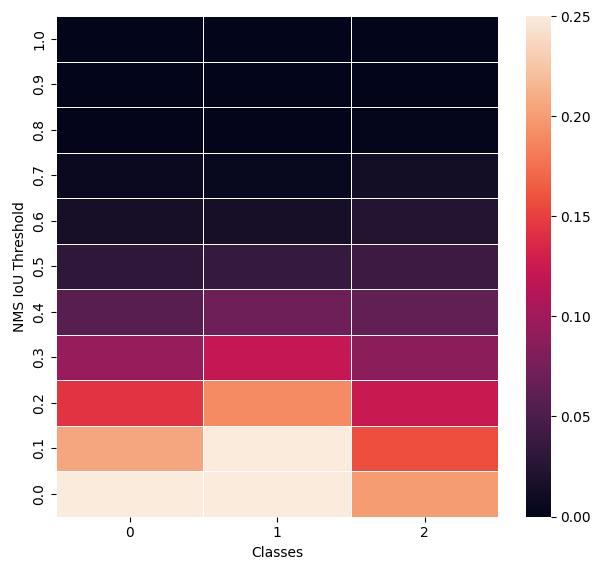}
        \caption{nuImages val}
        \end{subfigure}
        \caption{Recall Error of classes in a recall-accurate detector from different NMS IoU thresholds. Faster R-CNN 0.5-> 39.9, 0.6->40.0, 0.7->x.xx for COCO; 0.5-> 54.9, 0.6->55.5, 0.7->55.8, 0.8->55.2 for nuImages.}
        \label{fig:frcnnreliabilityhist}
        \vspace{-3ex}
\end{figure}

To demonstrate this, we make a simple analysis in which we assume that the raw detection set is exactly the ground truth boxes. We assign a confidence score of $1.00$ to these detections, making it perfectly model-calibrated. Then, we systematically increase $\tau^{\mathrm{post}}$ to see how the recall changes for each class and on average. We also plot $\mathrm{max}(\mathrm{IoU}(b_i, b_j))$. First, the trends of the classes are different and second even if the detector has perfect precision and localisation, the resulting AP is small when the threshold is not set properly.

\paragraph{Abstraction of Model from the Postprocessing} If a detector is model-calibrated, then $p^{\mathrm{post}}$ does not depend on the model at all. As for $\tau^{\mathrm{post}}$, the dataset and error statistics are both important. Assuming similar dataset statistics, calibration enables abstraction of the model from postprocessing.

}
\section{Experiments} \label{sec:exp}
In this section we seek to first outline: the criticality of calibrating individual object detectors when constructing an \gls{MoE}~(\cref{subsec:cal}); the effectiveness of \gls{CC} on many standard detection benchmarks, reaching \emph{state-of-the-art} on COCO, and DOTA~(\cref{subsec:benchmark}); followed by highlighting the reliability of \gls{CC} under domain shift~(\cref{subsec:reliability}); before finally outlining the limitations~(\cref{subsec:ablation}).
%
Our extensive experiments clearly show that \gls{CC} is consistently superior to the single models, Vanilla \gls{MoE}, and \glspl{DE} even while using fewer number of components in the mixture.

\begin{table}[t]
    \small
    \setlength{\tabcolsep}{0.4em}
    \centering
    \caption{Effect of calibration. All \glspl{MoE} use standard NMS.  
    \vspace{-2.5ex}}
    \scalebox{0.85}{
    \begin{tabular}{c|c|c|c|c||c|c|c} 
    \toprule
    \midrule
    Model&\multirow{2}{*}{Calibration}&\multicolumn{3}{c||}{\underline{Combined Detectors}}&\multicolumn{3}{c}{\underline{Detection Performance}} \\  
    Type&&RS R-CNN&ATSS&PAA&$\mathrm{AP}$&$\mathrm{AP_{50}}$&$\mathrm{AP_{75}}$ \\ \midrule
    Single&N/A&\cmark& & &$42.4$&$62.1$&$46.2$\\
    Models&N/A& &\cmark& &$43.1$&$61.5$&$47.1$\\
    &N/A& & &\cmark&$43.2$&$60.8$&$47.1$\\ \midrule 
    \multirow{8}{*}{\glspl{MoE}}&\xmark&\cmark&\cmark& &$42.4$&$62.1$&$46.3$\\
    &\cmark&\cmark&\cmark& &$\mathbf{44.1}$&$\mathbf{63.0}$&$\mathbf{48.4}$\\ 
    &\xmark&\cmark& &\cmark&$43.4$&$62.5$&$47.1$ \\
    &\cmark&\cmark& &\cmark&$\mathbf{44.0}$&$\mathbf{62.7}$&$\mathbf{47.9}$\\ 
    &\xmark& &\cmark&\cmark&$43.3$&$60.9$&$47.2$ \\
    &\cmark& &\cmark&\cmark&$\mathbf{44.4}$&$\mathbf{62.5}$&$\mathbf{48.5}$\\ 
    &\xmark&\cmark&\cmark&\cmark&$43.4$&$62.5$&$47.1$\\ 
    &\cmark&\cmark&\cmark&\cmark&$\mathbf{44.7}$&$\mathbf{63.1}$&$\mathbf{48.9}$\\ 
    \midrule
    \bottomrule
    \end{tabular}
     }
    \label{tab:ensembles_cal}
     \vspace{-1.5ex}
\end{table}

\subsection{Effect of Calibration on Obtaining \glspl{MoE}}
\label{subsec:cal}
To demonstrate the effect of calibration, we use the common COCO dataset \cite{COCO}.  
Similar to \cite{mvcalibrationod}, we randomly split COCO val set with 5K images into two, and use 2.5K images as COCO \textit{minival} to calibrate the detectors and keep the remaining 2.5K images  for testing as COCO \textit{minitest}. 
In our experiments, we mainly use COCO-style \gls{AP} and also report (i) $\mathrm{AP_{50}}$, $\mathrm{AP_{75}}$ as the \glspl{AP} measured at \gls{IoU} thresholds $0.50$ and $0.75$; as well as (ii)  $\mathrm{AP_{S}}$, $\mathrm{AP_{M}}$ and $\mathrm{AP_{L}}$ to present the accuracy on small, medium and large objects. 
In terms of models, here we combine RS R-CNN, ATSS and PAA with ResNet-50 \cite{ResNet} with FPN \cite{FeaturePyramidNetwork} backbone.
These detectors have different characteristics making them non-trivial to combine.
Specifically, RS R-CNN \cite{RSLoss} is a two-stage detector optimizing a ranking-based loss function, whereas ATSS \cite{ATSS} and PAA \cite{paa} are both one-stage detectors trained by focal loss \cite{FocalLoss}.
Also, different from ATSS with a centerness head, PAA employs an \gls{IoU} prediction head and they differ in obtaining the confidence score.

\begin{table*}
    \small
    \centering
    \caption{Object detection performance on COCO \textit{minitest}. \glspl{MoE} obtained by our \gls{CC} outperforms \gls{DE}s significantly even with less detectors. Our gains in green are obtained compared to the best single model for each performance measure, represented as underlined. \vspace{-2.5ex}}
    \label{tab:ensemble_moe}
    \scalebox{1.0}{
    \begin{tabular}{c|c||c|c|c||c|c|c} 
    \toprule
    \midrule
    Model Type&Detector&$\mathrm{AP}$&$\mathrm{AP_{50}}$&$\mathrm{AP_{75}}$&$\mathrm{AP_{S}}$&$\mathrm{AP_{M}}$&$\mathrm{AP_{L}}$\\ \midrule
    \multirow{3}{*}{Single Models}&RS R-CNN&$42.4$&$\underline{62.1}$&$46.2$&$26.8$&$46.3$&$56.9$\\
    &ATSS&$43.1$&$61.5$&$\underline{47.1}$&$\underline{27.8}$&$\underline{47.5}$&$54.2$\\ 
    &PAA&$\underline{43.2}$&$60.8$&$\underline{47.1}$&$27.0$&$47.0$&$\underline{57.6}$\\ \midrule
    \multirow{3}{*}{Deep Ensembles}
    %
    %
    %
    &RS R-CNN $\times$ 5&$43.4$&$63.0$&$47.7$&$28.0$&$47.5$&$57.0$\\ 
    &ATSS $\times$ 5&$44.1$&$62.3$&$48.4$&$29.4$&$49.0$&$56.3$\\ 
    &PAA $\times$ 5 &$44.4$&$62.0$&$48.4$&$28.9$&$49.0$&$59.2$\\  \midrule
    %
    \multirow{3}{*}{Mixtures of Experts}&Vanilla \gls{MoE} (RS R-CNN, ATSS, PAA)&$43.4$&$62.5$&$47.1$&$27.3$&$47.3$&$58.0$\\ \cdashlinelr{2-8}
    &\gls{CC} (ATSS and PAA) - Ours&$44.8$&$62.4$&$49.2$&$29.4$&$49.1$&$57.6$\\
    &\gls{CC} (RS R-CNN, ATSS, PAA) - Ours&$\textbf{45.5}$&$\textbf{63.2}$&$\textbf{50.0}$&$\textbf{29.7}$&$\textbf{49.7}$&$\textbf{59.3}$\\
    &&$\imp{2.3}$&$\imp{1.1}$&$\imp{2.9}$&$\imp{1.9}$&$\imp{2.2}$&$\imp{1.7}$\\
    \midrule
    \bottomrule
    \end{tabular}
     }
     \vspace{-1.5ex}
\end{table*}

%

%
%

\paragraph{Calibration is crucial for accurate \glspl{MoE}}
In order to highlight the effect of calibration in different settings, we construct three \glspl{MoE} from pair-wise combinations of RS R-CNN, ATSS and PAA as well as one \gls{MoE} that combines all three.
In order to focus only on calibration, here we use the standard \gls{NMS} with an IoU threshold of $0.65$ as in \cite{yolov7}.
\cref{tab:ensembles_cal} presents the results of \glspl{MoE} with uncalibrated and calibrated detectors.
The striking observation is that \textit{without calibration, the \glspl{MoE} perform similar to the single models and calibration enables accurate \glspl{MoE} for all four settings.}
Specifically, using two calibrated \glspl{MoE} yields $\sim 1$AP gain, and using three improves AP by $1.5$ compared to single models; showing the effectiveness of calibration.

\paragraph{Even with fewer models, \gls{CC} is superior to \glspl{DE}} 
Next we compare \gls{CC}, Vanilla \gls{MoE} and \glspl{DE}.
%
%
We calibrate the components of \glspl{DE} while combining them, though we do not observe a significant effect from the calibration in their performance (see App. \ref{app:exp} for details); which is an expected outcome.  
%
%
\cref{tab:ensemble_moe} shows that the \glspl{DE} perform consistently better than the single models; validating them as strong baselines.
The main observation in \cref{tab:ensemble_moe} is that \textit{combining different types of few detectors into an \gls{MoE} performs significantly better than \glspl{DE}.}
Specifically, \gls{CC} with only two detectors, ATSS and PAA, outperforms all \glspl{DE}, each with five components.
Also, combining three detectors by \gls{CC} performs $1.1$ AP better than its closest counterpart \gls{DE}.
This is because the same type of detectors make similar errors, which yields less gain once they are combined together.
However, different types of detectors complement each other thanks to their diversity.
%
Finally, \gls{CC} outperforms Vanilla \gls{MoE} by $\sim 2$ AP in this setting.
Our final model obtains $45.5$ AP and outperforms the best single model in all \gls{AP} variants significantly.

\subsection{Benchmarking \gls{CC} on Various Tasks}\label{subsec:benchmark}
\label{subsec:objdet}
In this section, we demonstrate that combining off-the-shelf detectors via \gls{CC} improves single detectors on various detection tasks up to $2.5$ AP, which is a significant performance improvement.
Our  \gls{CC} reaches \gls{SOTA} results on COCO dataset among public models and on DOTA dataset for rotated object detection.
Specifically, we evaluate on four different tasks: object detection (COCO~\cite{COCO}), rotated object detection (DOTA~\cite{dota}), open vocabulary object detection (COCO and ODinW35~\cite{li2022elevater}) and instance segmentation (LVIS~\cite{LVIS}).
For these tasks, we use a total of 15 different detectors include one-stage and two-stage, convolutional, transformer-based ones and foundation models.

\begin{table}[t]
\small
\setlength{\tabcolsep}{0.7em}
\centering
\caption{Detection performance on COCO \textit{test-dev} using strong detectors. \textcolor{forestgreen}{Green}: Gain against best single model (\underline{underlined}).    
\vspace{-2.5ex}
}
\label{tab:testdev}
\scalebox{0.85}{
\begin{tabular}{c||c|c|c||c|c|c} 
\toprule
\midrule
Method&$\mathrm{AP}$&$\mathrm{AP_{50}}$&$\mathrm{AP_{75}}$&$\mathrm{AP_{S}}$&$\mathrm{AP_{M}}$&$\mathrm{AP_{L}}$ \\ \midrule
YOLOv7 \cite{yolov7}&$55.5$&$73.0$&$60.6$&$37.9$&$58.8$&$67.7$\\
QueryInst \cite{queryinst}&$55.7$&$\underline{75.7}$&$61.4$&$36.2$&$58.4$&$\underline{70.9}$\\
DyHead \cite{dyhead}&$\underline{56.6}$&$75.5$&$\underline{61.8}$&$\underline{39.4}$&$\underline{59.8}$&$68.7$\\ \midrule 
Vanilla \gls{MoE} &$57.6$&$76.6$&$63.2$&$40.0$&$60.9$&$70.8$\\
&$\imp{1.0}$&$\imp{0.9}$&$\imp{1.4}$&$\imp{0.6}$&$\imp{1.1}$ &$\nimp{0.1}$\\ \midrule 
\gls{CC} &$\mathbf{59.0}$&$\mathbf{77.2}$&$\mathbf{64.7}$&$\mathbf{41.1}$&$\mathbf{62.6}$&$\mathbf{72.4}$\\
(Ours)&$\imp{2.4}$&$\imp{1.5}$&$\imp{2.9}$&$\imp{1.7}$&$\imp{2.8}$&$\imp{1.5}$ \\ \midrule
\bottomrule
\end{tabular}
}
\vspace{-1.5ex}
\end{table}




\paragraph{Object Detection on COCO} 
%
Here, we evaluate on COCO \textit{test-dev} by submitting our result to the evaluation server.
In the first setting, we combine the following well-known and effective detectors:
\begin{compactitem}
\item YOLOv7 \cite{yolov7} with a large convolutional backbone following its original setting,
\item QueryInst \cite{queryinst} as a transformer-based detector with a Swin-L \cite{swin} backbone,
\item ATSS with transformer-based dynamic head \cite{dyhead} and again Swin-L backbone.
\end{compactitem}
These detectors differ from each other in terms of the pretraining data, backbone or architecture as summarized in App. \ref{app:exp}. 
\cref{tab:testdev} shows that our \gls{CC} reaches $59.0$ AP with a gain of $2.4$ AP on this challenging setting as well.
As our gain here is similar to that of \cref{tab:ensemble_moe}, we can easily say that the gain of our \gls{CC} has not saturated in this stronger setting, which is commonly the opposite in the literature.

Finally, we also evaluate \gls{CC} on the two most recent \gls{SOTA} publicly available detectors\footnote{Published at CVPR 2023 and ICCV 2023}:
\begin{compactitem}
    \item EVA \cite{EVA}, a foundation model for vision using Cascade Mask R-CNN \cite{CascadeRCNN} for detection,
    \item Co-DETR \cite{codetr}, a transformer-based detector.
\end{compactitem}
\textit{\cref{tab:found_model_testdev} shows that \gls{CC} reaches \gls{SOTA} with $65.1$ AP on COCO test-dev and outperforms all existing public detectors by $0.7$ AP.}
This further shows the effectiveness of \gls{CC}.

%
%
%
%
%
%
%
%

\begin{table}[t]
\small
\setlength{\tabcolsep}{0.7em}
\centering
\caption{Object detection performance on COCO \textit{test-dev} using SOTA detectors.  \gls{CC} improves the most accurate publicly available model by $0.7$ AP and reaches \gls{SOTA}.  
}
\label{tab:found_model_testdev}
\scalebox{0.85}{
\begin{tabular}{c||c|c|c||c|c|c} 
\toprule
\midrule
Method&$\mathrm{AP}$&$\mathrm{AP_{50}}$&$\mathrm{AP_{75}}$&$\mathrm{AP_{S}}$&$\mathrm{AP_{M}}$&$\mathrm{AP_{L}}$ \\ \midrule
EVA \cite{EVA}&$\underline{64.4}$&$\underline{82.3}$&$70.9$&$\underline{48.2}$&$\underline{67.6}$&$\underline{77.5}$\\
Co-DETR \cite{codetr}&$64.3$&$81.4$&$\underline{71.0}$&$48.1$&$67.1$&$\underline{77.5}$\\
\midrule 
Vanilla \gls{MoE} &$64.6$&$\mathbf{82.7}$&$71.0$&$48.5$&$67.6$&$77.5$\\
&$\imp{0.2}$&$\imp{0.4}$&$\textcolor{blue}{0.0}$&$\imp{0.3}$&$\textcolor{blue}{0.0}$ &$\textcolor{blue}{0.0}$\\ \midrule 
\gls{CC} &$\mathbf{65.1}$&$\mathbf{82.7}$&$\mathbf{71.9}$&$\mathbf{49.2}$&$\mathbf{68.1}$&$\mathbf{78.1}$ \\ 
(Ours)&$\imp{0.7}$&$\imp{0.4}$&$\imp{0.9}$&$\imp{1.0}$&$\imp{0.5}$&$\imp{0.6}$ \\ 
\midrule
\bottomrule
\end{tabular}
}
\vspace{-1.5ex}
\end{table}

\begin{table}[t]
    \setlength{\tabcolsep}{0.4em}
    \centering
    \small
    \caption{Rotated object detection performance on DOTA test set. $\mathrm{AP_{50}}$ is used following DOTA. See App. \ref{app:exp} for all classes.  
    \vspace{-1.5ex}
    }
    \label{tab:rotbb}
    \scalebox{0.85}{
    \begin{tabular}{c||c||c|c|c|c|c} 
    \toprule
    \midrule
    \multirow{2}{*}{Method}&\multirow{2}{*}{$\mathrm{AP_{50}}$}&\multicolumn{5}{c}{\underline{5 Classes with Lowest Performance}} \\
    & &Bridge&Soccer&Roundab.&Harbor&Helico.\\ \midrule
    RTMDet \cite{rtmdet}&$81.32$&$58.50$&$\underline{72.12}$&$70.85$&$\underline{81.16}$&$77.24$\\
    LSKN \cite{lskn}&$\underline{81.85}$&$\underline{61.47}$&$71.67$&$\underline{71.35}$&$79.19$&$\underline{80.85}$\\ \midrule 
    Vanilla \gls{MoE}&$80.60$&$\mathbf{61.77}$&$70.98$&$65.92$&$84.28$&$77.57$ \\
    &$\nimp{1.25}$&$\imp{0.30}$&$\nimp{1.14}$&$\nimp{5.43}$&$\imp{3.12}$&$\nimp{3.28}$\\ \midrule
    \gls{CC}&$\mathbf{82.62}$&$61.38$&$\mathbf{75.50}$&$\mathbf{74.12}$&$\mathbf{84.49}$&$\mathbf{81.93}$\\
    (Ours)&$\imp{0.77}$&$\nimp{0.09}$&$\imp{3.38}$&$\imp{2.77}$&$\imp{3.33}$&$\imp{1.08}$\\ \midrule
    \bottomrule
    \end{tabular}
    }
\vspace{-1.5ex}
\end{table}

\paragraph{Rotated Object Detection on DOTA} 
We now investigate \gls{CC} for rotated object detection on DOTA v1.0 dataset \cite{dota} with 15 classes.
DOTA is also a challenging dataset comprising of aerial images that are very dense in terms of objects.
Specifically, DOTA dataset has $67.1$ objects on average per image.
We use all 458 images in the validation set to calibrate the detectors and report $\mathrm{AP_{50}}$ on the test set by submitting our results to the evaluation server.
We combine LSKN \cite{lskn} and RTMDet \cite{rtmdet} as two recent \gls{SOTA} detectors.
Following the literature, we use \gls{NMS} with an IoU threshold of $0.35$ as Soft \gls{NMS} and Score Voting are not straightforward to use in  this task.
\cref{tab:rotbb} suggests that \textit{we establish a new \gls{SOTA}} with $82.62$ $\mathrm{AP_{50}}$ on DOTA; improving the previous \gls{SOTA} by $0.77$.
Having examined the classes, we note that our improvement originates mostly from the classes with relatively lower performance; with the exception of the class `bridge' which performs marginally worse.
For example, on `soccer-field', `roundabout', `harbor' classes where the single detectors have between $70-80$ $\mathrm{AP_{50}}$, the improvement is around $3$ $\mathrm{AP_{50}}$.
These gains enable us to demonstrate the ability of \gls{CC}  to set a new \gls{SOTA} in rotated object detection.

\begin{table}[t]
\small
\setlength{\tabcolsep}{0.42em}
\centering
\caption{Comparison on open vocabulary object detection. \textcolor{forestgreen}{Green}: Gain against best single model (\underline{underlined}). 
\vspace{-2.5ex}
}
\label{tab:open_vocab}
\scalebox{0.85}{
\begin{tabular}{c||c|c|c||c|c} 
\toprule
\midrule
Method&\multicolumn{3}{c||}{COCO \cite{COCO}}&\multicolumn{2}{c}{ODinW-35 \cite{li2022elevater}}\\
&$\mathrm{AP}$&$\mathrm{AP_{50}}$&$\mathrm{AP_{75}}$&$\mathrm{AP_{average}}$&$\mathrm{AP_{median}}$\\
\midrule
Grounding DINO-T \cite{groundingdino}& $\underline{48.5}$ & $\underline{64.5}$ & $\underline{52.9}$ & $\underline{22.7}$ & $\underline{13.8}$ \\
MQ-GLIP-T \cite{mqdet}               & $46.3$ & $62.7$ & $50.6$ & $21.4$ & $8.2$  \\

\midrule

Vanilla \gls{MoE}                    & $48.1$ & $64.8$ & $52.5$ & $22.4$ & $11.8$ \\
                                     &$\nimp{0.4}$  & $\imp{0.3}$ &  $\nimp{0.4}$   & $\nimp{0.3}$  & $\nimp{2.0}$\\ 

\midrule 

\gls{CC}                             &$\mathbf{49.5}$       & $\mathbf{66.4}$        & $\mathbf{54.3}$ & $\mathbf{23.2}$ & $\mathbf{14.9}$\\
(Ours)                               &$\imp{1.0}$  & $\imp{2.1}$ &  $\imp{1.4}$   & $\imp{0.5}$  & $\imp{1.1}$\\ 
\midrule
\bottomrule
\end{tabular}
}
\vspace{-1.5ex}
\end{table}

\paragraph{Open Vocabulary Object Detection (OVOD)} We now investigate the effect of \gls{CC} on \gls{OVOD} task \cite{glip,li2022elevater}.
\gls{OVOD} task is a recently proposed challenging task in which the aim is to detect the objects pertaining to the classes in a given text prompt.
This requires the models to be able to interpret the given text prompt as well as the image and yield the detection results.
Furthermore, the text prompt can contain phrases that are not necessarily in the training data.
For this challenging task, we combine two strong and recent models: (i) Grounding DINO \cite{groundingdino}, a transformer-based detector and MQ-Det \cite{mqdet}, an anchor-based detector relying on GLIP \cite{glip}.
We obtain the calibrators on a subset of the Objects365 dataset \cite{Objects365}, which is included in the pretraining data for both of these models.
Therefore, the calibrators are also limited to the pretraining data only.
In our evaluation, we evaluate the models on COCO and ODinW-35\cite{li2022elevater} datasets following the common convention \cite{glip,groundingdino,mqdet}.
Note that ODinW-35 is a challenging dataset with 35 different subdatasets, some of which are substantially different from the pretraining dataset.
To illustrate, ODinW-35 includes subdatasets specifically for `potholes` on the road and \emph{infrared} images of `dogs` and `people`.
\cref{tab:open_vocab} shows that \gls{CC} improves the single models on both of these datasets on all performance measures notably.
As an example, the median AP of the best single model on ODinW-35 increases from $13.8$ to $14.9$, which suggests an $8 \%$ relative gain.
%

\begin{table}[t]
\small
\setlength{\tabcolsep}{0.35em}
\centering
\caption{Instance segmentation performance on LVIS val set. $\mathrm{AP_{box}}$ represents detection \gls{AP}. \vspace{-2.5ex}}
\label{tab:lvis}
\scalebox{0.85}{
\begin{tabular}{c||c|c|c||c|c|c||c} 
\toprule
\midrule
Method&$\mathrm{AP}$&$\mathrm{AP_{50}}$&$\mathrm{AP_{75}}$&$\mathrm{AP_{r}}$&$\mathrm{AP_{c}}$&$\mathrm{AP_{f}}$&$\mathrm{AP_{box}}$ \\ \midrule
Mask R-CNN \cite{MaskRCNN}&$\underline{25.4}$&$39.2$&$\underline{27.3}$&$15.7$&$\underline{24.7}$&$\underline{30.4}$&$\underline{26.6}$\\
RS Loss \cite{RSLoss}&$25.1$&$38.2$&$26.8$&$\underline{16.5}$&$24.3$&$29.9$&$25.8$\\
Seesaw Loss \cite{seesawloss}&$\underline{25.4}$&$\underline{39.5}$&$26.9$&$15.8$&$\underline{24.7}$&$\underline{30.4}$&$25.6$\\
\midrule 
Vanilla \gls{MoE}&$25.2$&$38.3$&$26.8$&$16.5$&$24.3$&$29.9$&$25.9$ \\
&$\nimp{0.2}$&$\nimp{1.2}$&$\nimp{0.5}$&$\textcolor{blue}{0.0}$&$\nimp{0.4}$&$\nimp{0.5}$&$\nimp{0.7}$ \\
\midrule
\gls{CC}&$\mathbf{27.7}$&$\mathbf{42.8}$&$\mathbf{29.4}$&$\mathbf{18.2}$&$\mathbf{27.3}$&$\mathbf{32.4}$&$\mathbf{29.1}$\\
(Ours)&$\imp{2.3}$&$\imp{3.3}$&$\imp{2.1}$&$\imp{1.7}$&$\imp{2.4}$&$\imp{2.0}$&$\imp{2.5}$ \\
\bottomrule  \midrule
\end{tabular}
}
\vspace{-1.5ex}
\end{table}


\label{subsec:ins}
\paragraph{Instance Segmentation} 
Given that \gls{CC} is beneficial for object detection, one would expect it to improve performance on the instance segmentation task.
To verify this, we use LVIS \cite{LVIS} as a long-tailed dataset for instance segmentation with more than 1K classes. 
Following its standard evaluation, we also report the AP on rare ($\mathrm{AP_{r}}$), common ($\mathrm{AP_{c}}$) and frequent ($\mathrm{AP_{f}}$) classes.
Similar to COCO, we reserve 500 images from val set to calibrate the detectors, and test our models on the remaining 19.5K images of the val set.
We combine three recent and diverse off-the-shelf Mask R-CNN variants in a \gls{MoE}: 
\begin{compactitem} 
    \item The vanilla Mask R-CNN \cite{MaskRCNN} with ResNeXt-101 \cite{ResNext} backbone, softmax classifier and using Repeat Factor Sampling (RFS) used for the long-tailed nature of LVIS,
    \item Mask R-CNN with ResNet-50, sigmoid classifier, trained with RS Loss \cite{RSLoss} and RFS,
    \item Mask R-CNN with ResNet-50, softmax classifier, trained with Seesaw Loss\cite{seesawloss} but no RFS.
\end{compactitem}

%
%
%

\cref{tab:lvis} shows that \textit{while the Vanilla \gls{MoE} performs worse than the best single model, \gls{CC} boosts the segmentation \gls{AP} by $2.3$}, an improvement of $\sim 10\%$ over the best single model.
Also, the detection \gls{AP} ($\mathrm{AP_{box}}$) improves by $2.5$ aligned with our previous findings.
%

%
%
%


\subsection{How Reliable is \gls{CC}?}\label{subsec:reliability}
As we are ensembling detectors in the form of \gls{MoE}, it is naturally to evaluate how reliable \gls{CC} is.
To evaluate this, we investigate how \gls{CC} performs under domain shift  \cite{RobustnessODBenchmark, hendrycks2019robustness}, and also test \gls{CC} on the recently proposed \gls{SAOD} task \cite{saod}.
%
Here, we use our setting in \cref{subsec:cal}, in which we combine RS R-CNN, ATSS and PAA.
That is, we use the calibrators trained on clean COCO and do not train a new calibrator.

\paragraph{Domain Shift (Synthetic and Natural)} Following the convention \cite{RobustnessODBenchmark,saod,calibrationdomainshiftod}, we apply 15 ImageNet-C style corruptions \cite{hendrycks2019robustness} under 5 different severities for synthetic domain shift on COCO.
Similar to the clean data, we observe in \cref{tab:corr} that combining only three models (RS R-CNN, ATSS and PAA) outperforms \glspl{DE} with five components thanks to the diversity of the detectors.
Here we see, that the performance over the best single model improves by $1.5$ AP.
Next, we evaluate the performance of \gls{CC} on Objects45K \cite{saod}.
Please note that this dataset has the same set of classes as in COCO, but collected and annotated separately, thereby implying a natural domain shift and a similar setting was used in \cite{RegressionUncOD}.
\cref{tab:natural_domain_shift} shows that the \glspl{DE} do not provide notable gains in this setting.
For example, PAA $\times$ 5 only improves the single PAA only by $0.7$ AP.
However, \gls{CC} improves the best single model by $\sim2$ AP, outperforming all \glspl{DE} and Vanilla MoE.

\paragraph{Self-aware Object Detection (SAOD)} Finally, we evaluate \gls{CC} on the recently proposed \gls{SAOD} task \cite{saod}, which requires the object detectors to be self-aware.
This task requires detectors to provide reliable uncertainty estimates along with accurate and calibrated detections in a holistic manner; which is evaluated using the \gls{DAQ}.
To evaluate the models on this task, we convert RS R-CNN, ATSS, PAA, Vanilla \gls{MoE} and \gls{CC} to a self-aware detector following \cite{saod}.
\gls{CC} improves the \gls{DAQ} of the best single model from $40.9$ to $42.9$ and outperforms Vanilla \gls{MoE} by $\sim 0.5$ DAQ.
Overall, this suggest that \gls{CC} is more reliable than the single detectors.
%
%
Details are provided in App. \ref{app:exp} due to space limitation.

\begin{table}
    \small
    \setlength{\tabcolsep}{0.5em}
    \centering
    \caption{Comparison on COCO \textit{mini-test} with ImageNet-C style corruptions. \textcolor{forestgreen}{Green}: Gain against best single model (\underline{underlined}). 
    \vspace{-2.5ex}
    }
    \label{tab:corr}
    \scalebox{0.85}{
    \begin{tabular}{c|c||c|c|c|c|c||c} 
    \toprule
    \midrule
    Model&\multirow{2}{*}{Detector}&\multicolumn{5}{c}{{Severity of the Corruption}}&mean\\
    Type&&$\mathrm{1}$&$\mathrm{2}$&$\mathrm{3}$&$\mathrm{4}$&$\mathrm{5}$ &$\mathrm{AP}$\\ \midrule
    Single&RS R-CNN&$33.2$&$27.7$&$21.9$&$16.2$&$11.8$&$22.2$\\
    Models&ATSS&$33.8$&$28.2$&$22.2$&$16.3$&$11.8$&$22.5$\\ 
    &PAA&$\underline{34.4}$&$\underline{29.2}$&$\underline{23.1}$&$\underline{16.9}$&$\underline{12.1}$&$\underline{23.1}$\\ \midrule
    \multirow{3}{*}{\glspl{DE}}
    &RS R-CNN $\times$ 5&$34.6$&$29.3$&$23.5$&$17.5$&$12.8$&$23.5$\\ 
    &ATSS $\times$ 5&$35.0$&$29.5$&$23.4$&$17.4$&$12.7$&$23.6$\\
    &PAA $\times$ 5&$35.7$&$30.4$&$24.4$&$\mathbf{18.2}$&$\mathbf{13.2}$&$24.4$\\
    \midrule
    \multirow{3}{*}{\glspl{MoE}}
    &Vanilla \gls{MoE}&$34.5$&$29.1$&$23.1$&$17.2$&$12.5$&$23.3$\\
    %
    \cdashlinelr{2-8}
    &\gls{CC}&$\mathbf{36.3}$&$\mathbf{30.8}$&$\mathbf{24.6}$&$\mathbf{18.2}$&$\mathbf{13.2}$&$\mathbf{24.6}$\\ &(Ours)&$\imp{1.9}$&$\imp{1.6}$&$\imp{1.5}$&$\imp{1.3}$&$\imp{1.1}$&$\imp{1.5}$\\
    \midrule
    \bottomrule
    \end{tabular}
     }
     \vspace{-1.5ex}
\end{table}

\begin{table}[t]
\small
\setlength{\tabcolsep}{0.9em}
\centering
\caption{Results on Object45K, a natural domain shift from COCO. \textcolor{forestgreen}{Green}: Gain against best single model (\underline{underlined}). 
\vspace{-1.5ex}
}
\label{tab:natural_domain_shift}
\scalebox{0.85}{
\begin{tabular}{c|c||c|c|c} 
\toprule
\midrule
Model Type&Detector&$\mathrm{AP}$&$\mathrm{AP_{50}}$&$\mathrm{AP_{75}}$\\
\midrule
\multirow{3}{*}{Single Models}
&RS-RCNN&$28.6$&$\underline{41.2}$&$\underline{31.2}$\\
&ATSS&$\underline{28.7}$&$39.8$&$31.2$\\
&PAA&$\underline{28.7}$&$39.5$&$31.0$\\
\midrule

\multirow{3}{*}{Deep Ensembles}
%
%
%
%
&RS-RCNN $\times$ 5&$29.8$&$\mathbf{42.6}$&$32.7$\\
&ATSS $\times$ 5 &$29.3$&$40.4$&$31.8$\\
&PAA $\times$ 5 &$29.4$&$40.1$&$31.8$\\
%
%
%
\midrule 
\multirow{3}{*}{Mixture of Experts}
&Vanilla \gls{MoE} &$29.3$&$41.4$&$31.8$\\
\cdashlinelr{2-5} 
&\gls{CC} &$\mathbf{30.6}$&$41.6$&$\mathbf{33.3}$ \\
&(Ours)&$\imp{1.9}$&$\imp{0.4}$&$\imp{2.1}$\\ 
\midrule
\bottomrule
\end{tabular}
}
\vspace{-1.5ex}
\end{table}

\subsection{Ablation Analysis and Limitations}
\label{subsec:ablation}

    \begin{table}[t]
    \centering
    \caption{Ablation analysis of \gls{CC}. We use the \gls{MoE} combining RS R-CNN, ATSS and PAA on COCO \textit{mini-test}. 
    }
    \label{tab:ablation}
    \scalebox{0.85}{
    \begin{tabular}{c|c|c||c} 
    \toprule
    \midrule
    \multirow{2}{*}{Calibration}&\multicolumn{2}{c||}{Refining NMS}&\multirow{2}{*}{$\mathrm{AP}$}\\
    &Soft NMS&Score Voting&\\ \midrule
    \xmark&\xmark&\xmark&$43.3$\\
    \cmark&\xmark&\xmark&$44.7$\\
    \cmark&\cmark&\xmark&$44.8$\\
    \xmark&\cmark&\xmark&$43.4$\\
    \xmark&\cmark&\cmark&$44.4$\\
    \cmark&\cmark&\cmark&$\mathbf{45.5}$\\
    \midrule
    \bottomrule
    \end{tabular}
     }
\end{table}

%
\paragraph{Ablation Analysis} Here, we provide further ablation of \gls{CC} in \cref{tab:ablation} in which the calibration appears to be the major factor of the performance gain.
%
%
Soft NMS yields a small gain of $0.1$ \gls{AP} and Score Voting, combining boxes from different detectors to extract a new bounding box, improves \gls{AP} notably from $44.8$ to $45.5$.
Please refer to App.\ref{app:exp} for further details.

\paragraph{Limitations}
In cases where the performance gap is large between the experts, we observe that both \gls{CC} and Vanilla \gls{MoE} tend to perform worse than the best single model.
For example, combining RS R-CNN, ATSS, PAA (APs $\approx$ 40) with EVA and Co-DETR (APs $\approx$ 65) does not result in a stronger \gls{MoE} than EVA or Co-DETR (refer to App. \ref{app:exp} for the details).
However, once we use an Oracle \gls{MoE}, which is obtained by directly assigning the calibration targets as the confidence, \gls{CC} reaches $86.7$AP achieving more than $20$ AP improvement compared to the best single model; an observation which aligns with our theoretical insights.
This suggests that calibration is critical to obtain effective \glspl{MoE} and that more effort is required from the community in calibrating object detectors.
%

\blockcomment{
\textbf{Validating the Calibrator} 
We first justify why we prefer Class-agnostic (CA) \gls{IR} calibrator in \gls{CC} .
We would ideally expect a calibrator to improve the calibration also by preserving the accuracy of the detector. 
To see that, we investigate \gls{LR} and \gls{IR} both \gls{CA} and \gls{CW} on late as well as early calibration in \cref{tab:ensemble_accuracy}.
We observe in the red cells in \cref{tab:ensemble_accuracy} that all calibrators, except \gls{CA} \gls{IR}, decrease \gls{AP} especially for early calibration.
Furthermore, \gls{CA} \gls{IR} improves \gls{LaECE} in all cases.
We also discuss in App. \ref{app:exp} that \gls{CA} \gls{IR} improves average and maximum calibration errors; thereby providing a better gain for \glspl{MoE} compared to other calibrators.
These observations on accuracy and calibration led us to choose \gls{CA} \gls{IR} while calibrating the single models in \gls{CC}.

\begin{table}[t]
    \setlength{\tabcolsep}{0.15em}
    \centering
    \caption{\gls{AP} and \gls{LaECE} of uncalibrated, early and late calibrated models on COCO \textit{mini-test}. Calibrator is N/A for baseline. \textcolor{bubblegum}{Red}: A notable \gls{AP} drop, \textcolor{inchworm}{green}: consistent \gls{AP}, \textbf{bold}: best \gls{LaECE}, \underline{underlined}: second in \gls{LaECE}. 
    \vspace{-1.ex}
    }
    \label{tab:ensemble_accuracy}
    \scalebox{0.85}{
    \begin{tabular}{c|c||c|c|c||c|c|c} 
    \toprule
    \midrule
    Cal.&Calibrator& \multicolumn{3}{c||}{AP}&\multicolumn{3}{c}{\gls{LaECE}} \\ 
    Type&&RS R-CNN&ATSS&PAA&RS R-CNN&ATSS&PAA\\ \midrule
    \multirow{5}{*}{Early}&N/A&$42.4$&$43.1$&$43.2$&$13.79$&$0.20$&$3.39$\\ \cline{2-8}
    &CW LR&\cellcolor{bubblegum} $26.4$&\cellcolor{bubblegum} $42.0$&\cellcolor{bubblegum}$37.1$&$0.44$&$0.12$&$\underline{0.22}$\\ 
    &CW IR&\cellcolor{bubblegum} $41.9$&\cellcolor{bubblegum} $42.8$&\cellcolor{bubblegum} $42.5$&$\mathbf{0.03}$&$\mathbf{0.02}$&$1.39$\\ 
    &CA LR&\cellcolor{inchworm}$42.4$&\cellcolor{bubblegum}$42.8$&\cellcolor{inchworm}$43.2$&$0.65$&$0.17$&$0.37$\\
    &CA IR&\cellcolor{inchworm}$42.4$&\cellcolor{inchworm}$43.1$&\cellcolor{inchworm}$43.2$&$\underline{0.14}$&$\underline{0.10}$&$\mathbf{0.14}$\\ \midrule \midrule
    \multirow{5}{*}{Late}&N/A&$42.4$&$43.1$&$43.2$&$36.45$&$5.01$&$11.23$\\ \cline{2-8}
    &CW LR&\cellcolor{inchworm}$42.4$&\cellcolor{inchworm}$43.1$&\cellcolor{inchworm}$43.2$&$4.36$&$\underline{2.69}$&$\underline{1.39}$\\ 
    &CW IR&\cellcolor{bubblegum}$41.8$&\cellcolor{bubblegum}$42.5$&\cellcolor{bubblegum}$42.4$&$\mathbf{1.56}$&$\mathbf{2.35}$&$\mathbf{1.21}$\\ 
    &CA LR&\cellcolor{inchworm}$42.4$&\cellcolor{inchworm}$43.0$&\cellcolor{inchworm}$43.2$&$5.83$&$4.46$&$1.63$\\ 
    &CA IR&\cellcolor{inchworm}$42.3$&\cellcolor{inchworm}$43.1$&\cellcolor{inchworm}$43.2$&$\underline{3.15}$&$4.51$&$1.62$\\ 
    \midrule
    \bottomrule
    \end{tabular}
    }
\end{table}
}
\blockcomment{
\begin{figure*}[t]
        \captionsetup[subfigure]{}
        \centering
        \begin{subfigure}[b]{0.4\textwidth}
            \includegraphics[width=\textwidth]{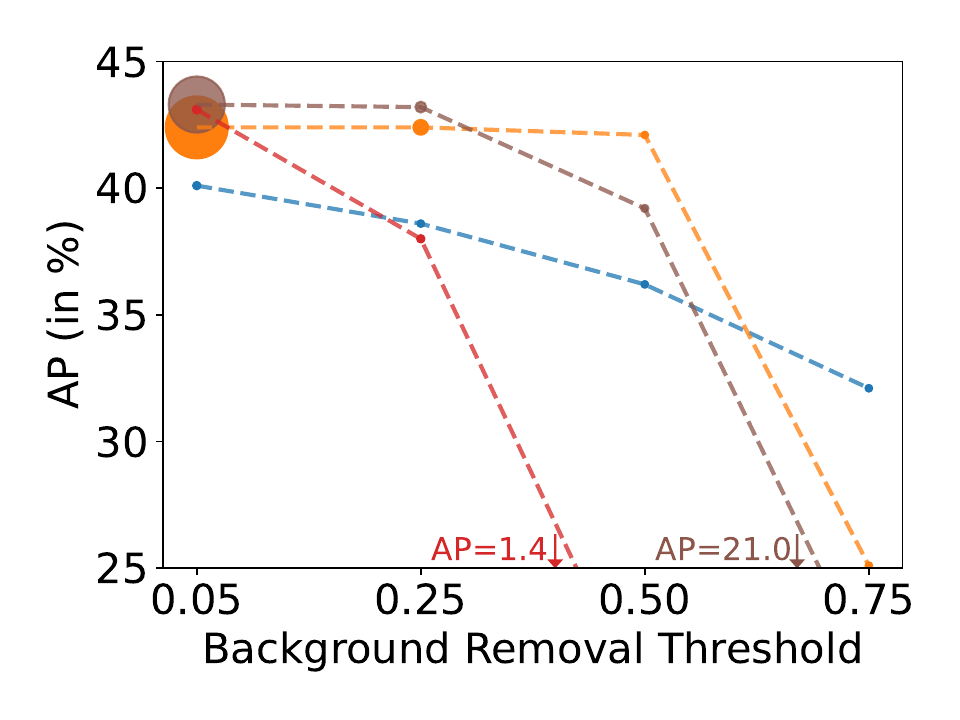}
            \caption{Uncalibrated}
        \end{subfigure}
        \begin{subfigure}[b]{0.4\textwidth}
            \includegraphics[width=\textwidth]{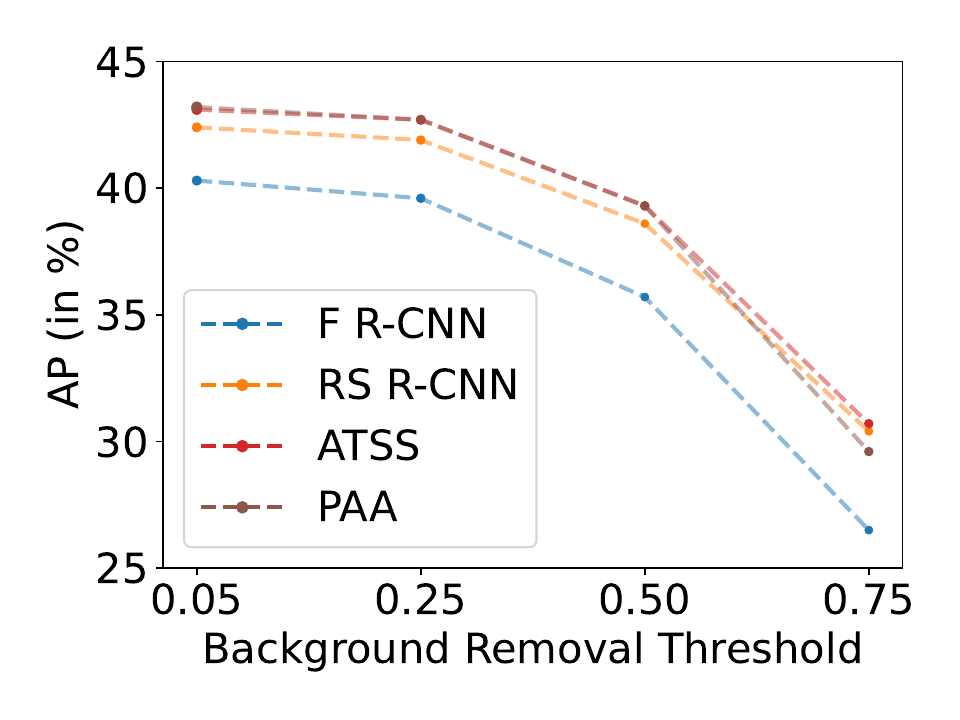}
            \caption{Early Calibrated}
        \end{subfigure}
        %
        \vspace{-2.ex}
        \caption{The effect of background removal threshold on \gls{AP} and \gls{NMS} processing time for \textbf{(a)} uncalibrated and \textbf{(b)} early calibrated detectors on COCO. The area of the dots are proportional to the NMS processing time of the detectors. This threshold is typically set to $0.05$, in which case PAA and RS R-CNN have large \gls{NMS} processing time in \textbf{(a)}. Please see text for the discussion. 
        \vspace{-2.ex}
        }
        \label{fig:prenms}
\end{figure*}

\textbf{Early calibration reduces the sensitivity to background removal threshold} 
Before we conclude, we investigate an additional use-case of early calibration in which it reduces the sensitivity of the detectors to background removal threshold in terms of both \gls{AP} and efficiency.
As \gls{AP} provably benefits from more detections \cite{saod}, detectors prefer a small background removal threshold as the first step of post-processing (\cref{fig:cc}(b)).
\textit{To illustrate, $0.05$ is the common choice for COCO \cite{ATSS,paa,FasterRCNN} and it is as low as $10^{-4}$ for LVIS \cite{mmdetection,LVIS}.}
While this convention is preferred by AP, it can easily increase \gls{NMS} processing time especially for over-confident detectors.
This is because, for such detectors, the background removal step accepts redundant \glspl{TN}, which should have been rejected. 
Hence, due to this large number of redundant \glspl{TN} propagated to the \gls{NMS}, \gls{NMS} processing time significantly increases.
To illustrate on PAA, which uses a threshold of $0.05$ for COCO, NMS takes $29.2$ ms/image on a Nvidia 1080Ti GPU, while it only takes $\sim 0.6$ ms/image for ATSS and Faster R-CNN (F R-CNN).
This difference among the detectors can easily be noticed by comparing the areas of the dots at $0.05$ in \cref{fig:prenms}(a).
%
%
Specifically, we observed for PAA that $\sim 45$K detections are propagated to the NMS per image on average. 
After early calibration, this number of detections from the same threshold reduces to $\sim 2$K per image,  which now enables \gls{NMS} to take only $0.8$ ms/image as ideally expected. 
\cref{fig:prenms}(b) presents that NMS takes consistently between $0.6$ to $0.8$ ms/image for all detectors as the behaviour of the detectors are aligned.
%
}

%

%
\section{Conclusions} \label{sec:conc}
A direct result of the vastly different training regimes employed in training object detectors is that their predictions are miscalibrated such that some are more confident than the others.
%
This lack of consistency makes constructing an \gls{MoE} in a na\"ive approach futile, as the most confident detector dominates the predictions, even though its performance may not warrant this weighting.
Consequently, to address this we introduced \gls{CC} as a simple, principled and effective approach, which first calibrates the individual detectors before combining them appropriately through our refinement strategy.
Specifically, in the calibration stage we aligned the confidence with the \gls{IoU} of the detection with the object that it overlaps the most with.
We showed that this is an effective calibration target, resulting in accurate \glspl{MoE} with consistent gains across different detection tasks, reaching SOTA on many challenging detection benchmarks such as COCO and DOTA.
Whilst our choice of calibration function performed well on the settings demonstrated here, we further observe that increased gains can be achieved if the community develops more sophisticated calibration methods, an objective we leave to future work.
\clearpage

%
%

%
%
%
%
{
    \small
    \bibliographystyle{ieeenat_fullname}
    \bibliography{references}
}

\clearpage

\section*{APPENDICES}
\tableofcontents
\renewcommand{\thefigure}{A.\arabic{figure}}
\renewcommand{\thetable}{A.\arabic{table}}
\renewcommand{\theequation}{A.\arabic{equation}}
\renewcommand{\thesection}{A}
\section{Related Work} \label{app:relatedwork}


\paragraph{Mixture of Experts (MoEs)} The main aim of combining multiple  experts in the form of an \gls{MoE} is to leverage the expertise of each expert, which ideally specialises in different subpopulations of the input data \cite{jacobs1991adaptive, jordan1994hierarchical, xu1994alternative,yuksel2012twenty}. 
In the literature, this aim is achieved by a variety of techniques.
Some of the methods aggregates the predictions of individual experts \cite{ensembleod}.
Broadly speaking, from this perspective, \glspl{DE} \cite{ensembles}, in which multiple models are trained and then aggregated, can also be considered as an example in this group.
Another set of methods \cite{zhou2020bbn, MERCNN, wang2022longtailed, zhang2022diverse,cai2021ace}, as the majority of the techniques proposed in the deep learning era, allows the experts to share certain network components among experts, which are then combined (or selected) by a gating (a.k.a. routing) mechanism for the sake of efficiency.
Essentially, this gating mechanism decides on which expert to rely on conditioned on a particular subpspace of the input space.
%
%
%
%
%
 %
One particular and intuitive use-case of this group of methods is long-tailed classification \cite{zhou2020bbn, wang2022longtailed, cai2021ace, zhang2022diverse}, in which different experts specialise on different classes with various cardinalities, i.e. the classes with few or many training examples.
%
%
%
The shared features are then routed by a gating mechanism to leverage experts' specialisation in an efficient manner.
%
%
Another effective application of \glspl{MoE} is within the domain of variational autoencoders \cite{wu2018multimodal, shi2019variational, joy2022learning}, where different experts are generally utilised for different modalities.

\paragraph{Ensemble Methods in Object Detection}
Despite their aforementioned success in the classification literature, ensembling detectors either in the form of a \gls{DE} by using the same model with different initialisations or as an \gls{MoE} by combining different type of models has received very little attention \cite{MERCNN,ensembleod}.
Among the few existing works, \cite{ensembleod} combines  a set of detectors through various aggregation strategies, such as unanimous agreement between the detectors.
Therefore, this method combines off-the-shelf detectors, and accordingly, we refer to as Vanilla \gls{MoE} in the paper as a baseline.
However, we note that it does not consider either calibrating the detector or advanced aggregation techniques unlike our work. 
%
%
As the second work in this domain, Multi-Expert R-CNN\cite{MERCNN} is designed to exploit multiple experts in the R-CNN family, in which different R-CNN models correspond to different experts for a specific RoI.
Consequently, it is not applied to a wide range of different detectors such as the common one-stage detectors or transformer-based detectors.
%
%
%

%


\paragraph{Calibration in Object Detection}
As extensively studied for classification, calibration refers to the alignment of accuracy and confidence of a model~\cite{calibration,AdaptiveECE,verifiedunccalibration,rethinkcalibration,FocalLoss_Calibration,calibratepairwise}.
Specifically, a classifier is said to be \textit{calibrated} if it yields an accuracy of $p$ on its predictions with a confidence of $p$ for all $p \in [0,1]$. 
Earlier definitions for the calibration of detectors \cite{CalibrationOD,RelaxedSE} extend this definition with an objective to align the confidence of a detector with its precision,
\begin{align}\label{eq:dece}
    \mathbb{P}(\hat{c}_i = c_i | \hat{p}_i)  = \hat{p}_i, \forall \hat{p}_i \in [0,1],
\end{align}
where $\mathbb{P}(\hat{c}_i = c_i | \hat{p}_i)$ denotes the precision as the ratio of correctly classified predictions among all detections.
Extending from this definition, \cite{saod} takes into account that object detection is a joint task of classification and localisation.
Thereby defining the accuracy as the product of precision and average \gls{IoU} of \glspl{TP}, calibration of object detectors requires the following to be true
%
\begin{align}\label{eq:locaware_cal}
    \mathbb{P}(\hat{c}_i = c_i | \hat{p}_i) \mathbb{E}_{\hat{b}_i \in B_i(\hat{p}_i)}[ \mathrm{IoU}(\hat{b}_i, b_{\psi(i)})] = \hat{p}_i, \forall \hat{p}_i \in [0,1],
\end{align}
%
where $B_i(\hat{p}_i)$ is the set of \glspl{TP} with the confidence of $\hat{p}_i$ and $b_{\psi(i)}$ is the object that $\hat{b}_i$ matches with.
Then, \gls{LaECE} is obtained by discretizing the confidence score space into $J$ bins for each class. 
Specifically for class $c$, denoting the set of detections by $\hat{\mathcal{D}}^{c}$ and those in the $j$th bin by $\hat{\mathcal{D}}^{c}_j$ as well as the average confidence, precision and average \gls{IoU} of $\hat{\mathcal{D}}^{c}_j$ by $\bar{p}^{c}_{j}$, $\mathrm{precision}^{c}(j)$ and $\bar{\mathrm{IoU}}^{c}(j)$ respectively, \gls{LaECE} for class $c$ is defined as
%
\begin{align}
\label{eq:laece__}
   \mathrm{LaECE}^c 
    = \sum_{j=1}^{J} \frac{|\hat{\mathcal{D}}^{c}_j|}{|\hat{\mathcal{D}}^{c}|} \left\lvert \bar{p}^{c}_{j} - \mathrm{precision}^{c}(j) \times \bar{\mathrm{IoU}}^{c}(j)  \right\rvert.
\end{align}
%
Finally, the detector \gls{LaECE} is the average of $\mathrm{LaECE}^c$s over classes in the dataset, measuring the calibration error for a detector as a lower-better measure.

\paragraph{Comparative Summary} Different from the existing few works on the ensemble methods of object detectors, we comprehensively investigate how to obtain \glspl{MoE} in object detection using off-the-shelf detectors. While doing so, unlike existing work, (i) we identify that miscalibration of different detectors prevents them to be combined properly due to the peculiarities of the detectors; (ii) we introduce a strong aggregation technique which we refer to as Refining \gls{NMS} combining Soft \gls{NMS} and Score Voting from previous work; and (iii) we employ a very diverse set of detectors from a wide range of detection tasks. As for the miscalibration of the detectors, we rely on LaECE in Eq. \ref{eq:locaware_cal}. Consequently, \gls{CC} follows the main aim of the \gls{MoE} as the predictions from each expert are aggregated in a way that the strength of each detector is leveraged as we comprehensively demonstrate in our experiments. And besides, from the \gls{MoE} perspective  Refining \gls{NMS} can be considered as a form of a gating (or routing) mechanism, which decides the best way to combine the detections from different experts.
\renewcommand{\thesection}{B}
\section{Further Details on Calibration}
\label{app:calibration}
Throughout our paper, we calibrate the final confidence scores. One can also consider in object detection that, the raw confidence scores (before postprocessing) can be calibrated as well. For this reason, we first present early calibration in App. \ref{subsec:earlycal}. Then, for the sake of completeness, we discuss why we prefer post-hoc calibration methods in App. \ref{subsec:choice} as well as how to measure the calibration error of \gls{CC} based on \cref{eq:gen_calibration_} in App. \ref{subsec:calerror}.

\begin{figure}[]
\centering
\includegraphics[width=\textwidth]{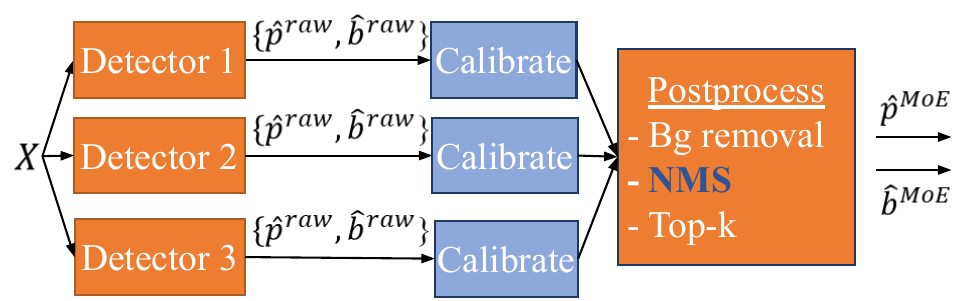}
\caption{Early calibration. Raw confidence scores $\hat{p}^{raw}$ are calibrated, and the standard  post-processing steps handle aggregation in which \gls{NMS} (in blue) removes the duplicates from multiple detectors.}
\label{fig:early_calibration}
\end{figure}

\subsection{Early Calibration of Object Detectors for Obtaining \glspl{MoE}}
\label{subsec:earlycal}
Up to now, we discussed calibrating final confidence scores $\hat{p}_i$ similar to \cite{CalibrationOD,saod}, which we show as \textit{late calibration} in \cref{fig:cc}.
Besides, we also investigate the effectiveness of \textit{early calibration} by calibrating the raw probabilities $\hat{p}^{raw}_i$ of the detectors as illustrated in \cref{fig:early_calibration}.
While we find both approaches to perform similar in \glspl{MoE}, we use late calibration as it is simpler owing to less number of final detections.
Still, as the first to investigate early calibration, we present an additional use-case of early calibration in App. \ref{app:exp} in which we show that it reduces the sensitivity of the model to the background removal threshold in post-processing.

\subsection{Choice of the Calibration Method} 
\label{subsec:choice}
There are multiple calibration methods used for object detection including training-time \cite{calibrationdomainshiftod,bridgeprecconf,MCCL} and post-hoc calibration methods \cite{saod,zadrozny2002transforming,calibration}.
Ideally, we would expect a calibration method to easily generalize to a wide range of detectors as well as detection tasks.
However, training-time calibration methods are typically incorporated into only a small set of object detectors and not tested on a variety of relevant tasks such as the rotated bounding box detection, which makes them unfit for purposes. 
On the other hand, post-hoc calibration methods formalize calibration as a regression task from the predicted final confidence to target confidence, making them easily applicable to any detectors and any  detection task.
%
%
Throughout this work, we investigate Linear Regression (LR) and Isotonic Regression (IR) \cite{zadrozny2002transforming} considering the criterion in Eq. \eqref{eq:gen_calibration_}.
\textit{We further observe that a \gls{CA} \gls{IR} calibrator for each detector obtained on 500 images is sufficient for \gls{CC}.} 
App \cref{app:exp} present extensive experimental analyses.

\subsection{Measuring the Calibration Error for \gls{MoE}} 
\label{subsec:calerror}
We introduce our calibration criterion in Eq. \eqref{eq:gen_calibration_}.
This criterion requires the confidence of a detection to align with its \gls{IoU} with the ground truth box that the detection overlaps the most.
As stated earlier, computing the calibration error based on this criterion corresponds to using an \gls{IoU} threshold of $0$ to validate \glspl{TP}. 
This is equivalent to using $\mathrm{precision}^{c}(j)=1$ for class $c$ in Eq. \eqref{eq:laece}, which then reduces to
\begin{align}
\label{eq:laece_}
   \mathrm{LaECE}^c 
    = \sum_{j=1}^{J} \frac{|\hat{\mathcal{D}}^{c}_j|}{|\hat{\mathcal{D}}^{c}|} \left\lvert \bar{p}^{c}_{j} - \bar{\mathrm{IoU}}^{c}(j)  \right\rvert.
\end{align}
where $ \hat{\mathcal{D}}^{c}$ denotes the set of detections for class $c$; $\hat{\mathcal{D}}^{c}_j$ is the set of detections in the $j$th bin for class $c$; $\bar{p}^{c}_{j}$ is the average confidence of the detections in $\hat{\mathcal{D}}^{c}_j$; and $\bar{\mathrm{IoU}}^{c}(j)$ is the average \gls{IoU} of the detections in $\hat{\mathcal{D}}^{c}_j$.
Following \cite{saod}, we use $J=25$ and average over $\mathrm{LaECE}^c$ of classes for the detector $\mathrm{LaECE}$.

In addition to $\mathrm{LaECE}$, here we define \gls{LaACE} and \gls{LaMCE} similar to the way how \gls{ECE} is extended to Average Calibration and Maximum Calibration Errors.
We find \gls{LaACE} and \gls{LaMCE} useful as they reduce the dominance of certain bins on the calibration error as in the case of \gls{LaECE}.
This is especially important for early calibration from which thousands of confidence scores are obtained from a single image, most of which have a confidence close to $0$.
Specifically, in our case, we define $\mathrm{LaACE}^c$ for class $c$ as 
\begin{align}
\label{eq:laace}
   \mathrm{LaACE}^c 
    = \sum_{j=1}^{J} \frac{1}{J} \left\lvert \bar{p}^{c}_{j} - \bar{\mathrm{IoU}}^{c}(j)  \right\rvert,
\end{align}
and $\mathrm{LaMCE}^c$ for class $c$ as 
\begin{align}
\label{eq:lamce}
   \mathrm{LaMCE}^c 
    = \max_{j \in \{1,2,..,J\}} \left\lvert \bar{p}^{c}_{j} - \bar{\mathrm{IoU}}^{c}(j)  \right\rvert.
\end{align}
Following \gls{LaECE}, we obtain \gls{LaACE} and \gls{LaMCE} for the detector by averaging over the classes.

\renewcommand{\thesection}{C}
\section{Theoretical Discussion on the Optimality of the Calibration Criterion in Eq. \eqref{eq:gen_calibration_}} \label{app:proof}

Lemma \ref{lemma:mocae} discusses the conditions under which an optimal  \gls{AP} can be achieved for a given set of pre-\gls{NMS} detections.

\begin{lemma}
\label{lemma:mocae}
Given a set of detection boxes for class $c$, denoted by $\mathcal{B}^{raw}=\{\hat{b}_1^{raw}, \hat{b}_2^{raw}, ... ,\hat{b}_L^{raw}\}$ , we first assume that the post-processing (\gls{NMS} in this case) does not remove \glspl{TP} and can remove duplicates in $\mathcal{B}^{raw}$\footnote{A \gls{TP} is a detection that has at least an IoU with a ground-truth of $\tau$ where $\tau$ is the IoU threshold to validate \glspl{TP}. In the case of more than one detection satisfy this criterion, the common convention is to accept the detection with the highest score as a \gls{TP} and the remaining ones are duplicates, which are counted as \glspl{FP} while computing the \gls{AP}.}. Let us denote the $k$th ground-truth box by $b_k$ and the detection set post \gls{NMS} by $\mathcal{B}=\{(\hat{b}_1,\hat{p}_1), (\hat{b}_2,\hat{p}_2) ..., (\hat{b}_N,\hat{p}_N)\}$ where $(\hat{b}_i,\hat{p}_i)$ correspond to a tuple with the $i$-th bounding box and associated confidence score. If the detections in $\mathcal{B}$ ensures that $\mathrm{IoU}(\hat{b_i},b_k) > \mathrm{IoU}(\hat{b_j},b_k)$ if $\hat{p}_i > \hat{p}_j$ for all $i \neq j$ and $k$ is the ground truth that each detection has maximum \gls{IoU} with, then $\mathcal{B}$ provides the optimal AP for class $c$ given $\mathcal{B}^{raw}$. The value of the optimal AP in this case is $\frac{\mathrm{N}_{TP}(\mathcal{B}^{raw})}{M}$,  where $\mathrm{N}_{TP}(\mathcal{B}^{raw})$ is the number of \glspl{TP} in $\mathcal{B}^{raw}$ and $M>0$ is the number of ground-truth objects from class $c$.
\end{lemma}
\begin{proof}
Below we show that $\mathcal{B}$ satisfies all the three necessary conditions required to maximize the \gls{AP} for any \gls{IoU} threshold $\tau$ (to identify   \glspl{TP}): 

\begin{compactenum}
    \item  \textit{$\mathcal{B}$ is to have $\mathrm{N}_{TP}(\mathcal{B}^{raw})$, that is, no \gls{TP} is to be removed in $\mathcal{B}^{raw}$ by postprocessing.} Note that this is handled by post-processing as an assumption in the Lemma.
    \item  \textit{The minimum confidence score of \glspl{TP} in $\mathcal{B}$ is to be higher than the maximum confidence score of \glspl{FP}.} This is also ensured considering (i) the assumption that is $\mathrm{IoU}(\hat{b_i},b_k) > \mathrm{IoU}(\hat{b_j},b_k)$ if $\hat{p}_i > \hat{p}_j$ for all $i \neq j$ and $k$ is the ground truth that each detection has maximum \gls{IoU} with; and (ii) the duplicates are removed by \gls{NMS}. As a result, all \glspl{TP} are ranked higher than all \glspl{FP}.
    \item \textit{$\mathcal{B}$ is to include the detection with the largest \gls{IoU} with each ground truth $k$.}\footnote{\label{fnlabel} When we consider the \gls{AP} for a single $\tau$, this criteria is not mandatory to be satisfied as the conventional \gls{AP} consider localisation performance loosely\cite{LRPPAMI}. However, the localisation performance is an important aspect of an object detector and it is considered by various performance measures such as COCO-style \gls{AP}, which corresponds to the average over \glspl{AP} with 10 different $\tau$ thresholds or LRP Error\cite{LRPPAMI}. As a result, while we consider the \gls{AP} for a single $\tau$ in this section, in order for our theoretical justifications to be applicable to other performance measures, we also take this criterion in account.}. \gls{NMS} selects the detection with the highest confidence score among a group of overlapping detections. Considering that \gls{NMS} does not remove any \glspl{TP} and $\mathrm{IoU}(\hat{b_i},b_k) > \mathrm{IoU}(\hat{b_j},b_k)$ for all $i$ and $j$ if $\hat{p}_i > \hat{p}_j$ and $i \neq j$ holds, \gls{NMS} survives the detections with the best localisation quality as they have the highest scores in their groups. As  a result, this condition is also satisfied.
\end{compactenum}
To compute the area under the precision-recall curve, we need the precision and recall pairs. Please note that once the aforementioned criteria are satisfied, the precision will be $1$ when recall interval is between  $[0, \frac{\mathrm{N}_{TP}(\mathcal{B}^{raw})}{M}]$; and will be zero beyond this. Therefore, the area under the precision-recall curve trivially turns out to be $\frac{\mathrm{N}_{TP}(\mathcal{B}^{raw})}{M}$.
\end{proof} 

\paragraph{On the Optimality of Eq. \eqref{eq:gen_calibration_}} Based on Lemma \ref{lemma:mocae}, we now present our theorem and its proof.

\begin{theorem}
\label{theorem:mocae}
Assume that $E$ different experts are combined in the form of an \gls{MoE} and the detection set of $e$-th expert for class $c$ is denoted by $\mathcal{B}_e$. Given the union of these detections (before aggregation) is denoted by $\mathcal{B}^{raw}=\bigcup_{e=1}^E \mathcal{B}_e$, the \gls{MoE} using perfectly calibrated experts in terms of Eq. \eqref{eq:gen_calibration_} yields optimal AP for class $c$ over any possible \glspl{MoE} following Lemma \ref{lemma:mocae}. Accordingly, denoting the number of \glspl{TP} in $\mathcal{B}^{raw}$ by $\mathrm{N}_{TP}(\mathcal{B}^{raw})$ and the number of ground-truth objects for class $c$ by $M>0$, the resulting \gls{AP} is $\frac{\mathrm{N}_{TP}(\mathcal{B}^{raw})}{M}$.
\end{theorem}

\begin{proof}
Please note that it is trivial to show that the detections in $\mathcal{B}$ of the perfectly calibrated detectors ensures the  requirement in Lemma \ref{lemma:mocae} that $\mathrm{IoU}(\hat{b_i},b_k) > \mathrm{IoU}(\hat{b_j},b_k)$ if $\hat{p}_i > \hat{p}_j$ for all $i \neq j$ and $k$ is the ground truth that each detection has maximum \gls{IoU} with. This is because the calibration target ensures that $\hat{p}_i=\mathrm{IoU}(\hat{b_i},b_k)$ for each detection $i$. As a result, following from Lemma \ref{lemma:mocae}, the theorem holds.
\end{proof}

\blockcomment{
\begin{proof}
    We show that $\mathcal{B}$, the detections of \gls{MoE} after \gls{NMS}, obtained by Eq. \eqref{eq:gen_calibration_} satisfies all three conditions outlined in the proof of Lemma \ref{lemma:mocae}. 
    \begin{compactitem}
        \item As \gls{NMS} does not remove any \glspl{TP}, the number of \glspl{TP} in $\mathcal{B}$ remains the same with that before \gls{NMS}, satisfying the first condition.
        \item The second condition requires ranking \glspl{TP} before \glspl{FP}. To show that, we will benefit from the fact that the confidence scores are perfectly calibrated, in which the confidence matches the \gls{IoU} (Eq. \eqref{eq:gen_calibration_}). As a result, while the minimum-scoring \gls{TP} has a larger score than $\tau$ (the \gls{TP} validation threshold), the maximum-scoring \gls{FP} has a smaller score than $\tau$ also considering that \gls{NMS} removes the duplicates.
        \item The third condition requires choosing the best-localised detection for each ground truth. Please note that \gls{NMS} picks the detection with the highest confidence, and as the scores are perfectly calibrated, the best-localised detection will be survived. 
    \end{compactitem}
    As all three conditions to achieve optimal \gls{AP} is satisfied, the \gls{AP} of \gls{CC} is $\frac{\mathrm{N}_{TP}(\mathcal{B}^{raw})}{M}$ which is the optimal \gls{AP}.
\end{proof}
}
Please note that it is trivial to show that Theorem \ref{theorem:mocae} generalizes to the cases in which the dataset involves multiple classes and COCO-style AP is used. This is because the former case is estimated as the average of the \glspl{AP} over different classes and the latter is simply the average of the \glspl{AP} over different \gls{IoU} thresholds (please see footnote \ref{fnlabel} for further discussion). Consequently, as the class-wise \glspl{AP} are optimal, so do the dataset \gls{AP} and the COCO-style \gls{AP}. Please refer to App. \ref{subsec:practicalimplications} for the experiment presenting the effect of \cref{theorem:mocae} using an Oracle \gls{MoE}.

\blockcomment{
\section{Theoretical Discussion on the Optimality of the Calibration Criterion in Eq. \eqref{eq:gen_calibration_}} \label{app:proof}

Before presenting our theorem on the optimality of Eq. \eqref{eq:gen_calibration_} in terms of \gls{AP}, we first build background including our assumptions on \gls{NMS} and what the optimal \gls{AP} is. 

\paragraph{The assumption on \gls{NMS}} Broadly speaking, the standard \gls{NMS} algorithm iteratively groups the detections by considering the maximum scoring detection among the detections.
Conventionally, the grouping criterion is the \gls{IoU} of the detections with the maximum scoring scoring detection.
Specifically, if the \gls{IoU} of a detection with the maximum scoring detection is larger than a predefined threshold, then the maximum scoring detection survives.
On the other hand, the detections ensuring this \gls{IoU}-based criterion are considered as duplicates and removed from the detection set.
For the purposes of theoretical analyses, it is not very trivial to foresee which detections will be survived or removed once we follow the definition of the standard \gls{NMS}. 
For example, in the case of highly-overlapping ground truths, \gls{NMS} can easily remove a \gls{TP}.
To exclude such edge cases and focus solely on the effect of the calibration target on obtaining \glspl{MoE} , we simplify how \gls{NMS} operates and define Reference \gls{NMS} in the following.

\begin{definition}
\label{def:ReferenceNMS}
Denoting $i$th ground truth and detection box (before \gls{NMS}) $b_i$ and $\hat{b}_i$ respectively, and the \gls{TP} validation threshold (in terms of \gls{IoU}) by $\tau$, Reference \gls{NMS} first returns the maximum scoring detection among the detections $\hat{b_i}$ that satisfy $\mathrm{IoU}(\hat{b_i},b_i)>\tau$ for each ground truth $b_i$. For the remaining detections, Reference \gls{NMS} follows the Standard \gls{NMS}. We note that  Reference \gls{NMS} operates class-wise similar to the common setting for the standard \gls{NMS}.
\end{definition}

Please note that Reference \gls{NMS} keeps the key aspect of the Standard \gls{NMS}, in which the detections are prioritized following the confidence of the detections.
However, differently we assume that Reference \gls{NMS} can survive the maximum scoring detection among \gls{TP} candidates for each ground truth; and remove the duplicates properly.

\paragraph{What is the Optimal \gls{AP} for \glspl{MoE}?}  Lemma \ref{lemma:mocae} shows that  the optimal AP value of an \gls{MoE} can be determined.

\begin{lemma}
\label{lemma:mocae}
Assume that $e$ different experts performs inference on image $X$, which includes $M>0$ objects from class $c$. Let us denote the detection set of $i$th expert by $\hat{D}_{final}^i$ for class $c$. Furthermore, following the \gls{MoE} pipeline (in \cref{fig:cc}), the detections are combined before they are aggregated as $\hat{D}=\bigcup_{i=1}^e \hat{D}_{final}^i$. Denoting the number of \glspl{TP} in $\hat{D}$ by $\mathrm{N}_{TP}(\hat{D})$ and using Reference \gls{NMS} (Definition \ref{def:ReferenceNMS}) on $\hat{D}$, the optimal AP of the resulting \gls{MoE} for class $c$ for any $\tau$ is $\frac{\mathrm{N}_{TP}(\hat{D})}{M}$\footnote{Note that a \gls{TP} detection is a detection in $\hat{D}_{final}^{MoE}$ that has at least an IoU with a ground truth of $\tau$. In the case of more than one detection satisfy this criterion for the same ground truth, one common convention is to accept the detection with the highest score as a \gls{TP} and the remaining ones are duplicates, that is, they are counted as \glspl{FP}.}.
\end{lemma}
\begin{proof}
We denote the final detection set of the \gls{MoE} by $\hat{D}_{final}^{MoE}$, that is $\hat{D}_{final}^{MoE}$ is the output of Reference \gls{NMS} given $\hat{D}$. Considering that \gls{AP} is defined as the area under the precision-recall curve, the \gls{AP} of $\hat{D}_{final}^{MoE}$ for any $\tau$ is maximized when the following criteria are satisfied: 
\begin{compactitem}
    \item  $\hat{D}_{final}^{MoE}$ is to have $\mathrm{N}_{TP}(\hat{D})$, as the number of total \glspl{TP} discovered by the union of the $e$ experts.
    \item  The minimum confidence score of \glspl{TP} in $\hat{D}_{final}^{MoE}$ is to be higher than the maximum confidence score of \glspl{FP}.
    \item $\hat{D}_{final}^{MoE}$ is to have the detection with the largest \gls{IoU} with the ground truth $i$ for each ground truth $i$\footnote{When we consider the \gls{AP} for a single $\tau$, this criteria is not mandatory to be satisfied as the conventional \gls{AP} consider localisation performance loosely\cite{LRPPAMI}. However, the localisation performance is an important aspect of an object detector and it is considered by various performance measures such as COCO-style \gls{AP}, which corresponds to the average over \glspl{AP} with 10 different $\tau$ thresholds or LRP Error\cite{LRPPAMI}.}. 
\end{compactitem}
To compute the area under the precision-recall curve, we need the precision and recall pairs. Please note that once the aforementioned criteria are satisfied, the precision will be $1$ when recall is between $[0, \frac{\mathrm{N}_{TP}(\hat{D})}{M}]$. Then, the precision will reduce to $0$ in the recall interval between $(\frac{\mathrm{N}_{TP}(\hat{D})}{M},1]$. Then, the area under the precision-recall curve, the \gls{AP}, is $\frac{\mathrm{N}_{TP}(\hat{D})}{M}$.
\end{proof} 
\blockcomment{
We now provide an intuition of what we mean by the upper bound of the $\mathrm{AP_\tau}$ of \gls{MoE} in an image $X$  where $\tau$ is the \gls{TP} validation threshold.
Given the detection sets from different $e$ experts by $\hat{D} = \hat{D}_{final}^1, \hat{D}_{final}^2, ..., \hat{D}_{final}^e$, the \gls{NMS} simply operates on the input $\hat{D}=\bigcup_{i=1}^e \hat{D}_{final}^i$ for each class $c$ and yields $\hat{D}_{final}$.
We are interested in finding $\hat{D}_{final}$ that can provide the best performance.
Considering how \gls{NMS} operates, as aforementioned $\hat{D}_{final} \subseteq \hat{D}=\bigcup_{i=1}^e \hat{D}_{final}^i$.
Therefore, in order to reach the upper bound $\mathrm{AP_\tau}$ of $X$ given $\hat{D}=\bigcup_{i=1}^e \hat{D}_{final}^i$, the output of \gls{NMS} for each class $c$ in each image should satisfy the following criteria:
\begin{compactitem}
    \item \textit{1.Maximum achievable number of \glspl{TP}.} $\hat{D}_{final}^i$ is to include the maximum achievable number of \gls{TP} detections defined as the number of \gls{TP} detections in $\hat{D}=\bigcup_{i=1}^e \hat{D}_{final}^i$. Specifically, please note that it does not necessarily correspond to the number of ground truths.\footnote{Note that a \gls{TP} detection is a detection in $\hat{D}_{final}$ that has at least an IoU with a ground truth of $\tau$. In the case of more than one detection satisfy this criterion, the common convention is to accept the detection with the highest score as a \gls{TP} and the remaining ones are duplicates, that is, they are \glspl{FP}. },
    \item \textit{2.Maximum achievable average \gls{IoU}.} The average \gls{IoU} of the \glspl{TP} in $\hat{D}_{final}^i$ is to correspond to the maximum achievable average \gls{IoU}, which can be obtained as follows: finding the \gls{TP} detections in $\hat{D}=\bigcup_{i=1}^e \hat{D}_{final}^i$ that have the maximum \gls{IoU} with the detections and averaging their \glspl{IoU}.
    \item \textit{3.Maximum achievable ranking of detections.} A mixture can satisfy the maximum achievable number of \glspl{TP} and average \gls{IoU} criteria, but still be suboptimal in terms of \gls{AP} in the case that the confidence scores of \glspl{TP} are low compared to those of \glspl{FP}. Ideally, one would expect that the minimum confidence scores of \gls{TP} detections to be higher than the maximum confidence score of \gls{FP} detections. However, this might be achieved due to suboptimality of \gls{NMS} in determining duplicate detections. As an example, consider a scenario in which there are two ground truths ($b_i$ and $b_j$) and three detections ($\hat{b}_i$, $\hat{b}_j$ and $\hat{b}_k$) are returned by \gls{NMS}. In this scenario, assume that $\mathrm{IoU}(\hat{b}_i, b_i)=0.75$, $\mathrm{IoU}(\hat{b}_j, b_j)=0.50$ and $\mathrm{IoU}(\hat{b}_k, b_i)=0.70$. This implies that $\hat{b}_k$ is a duplicate detection\footnote{A duplicate detection for object $b_i$ can be defined as follows: $\hat{b}_j$ is a duplicate detection if NMS returns both $\hat{b}_i$ and $\hat{b}_j$ such that $\mathrm{IoU}(b_i, \hat{b}_j)>\tau$, $\mathrm{IoU}(b_j, \hat{b}_j)>\tau$ and the confidence score of $\hat{b}_i$ is higher than $\hat{b}_j$.} for the object $b_i$. Accordingly, the upper bound precision of this setting from $\tau=0.50$ is $0.66$ due to the limitation of the standard \gls{NMS}. Therefore, this criterion should take into account the duplicates. Formally speaking, there are two criteria: (i)  the minimum confidence scores of \gls{TP} detections and duplicate detections are to be higher than the maximum confidence score of \gls{FP} detections; and (ii)  \gls{TP} for the object $b_i$ should be ranked higher than all of the duplicates of $b_i$.
\end{compactitem}
}

\paragraph{On the Optimality of Eq. \eqref{eq:gen_calibration_}} Having provided the background, we now present our theorem and its proof.

\begin{theorem}
\label{theorem:mocae}
Assuming that Reference \gls{NMS} is used to aggregate the detections from individual experts (Definition \ref{def:ReferenceNMS}), the \gls{MoE} using perfectly calibrated experts in terms of Eq. \eqref{eq:gen_calibration_} yields optimal AP for any $\tau$ for image $X$ on class $c$ over any possible \glspl{MoE}. This \gls{AP} is $\frac{\mathrm{N}_{TP}(\hat{D})}{M}$ following Lemma \ref{lemma:mocae}.
\end{theorem}

\begin{proof}
    We need to show that  $\hat{D}_{final}^{MoCaE}$ obtained by Eq. \eqref{eq:gen_calibration_} satisfies all three conditions outlined in the proof of Lemma \ref{lemma:mocae}. 
    Considering that Reference \gls{NMS} returns the maximum scoring detection for each ground truth, the number of \glspl{TP} in $\hat{D}_{final}^{MoCaE}$ is equal to $\mathrm{N}_{TP}(\hat{D})$, satisfying the first condition. The second condition requires ranking \glspl{TP} before \glspl{FP}. To show that, we will benefit from the fact that the confidence scores are perfectly calibrated, in which the confidence matches the \gls{IoU} (Eq. \eqref{eq:gen_calibration_}). As a result, while the minimum-scoring \gls{TP} has a larger score than $\tau$ (the \gls{TP} validation threshold), the maximum-scoring \gls{FP} has a smaller score than $\tau$ also considering that Reference \gls{NMS} removes the duplicates. Finally, the third condition requires choosing the best-localised detection for each ground truth. Please note that \gls{NMS} picks the detection with the highest confidence, and as the scores are perfectly calibrated, the best-localised detection will be survived. As all three conditions to achieve optimal \gls{AP} is satisfied, the \gls{AP} of \gls{CC} is $\frac{\mathrm{N}_{TP}(\hat{D})}{M}$, which is the optimal \gls{AP}.
\end{proof}

\blockcomment{
\begin{proof}
We use proof by induction on the number of ground truths in $X$.

\underline{Base Case:} In the case that there is only a single ground truth in $X$ and there are two possible cases.

\textit{Case (i). None of the experts has a \gls{TP} detection.} As $\hat{D}_{final} \subseteq \hat{D}=\bigcup_{i=1}^e \hat{D}_{final}^i$ and $\bigcup_{i=1}^e \hat{D}_{final}^i$ does not have any $TP$, then the upper bound performance is $0$ for AP, which is trivially satisfied by the perfectly calibrated experts.

\textit{Case (ii). At least one of the experts detects the single ground truth with $IoU \geq \tau$.} In this case, as the all the expert detectors are perfectly calibrated, the confidence score in each detection in $\hat{D}=\bigcup_{i=1}^e \hat{D}_{final}^i$ corresponds to the \gls{IoU} of the detection with the single ground truth in the image. As a result, the highest scoring detection has the highest \gls{IoU}, which makes it to be survived from the \gls{NMS} to $\hat{D}_{final}$. In fact, this satisfies all three criteria mentioned above for upper bound AP, completing the proof of the base case. In particular, the maximum number of \gls{TP} detections is achieved as $1$, the \gls{TP} has the largest \gls{IoU} and the confidence score of the \gls{TP} is the largest in $\hat{D}_{final}$

\underline{Induction Hypothesis:} If there are $M$ ground truths in image $X$ and relying on the assumed \gls{NMS}, then the \gls{MoE} using perfectly calibrated experts in terms of Eq. \eqref{eq:gen_calibration_} yields the upper bound AP of $X$ over any possible \glspl{MoE}.

\underline{Induction Step:} Now assume that there are $M+1$ ground truths in image $X$. Without loss of generality, let's leave a single ground truth $b_i$ out. Then, thanks to induction hypothesis, we already know that the upper bound AP is achieved by any $M$ ground truths in $X$, i.e., the three criteria for upper bound AP are satisfied. Now considering all $M+1$ ground truths, there are again two possible cases similar to the base case:

\textit{Case (i). None of the experts detects $b_i$.} In this case, the \glspl{TP} and \glspl{FP} as well as their ordering in  $\hat{D}_{final}$ do not change compared to the image with $M$ ground truths. Also, note that the number of maximum achievable \glspl{TP} does not change as none of the experts detects $b_i$. As a result, all three criteria naturally follows from the induction hypothesis. 

\textit{Case (ii). At least one of the experts detects $b_i$ with $IoU \geq \tau$.} In this case the number of maximum achievable \gls{TP} detections increases by $1$. Please note that the assumed \gls{NMS} algorithm will return the detection with the highest score for $b_i$, which also has the largest \gls{IoU} as each expert is perfectly calibrated. This implies that the first and the second criteria are satisfied for the upper bound \gls{AP}. As for the third criterione, \gls{NMS} returns one more \gls{TP} and potentially some duplicates. Both \gls{TP} and the duplicates will have confidence larger than all \glspl{FP}, ensuring the first condition of this criterion. Besides, thanks to calibration, the \gls{TP} corresponding to the box $b_i$ will have a score larger than the duplicates corresponding to the same box. Consequently, also considering the induction hypothesis, third criterion is also satisfied, concluding the proof.
\end{proof}
}
Please refer to App. \ref{subsec:practicalimplications} for the experiment presenting the effect of Theorem \cref{theorem:mocae} using an Oracle \gls{MoE}.

%
%
%
}

\renewcommand{\thesection}{D}
\section{Further Details on Refining NMS} \label{app:cc}
As described in \cref{subsec:cluster}, we combine \textit{Soft NMS} with \textit{Score Voting} while aggregating the detections of different detectors in our \gls{CC} approach.
Here, for the sake of completeness, we present these approaches.

Starting with the standard \gls{NMS},
%
given a set of raw detections after background removal (\cref{sec:relatedwork}), the standard \gls{NMS} first selects the maximum-scoring detection $\{\hat{b}_\alpha,\hat{p}_\alpha\}$ and then groups the detections that have an \gls{IoU} with  $\{\hat{b}_\alpha,\hat{p}_\alpha\}$ larger than a predefined \gls{IoU} threshold\footnote{Considering the common usage of \gls{NMS}, we assume that \gls{NMS} operates class-wise. Hence, the predicted class label is not explicitly included in the detection representation.}.
Then, the standard \gls{NMS} survives $\{\hat{b}_\alpha,\hat{p}_\alpha\}$ by placing it to the final detection set and discards all other raw detections from that group assuming that they are duplicates.
This process takes place until all raw detections are either moved to the final detection set or discarded.
Instead of removing the detections (i.e., the detections other than $\{\hat{b}_\alpha,\hat{p}_\alpha\}$) completely, Soft NMS decreases their confidence scores as a function of their overlap with $\{\hat{b}_\alpha,\hat{p}_\alpha\}$.
More specifically, Soft NMS has two variants determined by the type of this rescoring function.
The first one is called \textit{Linear Soft NMS}, in which the confidence scores are rescored such that
\begin{align}
    \label{eq:linearnms}
  \hat{p}_i =
    \begin{cases}
      \hat{p}_i & \text{if $\mathrm{IoU}(\hat{b}_i, \hat{b}_{\alpha}) < \mathrm{IoU}_{NMS}$}\\
      \hat{p}_i \times (1 - \mathrm{IoU}(\hat{b}_i, \hat{b}_{\alpha})) & \text{else,}
    \end{cases}       
\end{align}
where $\mathrm{IoU}_{NMS}$ is the predefined \gls{IoU} threshold for \gls{NMS}, set to $0.30$ in \cite{SoftNMS}. 
Note that Eq. \eqref{eq:linearnms} corresponds to the standard \gls{NMS} if $\hat{p}_i$ is set to $0$ for the case that  $\mathrm{IoU}(\hat{b}_i, \hat{b}_{\alpha}) \geq \mathrm{IoU}_{NMS}$ in which $\hat{b}_i$ overlaps with $\hat{b}_{\alpha}$ more than a threshold.
Differently, Soft NMS decreases the score $\hat{p}_i$ by considering the overlap of $\hat{b}_i$ with $\hat{b}_{\alpha}$.
In the case of a higher overlap, the score $\hat{p}_i$ is reduced more with the intuition that $\hat{b}_i$ is more likely to detect the same object with $\hat{b}_{\alpha}$.
With the same intuition, the second Soft NMS variant, for \textit{Gaussian Soft NMS} modifies the scores as follows,
\begin{align}
    \hat{p}_i = \hat{p}_i e^{-\frac{\mathrm{IoU}(\hat{b}_i, \hat{b}_{\alpha})^2}{\sigma_{\mathrm{NMS}}} },
\end{align}
where $\sigma_{\mathrm{NMS}}$ is a hyper-parameter to control how much to suppress the scores such that a smaller $\sigma_{\mathrm{NMS}}$ implies that $\hat{p}_i$ is suppressed more.
We provide experiments on how $\sigma_{\mathrm{NMS}}$ affects the performance and find that $\sigma_{\mathrm{NMS}} \in [0.40,0.60]$ typically performs well for \gls{CC}.
Therefore, as Soft NMS is less rigid in removing overlapping detections, it naturally leads to improved recall. 

Furthermore, we also consider Score Voting \cite{paa}, which aims to refine the final detections after \gls{NMS} (or Soft \gls{NMS}). 
Inspired by \cite{KLLoss}, the bounding box of a final detection is refined by utilizing the raw detections and their confidence.
Specifically, the refined box $\hat{b}_i$ is obtained as the weighted average of raw bounding boxes $\hat{b}_j$ (i.e., before \gls{NMS}) as follows,
\begin{align}
    \hat{b}_i = \frac{\sum_j \hat{p}_j \hat{IoU}_j \hat{b}_j }{\sum \hat{p}_j \hat{IoU}_j},
\end{align}
where $\hat{IoU}_j$ is 
\begin{align}
    \hat{IoU}_j =  e^{-\frac{1-\mathrm{IoU}(\hat{b}_i, \hat{b}_{j})^2}{\sigma_{\mathrm{SV}}} },
\end{align}
such that $\sigma_{\mathrm{SV}}$ is the hyper-parameter of Score Voting, which we set to $0.04$ in all of our experiments.

Note that while Soft \gls{NMS} keeps the box as it is and updates the confidence score, Score Voting does the opposite by keeping the score as it is and refines the box.
As a result, these two approaches adopted in Refining \gls{NMS} complement each other well, enabling us to obtain a strong aggregator for \gls{CC}.

\renewcommand{\thesection}{E}
\section{Further Experiments and Analyses} \label{app:exp}
Here, we present further experiments and analyses that are not included in the main text due to space limitation.

\subsection{Further Details on Used Models}
\label{app:further_details_models}
We provide the details of the used models as follows. 
We again note that we haven't trained any model but used off-the-shelf detectors with the exception of \glspl{DE}.
Here we provide further details on the used detectors.
Still, as it is not feasible to provide all of the details, we also present the papers and repositories that we borrow these off-the-shelf models in order to ensure the reproducibility of our results. 

\paragraph{Object Detection}
We use two different configurations.
In the first one, we employ three detectors with ResNet-50 \cite{ResNet} with FPN \cite{FeaturePyramidNetwork} backbone. These detectors are:
\begin{compactitem}
    \item Rank \& Sort R-CNN (RS R-CNN) \cite{RSLoss} is a recent representative of the two-stage R-CNN family \cite{FasterRCNN,RFCN,dynamicrcnn} optimizing a ranking-based loss function,
    \item Adaptive Training Sample Selection (ATSS) \cite{ATSS} is a common one stage baseline,
    \item Probabilistic Anchor Assignment (PAA) \cite{paa} relies on the one-stage ATSS architecture but with a different anchor assignment mechanism and postprocessing of the confidence scores.
\end{compactitem}
We obtain RS R-CNN and ATSS from \cite{saod} and PAA from \cite{mmdetection}.
All these detectors are trained for 36 epochs using multi-scale training data augmentation in which the shorter side of the image is resized within the range of [480, 800] for RS R-CNN and ATSS and [640, 800] for PAA.
We do not use Soft NMS and Score Voting for the single detectors.

In our second setting, we use the following detectors:
\begin{compactitem}
\item YOLOv7 \cite{yolov7} with a large convolutional backbone following its original setting,
\item QueryInst \cite{queryinst} as a transformer-based detector with a Swin-L \cite{swin} backbone,
\item ATSS with transformer-based dynamic head \cite{dyhead} and again Swin-L backbone.
\end{compactitem}
Again, we obtain YOLOv7 and dynamic head from mmdetection \cite{mmdetection} and use the official repository of QueryInst \cite{queryinst}.

Furthermore, to improve the \gls{SOTA} on COCO \textit{test-dev}, we use two of the most recent and strong public detectors, EVA \cite{EVA} and Co-DETR \cite{codetr}.
EVA \cite{EVA} is a foundation model for computer vision that utilises Cascade Mask R-CNN \cite{CascadeRCNN} to perform object detection wheras Co-DETR \cite{codetr} is a recent transformer-based detector.
For both of them, we directly consider the official repositories and do not change any settings, including the Soft-NMS for EVA \cite{EVA}.

\paragraph{Rotated Object Detection}
For rotated object detection, we use RTMDet and LSKN as two different detectors.
We obtain RTMDet again from mmdetection (which is also the official repository for RTMDet) and LSKN from its official repository \cite{lskn}.

\paragraph{Open-Vocabulary Object Detection}
We use two of the most recent works on vision-language foundation models literature, which can be listed as follows:
\begin{compactitem}
        \item Grounding DINO \cite{groundingdino} is a transformer-based detector.
        \item MQ-GLIP from the recent \cite{mqdet}, an anchor-based detector that introduces multi-modal queries on top of GLIP \cite{glip}.
\end{compactitem}
For both Grounding Dino \cite{groundingdino} and MQ-GLIP \cite{mqdet}, we consider the versions employing Swin-T \cite{swin} as the backbone.
We do not perform any prompt engineering and do not change any settings.
Furthermore, we directly utilise the official GitHub repositories for both of the models.

\paragraph{Instance Segmentation}
We use three different Mask R-CNN variants for instance segmentation:
\begin{compactitem} 
    \item The vanilla Mask R-CNN \cite{MaskRCNN} with ResNeXt-101 \cite{ResNext} backbone, softmax classifier and using Repeat Factor Sampling (RFS) to address the long-tailed nature of LVIS,
    \item Mask R-CNN with ResNet-50, sigmoid classifier, trained with RS Loss \cite{RSLoss} and RFS,
    \item Mask R-CNN with ResNet-50, softmax classifier, trained with Seesaw Loss\cite{seesawloss} but no RFS.
\end{compactitem}
We obtain Vanilla Mask R-CNN and Seesaw Loss from mmdetection. 
As for Mask R-CNN trained with RS Loss, we use the official repository of RS Loss \cite{RSLoss} in which it is trained for 12 epochs using multi-scale training augmentation.
The other Mask R-CNN variants also employ multi-scale training augmentation and the Vanilla Mask R-CNN is trained for 12 epochs as well.
Differently, Mask R-CNN with Seesaw Loss is trained for 24 epochs and uses the  mask normalization technique proposed in the same paper\cite{seesawloss}.

\blockcomment{
\begin{table}[t]
    \centering
    \caption{SAOD-Gen dataset results, which includes domain shift compared to the training set that is COCO.}
    \label{tab:app_earlier}
    \scalebox{0.90}{
    \begin{tabular}{c||c|c|c||c|c|c} 
    \toprule
    \midrule
    Method&$\mathrm{AP}$&$\mathrm{AP_{50}}$&$\mathrm{AP_{75}}$&$\mathrm{AP_{S}}$&$\mathrm{AP_{M}}$&$\mathrm{AP_{L}}$ \\ \midrule
    Faster R-CNN \cite{FasterRCNN}&$38.1$&$59.3$&$41.0$&$23.5$&$42.0$&$48.5$\\
    Retina Net \cite{FocalLoss, FocalLossConf}&$36.9$&$55.5$&$39.7$&$21.8$&$40.4$&$50.1$\\
    FoveaBox \cite{FoveaBox}&$36.7$&$55.7$&$38.8$&$21.3$&$40.3$&$49.8$\\ \midrule 
    Vanilla \gls{MoE}&$37.9$&$56.7$&$40.8$&$22.8$&$41.9$&$51.3$ \\
    &$\nimp{1.6}$&$\nimp{1.2}$&$\nimp{1.8}$&$\nimp{0.3}$&$\nimp{1.3}$&$\nimp{3.0}$  \\ \midrule 
    \gls{CC}&$\mathbf{31.9}$&$\mathbf{43.5}$&$\mathbf{34.9}$&$\mathbf{12.2}$&$\mathbf{29.0}$&$\mathbf{44.4}$ \\
     (Ours) &$\imp{1.3}$&$\imp{0.8}$&$\imp{1.4}$&$\imp{1.3}$&$\imp{1.6}$&$\imp{0.7}$ \\ \midrule
    \bottomrule
    \end{tabular}
    }
\end{table}

\begin{table}[t]
    \centering
    \caption{SAOD-Gen dataset results, which includes domain shift compared to the training set that is COCO.}
    \label{tab:app_saodgen}
    \scalebox{0.90}{
    \begin{tabular}{c||c|c|c||c|c|c} 
    \toprule
    \midrule
    Method&$\mathrm{AP}$&$\mathrm{AP_{50}}$&$\mathrm{AP_{75}}$&$\mathrm{AP_{S}}$&$\mathrm{AP_{M}}$&$\mathrm{AP_{L}}$ \\ \midrule
    RS R-CNN \cite{RSLoss}&$28.6$&$41.2$&$31.3$&$10.5$&$25.7$&$40.1$\\
    ATSS \cite{ATSS}&$28.7$&$39.8$&$31.2$&$10.6$&$26.5$&$39.6$\\
    D-DETR \cite{DDETR}&$30.6$&$42.7$&$33.5$&$10.9$&$27.4$&$43.7$\\ \midrule 
    Vanilla \gls{MoE}&$29.0$&$41.5$&$31.7$&$10.6$&$26.1$&$40.7$ \\
    &$\nimp{1.6}$&$\nimp{1.2}$&$\nimp{1.8}$&$\nimp{0.3}$&$\nimp{1.3}$&$\nimp{3.0}$  \\ \midrule 
    \gls{CC}&$\mathbf{31.9}$&$\mathbf{43.5}$&$\mathbf{34.9}$&$\mathbf{12.2}$&$\mathbf{29.0}$&$\mathbf{44.4}$ \\
     (Ours) &$\imp{1.3}$&$\imp{0.8}$&$\imp{1.4}$&$\imp{1.3}$&$\imp{1.6}$&$\imp{0.7}$ \\ \midrule
    \bottomrule
    \end{tabular}
    }
\end{table}

\subsection{Results on Different Detectors and Datasets}

\paragraph{Instance Segmentation on COCO}

\paragraph{Object Detection on SAOD-Gen}

\cref{tab:app_saodgen}
We finally explore whether the improvement generalizes to a scenario with domain shift using SAOD-Gen dataset \cite{saod}.
SAOD-Gen dataset consists of 45K images from exactly the same classes of COCO dataset, but collected from Objects365 dataset \cite{Objects365}; thereby introducing natural domain shift.
Thanks to this overlap across classes, we use our previously-trained calibrators for COCO dataset.
In addition to RS R-CNN and ATSS, here we include Deformable DETR (D-DETR) \cite{DDETR} as a transformer-based detector.
The results are presented in \cref{tab:app_saodgen} in which two weaker detectors push the performance of D-DETR by up to $1.3$ AP using our Late \gls{CC} approach.
}
\begin{table*}[t]
    \setlength{\tabcolsep}{0.25em}
    \centering
    \caption{Accuracy and calibration performance of uncalibrated, early calibrated and late calibrated models on COCO \textit{mini-test}. A red cell indicates a notable  \gls{AP} drop compared to the uncalibrated detector while a green cell implies consistency. CW: Class-wise, CA: class-agnostic, bold: best calibration performance, underlined: second best. Calibrator is not available (N/A) for uncalibrated models. \gls{CA} \gls{IR} provides a good balance of \gls{AP} and calibration performance. The results are presented on COCO \textit{minitest}. }
    \label{tab:app_ensemble_accuracy}
    \scalebox{0.8}{
    \begin{tabular}{c|c|c||c|c|c||c|c|c||c|c|c|c|c|c} 
    \toprule
    \midrule
    Cal.&Class&Calibrator& \multicolumn{3}{c|}{AP}&\multicolumn{3}{c|}{LaECE}&\multicolumn{3}{c|}{LaACE}&\multicolumn{3}{c}{LaMCE} \\ 
    Type&Type&&RS R-CNN&ATSS&PAA&RS R-CNN&ATSS&PAA&RS R-CNN&ATSS&PAA&RS R-CNN&ATSS&PAA\\ \midrule
    &N/A&N/A&$42.4$&$43.1$&$43.2$&$13.79$&$0.20$&$3.39$&$\mathbf{5.18}$&$25.22$&$9.73$&$35.37$&$42.53$&$\underline{19.80}$\\ \cline{2-15}
    &\multirow{2}{*}{CW}&LR&\cellcolor{bubblegum} $26.4$&\cellcolor{bubblegum} $42.0$&\cellcolor{bubblegum}$37.1$&$0.44$&$0.12$&$\underline{0.22}$&$25.12$&$10.13$&$28.95$&$61.11$&$\underline{24.79}$&$60.05$\\ 
    Early&&IR&\cellcolor{bubblegum} $41.9$&\cellcolor{bubblegum} $42.8$&\cellcolor{bubblegum} $42.5$&$\mathbf{0.03}$&$\mathbf{0.02}$&$1.39$&$\underline{5.23}$&$\mathbf{5.40}$&$\mathbf{5.26}$&$\underline{25.10}$&$25.38$&$26.60$\\ 
    Cal.&\multirow{2}{*}{CA}&LR&\cellcolor{inchworm}$42.4$&\cellcolor{bubblegum}$42.8$&\cellcolor{inchworm}$43.2$&$0.65$&$0.17$&$0.37$&$27.48$&$7.86$&$25.49$&$68.81$&$23.07$&$54.03$\\
    &&IR&\cellcolor{inchworm}$42.4$&\cellcolor{inchworm}$43.1$&\cellcolor{inchworm}$43.2$&$\underline{0.14}$&$\underline{0.10}$&$\mathbf{0.14}$&$5.86$&$\underline{6.43}$&$\underline{6.70}$&$\mathbf{14.08}$&$\mathbf{15.59}$&$\mathbf{18.15}$\\ \midrule \midrule
    &N/A&N/A&$42.4$&$43.1$&$43.2$&$36.45$&$5.01$&$11.23$&$29.30$&$17.48$&$15.79$&$45.42$&$40.00$&$\underline{32.33}$\\ \cline{2-15}
    &\multirow{2}{*}{CW}&LR&\cellcolor{inchworm}$42.4$&\cellcolor{inchworm}$43.1$&\cellcolor{inchworm}$43.2$&$4.36$&$\underline{2.69}$&$\underline{1.39}$&$14.19$&$\mathbf{9.03}$&$12.01$&$40.20$&$\mathbf{29.51}$&$33.76$\\ 
    Late&&IR&\cellcolor{bubblegum}$41.8$&\cellcolor{bubblegum}$42.5$&\cellcolor{bubblegum}$42.4$&$\mathbf{1.56}$&$\mathbf{2.35}$&$\mathbf{1.21}$&$\underline{9.21}$&$9.81$&$\mathbf{9.24}$&$38.43$&$40.34$&$37.84$\\ 
    Cal.&\multirow{2}{*}{CA}&LR&\cellcolor{inchworm}$42.4$&\cellcolor{inchworm}$43.0$&\cellcolor{inchworm}$43.2$&$5.83$&$4.46$&$1.63$&$13.86$&$\underline{9.46}$&$11.88$&$\underline{37.79}$&$\underline{29.59}$&$\mathbf{29.91}$\\ 
    &&IR&\cellcolor{inchworm}$42.3$&\cellcolor{inchworm}$43.1$&\cellcolor{inchworm}$43.2$&$\underline{3.15}$&$4.51$&$1.62$&$\mathbf{8.93}$&$9.51$&$\underline{9.61}$&$\mathbf{35.72}$&$37.35$&$35.59$\\ 
    \midrule
    \bottomrule
    \end{tabular}
    }
\end{table*}

\begin{figure*}[t]
        \captionsetup[subfigure]{}
        \centering
        \begin{subfigure}[b]{0.32\textwidth}
            \includegraphics[width=\textwidth]{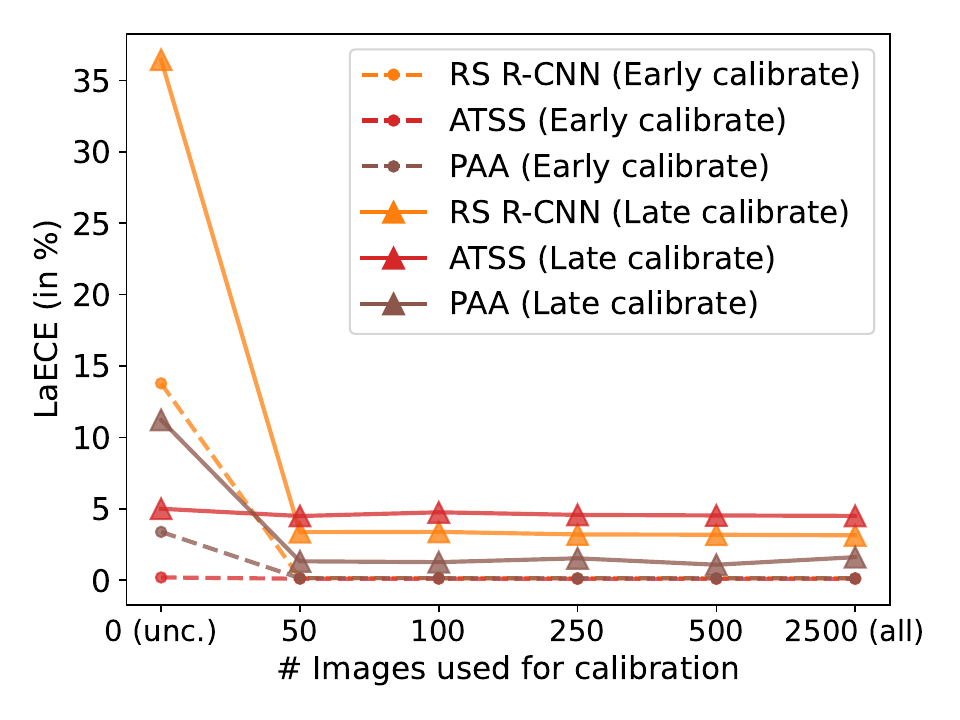}
            \caption{LaECE}
        \end{subfigure}
        \begin{subfigure}[b]{0.32\textwidth}
            \includegraphics[width=\textwidth]{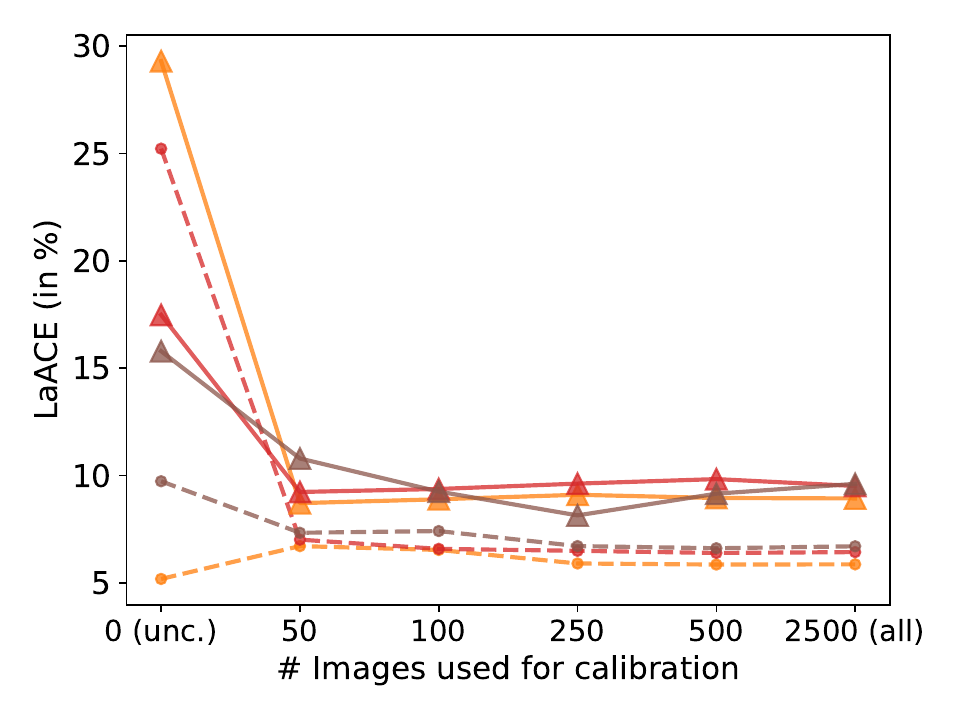}
            \caption{LaACE}
        \end{subfigure}
        \begin{subfigure}[b]{0.32\textwidth}
            \includegraphics[width=\textwidth]{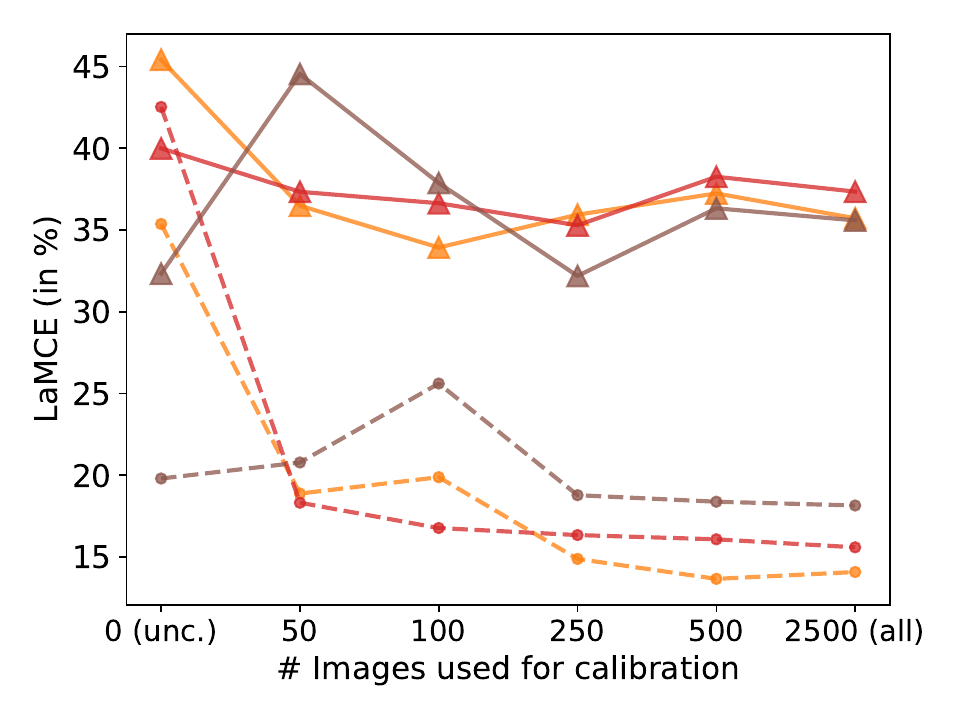}
            \caption{LaMCE}
        \end{subfigure}
        \caption{The effect of number of images on calibration using \gls{CA} \gls{IR}. We find it sufficient to use only 500 images for calibration.The results are presented using COCO \textit{minitest}.} 
        \label{fig:app_numimages}
\end{figure*}

\subsection{Further Ablation Experiments on \gls{CC}}
This section presents further ablation experiments by which we validate our design choices.
These include the validation of calibration method and the design choices in Refining \gls{NMS}.

\subsubsection{Further Details and Ablation on Calibration}

We use \gls{CA} \gls{IR} and \gls{LR} to calibrate the models given the final scores in \gls{CC}, i.e., late calibration.
More specifically, we use \gls{CA} \gls{IR} in all of the experiments unless otherwise explicitly specified.
One notable exception where we use the \gls{CA} \gls{LR} instead of \gls{IR} is the \gls{SOTA} experiment in \cref{tab:found_model_testdev} in which we observe that \gls{CA} \gls{LR} performs $0.1$ AP compared to \gls{CA} \gls{IR}.
Please note that, we obtain those calibrators on a held-out validation set consisting of only 500 images.
This section presents further  experiments to validate these design choices.

\paragraph{Validating the Calibrator} 
%
%
While calibrating the predictive distribution of the classifiers, commonly the accuracy of the classifier is preserved.
However, this might not be the case once the predictive confidence instead of the distribution is calibrated, which is the case in the common calibrators for object detection and in our case \cite{saod}.
This is mainly because the ranking among the detections can change unless the calibrator is a monotonically increasing function of the confidence. 
%
%
%
Therefore, we would ideally expect a calibrator to improve the calibration by at least preserving the accuracy of the detector. 
To ensure that, we investigate \gls{LR} and \gls{IR} both \gls{CA} and \gls{CW} on late as well as early calibration in \cref{tab:app_ensemble_accuracy}.
Here, we obtain the calibrators on COCO \textit{minival} and report the calibration error and accuracy on COCO \textit{minitest}.
We observe in the red cells in \cref{tab:app_ensemble_accuracy} that all calibrators, except \gls{CA} \gls{IR} and \gls{CA} \gls{IR}, decrease \gls{AP} especially for early calibration.
Furthermore, these calibrators improve \gls{LaECE} in all cases.
%
%
These observations on accuracy and calibration led us to choose \gls{CA} \gls{IR} and \gls{CA} \gls{LR} while calibrating the single models in \gls{CC}.

%

\paragraph{500 Images are Sufficient for calibration in \gls{CC}} 
Equipped with the aforementioned insights, we then investigate the sufficient number of images for calibration to enable \glspl{MoE} using \gls{CC}.
Our aim here is to determine the cardinality of the held-out validation set to properly calibrate the models,  which can help the practitioners to avoid reserving redundant data for this held-out validation set.
Following the literature \cite{calibration,saod}, we obtain calibrators on a held-out validation set (COCO minival) and then report the results on the test set (COCO minitest).
Furthermore, we obtain several calibrators on both early and late calibration settings by using different number of images.
\cref{fig:app_numimages} presents how \gls{LaECE}, \gls{LaACE} and \gls{LaMCE} change when the cardinality of the hold-out validation set changes on three different detectors.
We observe in general that calibration errors drop significantly even when only 50 images are used to learn the calibrators.
Through introducing more images, while we do not observe a notable gain in \gls{LaECE} (\cref{fig:app_numimages}(a)), \gls{LaACE} (\cref{fig:app_numimages}(b)) and \gls{LaMCE} (\cref{fig:app_numimages}(c)) continue to improve especially for early calibration, implying the necessity for these calibration errors.
Overall, as we have not observed a notable gain after 500 images, we keep 500 images on the held-out validation set while training the calibrators.
A noteworthy point is that 500 images is a small number images compared to large training sets in object detection, demonstrating the ease of applicability of using \gls{CC}.

\begin{table*}[t]
    \small
    \centering
    \caption{Further experiments comparing early and late calibration.  Standard NMS is used for both of the methods. While both approaches improve single models, late calibration performs slightly better. The results are presented on COCO \textit{minitest}.}
    \label{tab:app_ensembles_cal_2}
    \scalebox{1.0}{
    \begin{tabular}{c||c|c|c||c|c|c||c|c|c} 
    \toprule
    \midrule
    Calibration Type&RS R-CNN&ATSS&PAA&$\mathrm{AP}$&$\mathrm{AP_{50}}$&$\mathrm{AP_{75}}$&$\mathrm{AP_{S}}$&$\mathrm{AP_{M}}$&$\mathrm{AP_{L}}$ \\ \midrule
    N/A&\cmark& & &$42.4$&$62.1$&$46.2$&$26.8$&$46.3$&$56.9$\\
    (Single& &\cmark& &$43.1$&$61.5$&$47.1$&$27.8$&$47.5$&$54.2$\\
    Models)& & &\cmark&$43.2$&$60.8$&$47.1$&$27.0$&$47.0$&$57.6$\\ \midrule
    Early&\cmark&\cmark& &$43.9$&$63.5$&$47.9$&$28.6$&$47.9$&$56.9$\\
    Late&\cmark&\cmark& &$44.1$&$63.0$&$48.4$&$28.5$&$48.4$&$56.8$\\\midrule
    
    Early&\cmark& &\cmark&$43.8$&$63.3$&$47.7$&$28.1$&$47.8$&$57.6$\\
    Late&\cmark& &\cmark&$44.0$&$62.7$&$47.9$&$28.2$&$48.2$&$58.1$\\\midrule
    
    Early& &\cmark&\cmark&$44.3$&$63.1$&$47.8$&$28.8$&$48.1$&$56.7$\\
    Late& &\cmark&\cmark&$44.4$&$62.5$&$48.5$&$29.2$&$48.5$&$57.3$\\\midrule
    
    Early&\cmark&\cmark&\cmark&$44.5$&$\mathbf{63.7}$&$48.4$&$29.0$&$48.5$&$57.5$\\ 
    Late&\cmark&\cmark&\cmark&$\mathbf{44.7}$&$63.1$&$\mathbf{48.9}$&$\mathbf{29.2}$&$\mathbf{49.0}$&$\mathbf{58.2}$\\ 
    \midrule
    \bottomrule
    \end{tabular}
     }
\end{table*}

\paragraph{Comparison of Early and Late Calibration}
Here, we provide additional insights on early and late calibration.
In \cref{tab:app_ensembles_cal_2}, we can see that while both approaches improve single models, late calibration performs slightly better than early calibration consistently.
Furthermore, in practical terms, late calibration is significantly simpler as
%
the number of confidence scores obtained before post-processing is significantly larger than those obtained after postprocessing (i.e., final detections).
To illustrate this more concretely, RS R-CNN outputs 1K proposals for each image as raw detections, thus resulting with a single image containing 80K raw confidence scores for the COCO dataset and more than 1M scores for LVIS as each proposal has a score for each class.
In addition, a very large amount of these raw detections do not even overlap with any objects, 
complicating the problem due to this imbalanced nature of the data.
%
Furthermore, the number of raw confidence scores is significantly larger for one-stage detectors (ATSS and PAA in our case) as such detectors make predictions directly from a very large number of anchors; making early calibration even more impractical for them\footnote{To keep this number manageable, we use top-1000 detections predicted from each pyramid level for ATSS and PAA for early calibration.}.
On the other hand, we use only top-100 detections in COCO and top-300 detections in LVIS for each image following the evaluation specification of these datasets.
Thereby resulting in more practical scenarios with significantly smaller number of detections for late calibration compared to early.
Consequently, considering its slight accuracy gain as well as simplicity, we prefer late calibration over early to obtain \glspl{MoE} in \gls{CC}.

\blockcomment{
\paragraph{Effect of Refining NMS } \cref{fig:app_numimages}

\begin{table}[t]
    \small
    \setlength{\tabcolsep}{0.25em}
    \centering
    \caption{Effect of Refining NMS, which is a combination of Soft NMS and Score Voting. Using Refining NMS for Late \gls{CC} for clustering boosts the \gls{MoE} performance by $0.8$ AP, resulting in $2.3$ AP gain in total.}
    \label{tab:app_ablation}
    \scalebox{1.0}{
    \begin{tabular}{c|c||c|c|c||c|c|c||c|c|c} 
    \toprule
    \midrule
    Calibration&Clustering&RS R-CNN&ATSS&PAA&$\mathrm{AP}$&$\mathrm{AP_{50}}$&$\mathrm{AP_{75}}$&$\mathrm{AP_{S}}$&$\mathrm{AP_{M}}$&$\mathrm{AP_{L}}$ \\ \midrule
    \xmark&N/A&\cmark& & &$42.4$&$62.1$&$46.2$&$26.8$&$46.3$&$56.9$\\
    \xmark&N/A& &\cmark& &$43.1$&$61.5$&$47.1$&$27.8$&$47.5$&$54.2$\\
    \xmark&N/A& & &\cmark&$43.2$&$60.8$&$47.1$&$27.0$&$47.0$&$57.6$\\ \midrule
    \xmark&NMS&\cmark&\cmark&\cmark&$43.4$&$62.5$&$47.1$&$27.3$&$47.3$&$58.0$\\
    \xmark&Linear Soft NMS&\cmark&\cmark&\cmark&$43.4$&$62.5$&$47.1$&$27.3$&$47.3$&$58.0$\\
    \xmark&Gaussian Soft NMS&\cmark&\cmark&\cmark&$42.9$&$62.0$&$46.5$&$26.7$&$46.5$&$57.3$\\ 
    \xmark&Refining NMS&\cmark&\cmark&\cmark&$44.4$&$62.6$&$48.2$&$27.8$&$48.4$&$\mathbf{59.3}$\\
    & & & &&$\imp{1.2}$&$\imp{0.5}$&$\imp{1.1}$&$\textcolor{red}{0.0}$&$\imp{0.9}$&$\imp{1.7}$ \\
     \midrule
    \cmark&NMS&\cmark&\cmark&\cmark&$44.7$&$63.1$&$48.9$&$29.2$&$49.0$&$58.2$\\  
    \cmark&Linear Soft NMS&\cmark&\cmark&\cmark&$44.8$&$63.1$&$49.1$&$29.3$&$49.1$&$58.3$\\
    \cmark&Gaussian Soft NMS&\cmark&\cmark&\cmark&$44.8$&$\mathbf{63.3}$&$48.9$&$28.8$&$48.7$&$58.5$\\
    \cmark&Refining NMS&\cmark&\cmark&\cmark&$\mathbf{45.5}$&$63.2$&$\mathbf{50.0}$&$\mathbf{29.7}$&$\mathbf{49.7}$&$\mathbf{59.3}$\\ 
         & & & &&$\imp{2.3}$&$\imp{1.1}$&$\imp{2.9}$&$\imp{1.9}$&$\imp{2.2}$&$\imp{1.7}$ \\

    \midrule
    \bottomrule
    \end{tabular}
     }
\end{table}

\begin{table}
    \small
    \setlength{\tabcolsep}{0.25em}
    \centering
    \caption{Ensembling detectors. Refining NMS: Soft NMS + Score Voting. Observations: (1) All ensembles perform poor without calibration. Also soft nms does not have benefit. (2) Thanks to calibration, the performance improves and using soft nms boosts the performance more. Compared to best individual model, the gain is $1.7$AP. }
    \label{tab:app_ensembles_cal}
    \scalebox{1.0}{
    \begin{tabular}{c|c||c|c|c||c|c|c||c|c|c} 
    \toprule
    \midrule
    Calibrated&Refining NMS&RS R-CNN&ATSS&PAA&AP&AP50&AP75&APS&APM&APL \\ \midrule
    \xmark&\xmark&\cmark& & &$42.4$&$62.1$&$46.2$&$26.8$&$46.3$&$56.9$\\
    \xmark&\xmark& &\cmark& &$43.1$&$61.5$&$47.1$&$27.8$&$47.5$&$54.2$\\
    \xmark&\xmark& & &\cmark&$43.2$&$60.8$&$47.1$&$27.0$&$47.0$&$57.6$\\ \midrule
    Pre-NMS Uncalibrated& & & & & & & & & & \\        
    \xmark&\xmark&\cmark&\cmark& &$42.4$&$62.1$&$46.2$&$26.8$&$46.3$&$56.9$\\
    \xmark&\xmark&\cmark& &\cmark&$43.3$&$62.6$&$46.8$&$27.2$&$47.1$&$57.8$\\
    \xmark&\xmark& &\cmark&\cmark&$43.1$&$61.3$&$46.5$&$26.8$&$46.8$&$57.2$ \\
    \xmark&\xmark&\cmark&\cmark&\cmark&$43.3$&$62.6$&$46.8$&$27.2$&$47.1$&$57.8$\\ 
    \xmark&\cmark&\cmark&\cmark&\cmark&$42.9$&$62.0$&$46.5$&$26.8$&$46.4$&$57.3$\\ \midrule 
    Pre-NMS CA IR& & & & & & & & & & \\        
    \cmark&\xmark&\cmark&\cmark& &$43.9$&$63.5$&$47.9$&$28.6$&$47.9$&$56.9$\\
    \cmark&\xmark&\cmark& &\cmark&$43.8$&$63.3$&$47.7$&$28.1$&$47.8$&$57.6$\\
    \cmark&\xmark& &\cmark&\cmark&$44.3$&$63.1$&$47.8$&$28.8$&$48.1$&$56.7$\\
    \cmark&\xmark&\cmark&\cmark&\cmark&$44.5$&$\mathbf{63.7}$&$48.4$&$\mathbf{29.0}$&$48.5$&$57.5$\\ 
    \cmark&\cmark&\cmark&\cmark&\cmark&$\mathbf{44.9}$&$63.5$&$\mathbf{49.3}$&$\mathbf{29.0}$&$\mathbf{48.9}$&$\mathbf{58.3}$\\ \midrule
    Post-NMS Uncalibrated& & & & & & & & & & \\    
    \xmark&\xmark&\cmark&\cmark& &$42.4$&$62.1$&$46.3$&$26.8$&$46.3$&$56.9$\\
    \xmark&\xmark&\cmark& &\cmark&$43.4$&$62.5$&$47.1$&$27.3$&$47.3$&$58.0$ \\
    \xmark&\xmark& &\cmark&\cmark&$43.3$&$60.9$&$47.2$&$27.1$&$47.2$&$57.6$ \\
    \xmark&\xmark&\cmark&\cmark&\cmark&$43.4$&$62.5$&$47.1$&$27.3$&$47.3$&$58.0$\\ 
    \xmark&\cmark&\cmark&\cmark&\cmark&$43.1$&$61.2$&$46.9$&$26.7$&$46.8$&$57.9$\\ \midrule
    Post-NMS CW LR& & & & & & & & & & \\    
    \cmark&\xmark&\cmark&\cmark& &$43.0$&$61.4$&$46.9$&$28.1$&$47.2$&$54.4$\\
    \cmark&\xmark&\cmark& &\cmark&$42.7$&$61.8$&$46.7$&$27.1$&$46.5$&$57.6$\\
    \cmark&\xmark& &\cmark&\cmark&$42.6$&$60.7$&$46.5$&$26.6$&$47.2$&$53.9$\\
    \cmark&\xmark&\cmark&\cmark&\cmark&$43.0$&$61.3$&$47.1$&$28.2$&$47.4$&$54.6$\\ 
    \midrule
    Post-NMS CA LR& & & & & & & & & & \\    
    \cmark&\xmark&\cmark&\cmark& &$43.7$&$62.5$&$47.7$&$28.7$&$48.2$&$55.2$\\
    \cmark&\xmark&\cmark& &\cmark&$43.5$&$63.0$&$47.7$&$27.9$&$47.5$&$57.8$\\
    \cmark&\xmark& &\cmark&\cmark&$44.4$&$62.5$&$48.3$&$29.0$&$48.3$&$57.3$\\
    \cmark&\xmark&\cmark&\cmark&\cmark&$44.0$&$62.7$&$48.1$&$29.0$&$48.5$&$55.8$\\ 
    \midrule
    Post-NMS CA IR& & & & & & & & & & \\    
    \cmark&\xmark&\cmark&\cmark& &$44.1$&$63.0$&$48.4$&$28.5$&$48.4$&$56.8$\\
    \cmark&\xmark&\cmark& &\cmark&$44.0$&$62.7$&$47.9$&$28.2$&$48.2$&$58.1$\\
    \cmark&\xmark& &\cmark&\cmark&$44.4$&$62.5$&$48.5$&$29.2$&$48.5$&$57.3$\\
    \cmark&\xmark&\cmark&\cmark&\cmark&$44.7$&$63.1$&$48.9$&$29.2$&$49.0$&$58.2$\\ 
    \cmark& 0.75, 0.025&\cmark&\cmark&\cmark&$\mathbf{45.1}$&$63.1$&$\mathbf{49.4}$&$\mathbf{29.0}$&$\mathbf{49.1}$&$\mathbf{59.4}$\\
    \cmark& 0.60, 0.025&\cmark&\cmark&\cmark&$\mathbf{45.2}$&$63.4$&$\mathbf{49.4}$&$\mathbf{29.0}$&$\mathbf{49.1}$&$\mathbf{59.4}$\\
    \cmark& 0.60, 0.040&\cmark&\cmark&\cmark&$\mathbf{45.2}$&$63.4$&$\mathbf{49.5}$&$\mathbf{29.1}$&$\mathbf{49.1}$&$\mathbf{59.4}$\\
    \cmark& 0.60, 0.050&\cmark&\cmark&\cmark&$\mathbf{45.2}$&$63.4$&$\mathbf{49.5}$&$\mathbf{29.1}$&$\mathbf{49.1}$&$\mathbf{59.4}$\\
    \cmark& 0.60, 0.075&\cmark&\cmark&\cmark&$\mathbf{45.2}$&$63.4$&$\mathbf{49.5}$&$\mathbf{29.1}$&$\mathbf{49.1}$&$\mathbf{59.4}$\\
    \cmark& 0.75, 0.040&\cmark&\cmark&\cmark&$\mathbf{45.2}$&$63.1$&$\mathbf{49.5}$&$\mathbf{29.1}$&$\mathbf{49.2}$&$\mathbf{59.3}$\\
    \cmark& 0.75, 0.050&\cmark&\cmark&\cmark&$\mathbf{45.1}$&$63.1$&$\mathbf{49.4}$&$\mathbf{29.1}$&$\mathbf{49.2}$&$\mathbf{59.3}$\\
    \midrule
    \bottomrule
    \end{tabular}
     }
\end{table}
}

\begin{table*}
    \small
    \centering
    \caption{Comparison of different calibration methods to obtain \glspl{MoE}. \gls{CA} \gls{IR} performs better than other methods. The results are presented on COCO \textit{minitest}.}
    \label{tab:app_ensembles_cal} 
    \scalebox{1.0}{
    \begin{tabular}{c||c|c|c||c|c|c||c|c|c} 
    \toprule
    \midrule
    Calibrated&RS R-CNN&ATSS&PAA&AP&AP50&AP75&APS&APM&APL \\ \midrule
    N/A&\cmark& & &$42.4$&$62.1$&$46.2$&$26.8$&$46.3$&$56.9$\\
    (Single& &\cmark& &$43.1$&$61.5$&$47.1$&$27.8$&$47.5$&$54.2$\\
    Models)& & &\cmark&$43.2$&$60.8$&$47.1$&$27.0$&$47.0$&$57.6$\\ \midrule
    
    \xmark&\cmark&\cmark& &$42.4$&$62.1$&$46.3$&$26.8$&$46.3$&$56.9$\\
    \gls{CW} \gls{LR}&\cmark&\cmark& &$43.0$&$61.4$&$46.9$&$28.1$&$47.2$&$54.4$\\
    \gls{CA} \gls{LR}&\cmark&\cmark& &$43.7$&$62.5$&$47.7$&$28.7$&$48.2$&$55.2$\\
    \gls{CA} \gls{IR}&\cmark&\cmark& &$44.1$&$63.0$&$48.4$&$28.5$&$48.4$&$56.8$\\\midrule
    \xmark&\cmark& &\cmark&$43.4$&$62.5$&$47.1$&$27.3$&$47.3$&$58.0$ \\
    \gls{CW} \gls{LR}&\cmark& &\cmark&$42.7$&$61.8$&$46.7$&$27.1$&$46.5$&$57.6$\\
    \gls{CA} \gls{LR}&\cmark& &\cmark&$43.5$&$63.0$&$47.7$&$27.9$&$47.5$&$57.8$\\
    \gls{CA} \gls{IR}&\cmark& &\cmark&$44.0$&$62.7$&$47.9$&$28.2$&$48.2$&$58.1$\\\midrule
    \xmark&&\cmark&\cmark&$43.3$&$60.9$&$47.2$&$27.1$&$47.2$&$57.6$ \\
    \gls{CW} \gls{LR}& &\cmark&\cmark&$42.6$&$60.7$&$46.5$&$26.6$&$47.2$&$53.9$\\
    \gls{CA} \gls{LR}& &\cmark&\cmark&$44.4$&$62.5$&$48.3$&$29.0$&$48.3$&$57.3$\\
    \gls{CA} \gls{IR}&&\cmark&\cmark&$44.4$&$62.5$&$48.5$&$29.2$&$48.5$&$57.3$\\\midrule
    \xmark&\cmark&\cmark&\cmark&$43.4$&$62.5$&$47.1$&$27.3$&$47.3$&$58.0$\\ 
    \gls{CW} \gls{LR}&\cmark&\cmark&\cmark&$43.0$&$61.3$&$47.1$&$28.2$&$47.4$&$54.6$\\
    \gls{CA} \gls{LR}&\cmark&\cmark&\cmark&$44.0$&$62.7$&$48.1$&$29.0$&$48.5$&$55.8$\\ 
    \gls{CA} \gls{IR}&\cmark&\cmark&\cmark&$44.7$&$63.1$&$48.9$&$29.2$&$49.0$&$58.2$\\ 
    \midrule
    \bottomrule
    \end{tabular}
     }
\end{table*}

\paragraph{Effect of Different Calibration Methods on \glspl{MoE}}
While we choose \gls{CA} \gls{IR} as our calibration method in \gls{CC}, here we present how different calibration methods perform in obtaining \glspl{MoE}.
Specifically, we use late calibration with \gls{CA} \gls{LR} and \gls{CW} \gls{LR} as these two methods also preserve the accuracy of single models as shown in \cref{tab:app_ensembles_cal}.
\cref{tab:app_ensembles_cal} presents the results where we can see that \gls{CA} calibrators perform better than \gls{CW} \gls{LR}.
Also, while \gls{CA} \gls{LR} obtains on par performance with \gls{CA} \gls{IR} while combining ATSS and PAA, it performs worse once RS R-CNN is in the mixture.
This might be because the calibration error of \gls{CA} \gls{LR} is higher than \gls{CA} \gls{IR} (for late calibration) in terms of all calibration measures as shown in \cref{tab:app_ensembles_cal}).
%
%

\begin{table}[t]
    \centering
    
    \scalebox{0.85}{
    \begin{tabular}{c|c|c|c||c} 
    \toprule
    \midrule
    \gls{CC}&\multirow{2}{*}{Calibration}&\multicolumn{2}{c||}{Refining NMS}&\multirow{2}{*}{$\mathrm{AP}$}\\
    Pipeline& &Soft NMS&Score Voting&\\ \midrule
    Early&\xmark&\xmark&\xmark&$43.3$\\
    Early&\cmark&\xmark&\xmark&$44.5$\\
    Late &\cmark&\xmark&\xmark&$44.7$\\
    Late &\cmark&\cmark&\xmark&$44.8$\\
    Late &\xmark&\cmark&\xmark&$43.4$\\
    Late &\xmark&\cmark&\cmark&$44.4$\\
    Late &\cmark&\cmark&\cmark&$\mathbf{45.5}$\\
    \midrule
    \bottomrule
    \end{tabular}
     }
     \caption{Ablation analysis of \gls{CC}. We use the \gls{MoE} combining RS R-CNN, ATSS and PAA on COCO \textit{mini-test}}. 
    \label{tab:app_ablation_}
\end{table}

\paragraph{Ablation of \gls{CC} with Early Calibration}
\cref{tab:app_ablation_} presents a more detailed version of the \cref{tab:ablation} included in the paper.
In this version, we also include early calibration, which performs similar with late calibration as shown in \cref{tab:app_ablation_}.

\begin{table}
    \small
    \centering
    \caption{Sensitivity of Vanilla \gls{MoE} and \gls{CC} to different configurations of Soft NMS. The results are presented on COCO mini-test and 500 validation images that we used to train the calibrators for LVIS. We report box AP for COCO and mask AP for LVIS.}
    \label{tab:app_ablation}
    \scalebox{1.0}{
    \begin{tabular}{c|c||c|c} 
    \toprule
    \midrule
    Method&Soft NMS&COCO&LVIS\\ \midrule
    \multirow{7}{*}{Vanilla MoE}& \xmark &$\mathbf{43.4}$&$37.5$\\
     & Linear, $\mathrm{IoU}_{NMS}=0.65$ &$\mathbf{43.4}$&$37.5$\\
     & Gaussian, $\sigma_{NMS}=0.20$ &$41.6$&$\mathbf{37.9}$\\
     & Gaussian, $\sigma_{NMS}=0.40$ &$42.1$&$37.8$\\
     & Gaussian, $\sigma_{NMS}=0.60$ &$42.4$&$\mathbf{37.9}$\\
     & Gaussian, $\sigma_{NMS}=0.80$ &$42.7$&$37.8$\\
     & Gaussian, $\sigma_{NMS}=1.00$ &$42.9$&$37.9$\\ \midrule
    \multirow{7}{*}{\gls{CC}}& \xmark &$44.7$&$39.8$\\
     & Linear, $\mathrm{IoU}_{NMS}=0.65$ &$\mathbf{44.8}$&$39.9$\\
     & Gaussian, $\sigma_{NMS}=0.20$ &$43.7$&$40.6$\\
     & Gaussian, $\sigma_{NMS}=0.40$ &$44.4$&$\mathbf{40.8}$\\
     & Gaussian, $\sigma_{NMS}=0.60$ &$\mathbf{44.8}$&$40.6$\\
     & Gaussian, $\sigma_{NMS}=0.80$ &$44.7$&$40.2$\\
     & Gaussian, $\sigma_{NMS}=1.00$ &$44.6$&$39.9$\\ \midrule
    \bottomrule
    \end{tabular}
     }
\end{table}

\subsubsection{Sensitivity of \gls{CC} to Design Choices in Refining \gls{NMS}}
As introduced in App \ref{app:cc}, Refining NMS combines Soft NMS and Score Voting. 
Specifically, Soft NMS can be linear or gaussian; furthermore both Soft NMS (either linear or gaussian) and Score Voting have hyper-parameters.
Here, we investigate the sensitivity of \gls{CC} to such design choices using Soft NMS as an example using RS R-CNN, ATSS and PAA for COCO; and the setting described in \cref{sec:exp} for LVIS.
We can easily see in \cref{tab:app_ablation} that Vanilla \gls{MoE} does not benefit properly from Soft NMS without calibration.
For example, there is no gain for Linear Soft NMS, the performance degrades for the Gaussian Soft NMS on COCO and the gain is only $0.4$ for LVIS.
This is expected as a single hyper-parameter to reconciliate the scores all detectors might not be sufficient especially for the Gaussian Soft \gls{NMS}.
On the other hand, after calibration, we consistently see the gains for our \gls{CC}: \gls{CC} benefits slightly on COCO dataset both for linear and gaussian cases; and besides, the gain on LVIS is $1.0$ mask AP.
This is because, the scores are compatible for each detector after calibration and a single hyperparameter allows Soft NMS to properly adjust the scores from different detectors.
We choose Linear Soft NMS on COCO resulting in the best results for Vanilla \gls{MoE} and \gls{CC}.
For LVIS, we use Gaussian Soft NMS with $\sigma_{NMS}=0.40$ for \gls{CC}.
In a similar way, we validate the hyper-parameter of Score Voting as $0.04$.

\blockcomment{
\begin{figure}[]
\centering
\includegraphics[width=\textwidth]{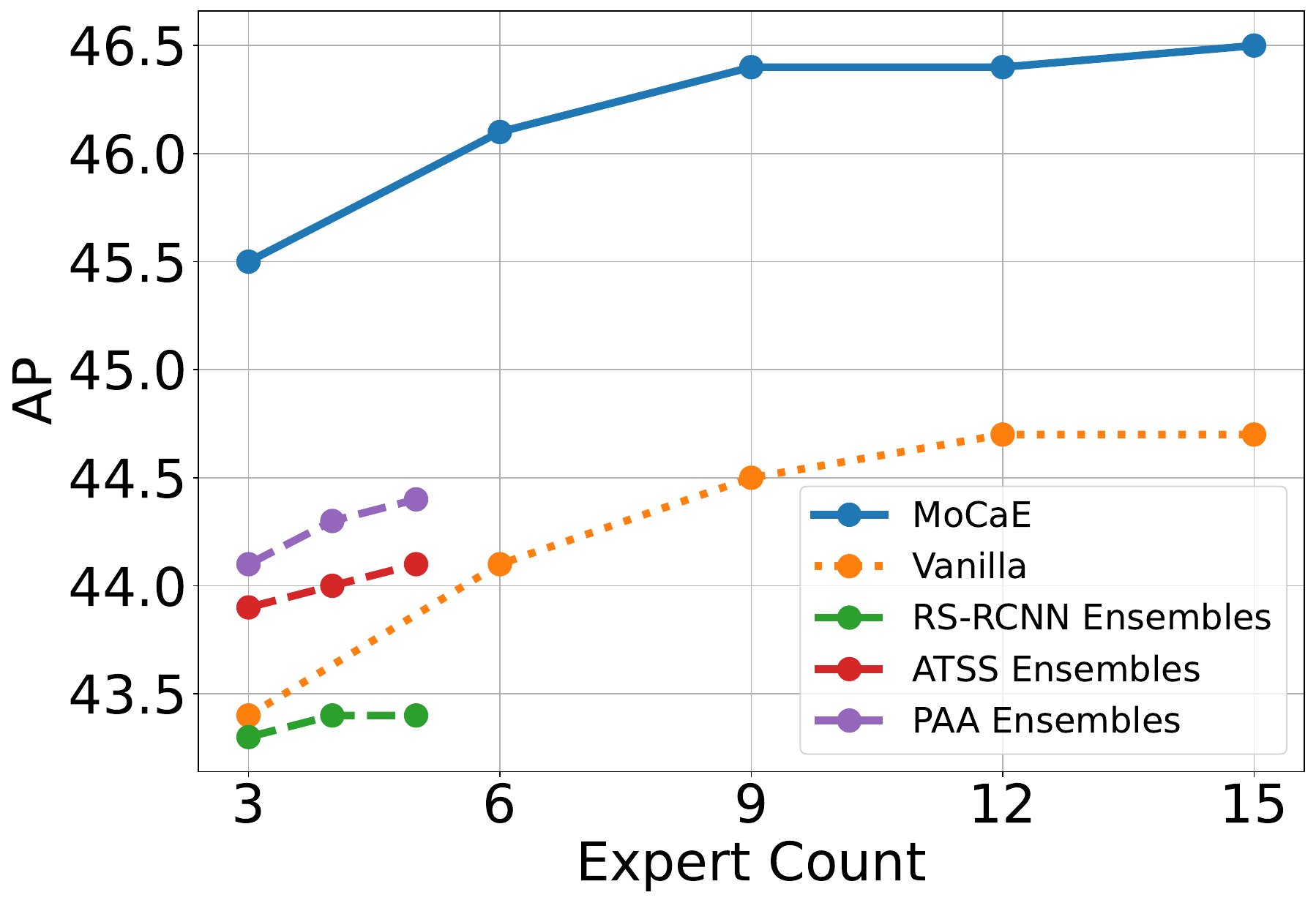}
\caption{$\mathrm{AP}$ scores of ensembles, Vanilla \gls{MoE} and \gls{CC} of different expert counts on COCO \textit{mini-test} dataset. \gls{CC} consistently outperforms Vanilla \gls{MoE} and other ensembles. Furthermore, its performance increases up to 1 $\mathrm{AP}$ as more experts are added.}

\label{tab:different_expert_counts_ap_scores}
\end{figure}

\subsubsection{Using More Components in \gls{CC}}  
We observed in the previous sections that combining 3 detectors are useful to improve the performance of \gls{CC}.
We investigate combining more detectors in \cref{tab:different_expert_counts_ap_scores}, in which we combine all 15 detectors (3 different types of detectors and 5 components from each) that we trained to obtain ensembles. 
Even if the comp
}

\subsection{Further Details and Analyses on Deep Ensembles}
This section provides further details and analyses on \glspl{DE}.

\begin{table}[t]
    \small
    \setlength{\tabcolsep}{0.4em}
    \centering
    \caption{Single model performance of the detectors that we used in \glspl{DE}. While obtaining \glspl{MoE}, we combine ``Model 1'' of different types of detectors.}
    \label{tab:app_ensemble_moe}
    \scalebox{0.9}{
    \begin{tabular}{c||c|c|c||c|c|c} 
    \toprule
    \midrule
    Model&$\mathrm{AP}$&$\mathrm{AP_{50}}$&$\mathrm{AP_{75}}$&$\mathrm{AP_{S}}$&$\mathrm{AP_{M}}$&$\mathrm{AP_{L}}$\\ \midrule
    RS R-CNN (Model 1)&$42.4$&$62.1$&$46.2$&$26.8$&$46.3$&$56.9$\\ 
    RS R-CNN (Model 2)&$42.6$&$62.7$&$46.7$&$27.3$&$46.5$&$55.8$\\ 
    RS R-CNN (Model 3)&$42.6$&$62.6$&$46.5$&$27.2$&$46.8$&$56.1$\\ 
    RS R-CNN (Model 4)&$42.6$&$62.3$&$46.1$&$27.9$&$46.1$&$55.9$\\ 
    RS R-CNN (Model 5)&$42.2$&$62.8$&$45.8$&$26.7$&$46.5$&$55.7$\\ \midrule
    ATSS (Model 1)&$43.1$&$61.5$&$47.1$&$27.8$&$47.5$&$54.2$\\ 
    ATSS (Model 2)&$43.3$&$61.5$&$47.5$&$28.9$&$47.8$&$55.0$\\ 
    ATSS (Model 3)&$43.3$&$61.5$&$47.0$&$28.9$&$47.9$&$55.8$\\ 
    ATSS (Model 4)&$43.0$&$61.2$&$46.8$&$29.0$&$47.4$&$54.6$\\ 
    ATSS (Model 5)&$43.3$&$61.5$&$47.5$&$28.2$&$47.7$&$54.8$\\ \midrule
    PAA (Model 1)&$43.2$&$60.8$&$47.1$&$27.0$&$47.0$&$57.6$\\ 
    PAA (Model 2)&$43.5$&$61.0$&$47.3$&$27.5$&$47.7$&$57.7$\\ 
    PAA (Model 3)&$43.4$&$61.1$&$47.2$&$27.7$&$47.6$&$57.8$\\ 
    PAA (Model 4)&$43.6$&$61.4$&$47.2$&$27.6$&$47.6$&$57.6$\\ 
    PAA (Model 5)&$43.6$&$61.3$&$47.2$&$27.9$&$47.9$&$58.0$\\
    \midrule
    \bottomrule
    \end{tabular}
     }
\end{table}

\begin{table}[t]
    \small
    \setlength{\tabcolsep}{0.4em}
    \centering
    \caption{The effect of increasing the components and using calibration on \glspl{DE}. Increasing the components improves the performance while calibration does not have a notable effect on performance for \glspl{DE} unlike their importance for \glspl{MoE}.}
    \label{tab:app_ensemble_moe2}
    \scalebox{0.9}{
    \begin{tabular}{c|c||c|c|c||c|c|c} 
    \toprule
    \midrule
    Model&Calibration&$\mathrm{AP}$&$\mathrm{AP_{50}}$&$\mathrm{AP_{75}}$&$\mathrm{AP_{S}}$&$\mathrm{AP_{M}}$&$\mathrm{AP_{L}}$\\ \midrule
    \multirow{2}{*}{PAA $\times$ 2}&\xmark&$43.8$&$61.4$&$47.6$&$28.2$&$48.3$&$58.3$\\
    &\cmark&$43.9$&$61.4$&$47.7$&$28.0$&$48.3$&$58.6$\\ \cdashlinelr{1-8}
    \multirow{2}{*}{PAA $\times$ 3}&\xmark&$44.1$&$61.5$&$48.0$&$28.8$&$48.7$&$58.8$\\
    &\cmark&$44.1$&$61.6$&$48.0$&$28.7$&$48.7$&$58.9$\\ \cdashlinelr{1-8}
    \multirow{2}{*}{PAA $\times$ 4}&\xmark&$44.3$&$61.7$&$48.3$&$28.9$&$48.8$&$59.1$\\
    &\cmark&$44.3$&$61.7$&$48.3$&$28.9$&$48.8$&$59.1$\\ \cdashlinelr{1-8}
    \multirow{2}{*}{PAA $\times$ 5}&\xmark&$44.4$&$61.7$&$48.6$&$29.1$&$49.0$&$59.3$\\
    &\cmark&$44.4$&$61.7$&$48.6$&$28.9$&$49.0$&$59.3$\\ 
    \midrule
    \bottomrule
    \end{tabular}
     }
\end{table}

\subsubsection{The Effect of Calibration on \glspl{DE}}
\glspl{DE} combine the same models that are trained from different initialization of the parameters.
Ideally, the expectation over the predictive distributions of the components in a \gls{DE} yields the prediction of the \gls{DE}.
This can be easily obtained for classifiers which predict a categorical distribution over the classes given an input image.
On the other hand, it is not straightforward to use \glspl{DE} for detectors as there is no clear way to associate detections from different detectors.
As a result, similar to \gls{CC}, we obtain \glspl{DE} by using late calibration as shown in \cref{fig:cc}(a), which turns out to be an effective method.
To see that, we first present the single model performance of the five different components comprising \glspl{DE} in \cref{tab:app_ensemble_moe}.
Then, from these single detectors, we obtain \glspl{DE} for PAA with and without calibration using the standard \gls{NMS}.
\cref{tab:app_ensemble_moe2} shows that this way of obtaining \glspl{DE} is effective as the performance increases when the number of components increases.
We observe that increasing the number of components improve the performance between $0.1-0.3$ AP.
On the other hand, as there is no incompatibility among different detectors in a \gls{DE}, the effect of calibration is not notable for \glspl{DE}.
Still, having observed that PAA with 2 components has a slightly better ($0.1$ AP) performance once calibrated, in our comparisons we use \glspl{DE} with calibration.

\begin{table*}
    \small
    \centering
    \caption{The effect of Refining \gls{NMS} on Deep Ensembles and Vanilla MoE.}
    \label{tab:app_ensemble_moe_}
    \scalebox{1.0}{
    \begin{tabular}{c|c||c|c|c||c|c|c} 
    \toprule
    \midrule
    Model Type&Detector&$\mathrm{AP}$&$\mathrm{AP_{50}}$&$\mathrm{AP_{75}}$&$\mathrm{AP_{S}}$&$\mathrm{AP_{M}}$&$\mathrm{AP_{L}}$\\ \midrule
    \multirow{3}{*}{Single Models}&RS R-CNN&$42.4$&$\underline{62.1}$&$46.2$&$26.8$&$46.3$&$56.9$\\
    &ATSS&$43.1$&$61.5$&$\underline{47.1}$&$\underline{27.8}$&$\underline{47.5}$&$54.2$\\ 
    &PAA&$\underline{43.2}$&$60.8$&$\underline{47.1}$&$27.0$&$47.0$&$\underline{57.6}$\\ \midrule
    \multirow{12}{*}{Deep Ensembles}
    &RS R-CNN $\times$ 3&$43.3$&$63.1$&$47.4$&$27.7$&$47.5$&$57.1$\\ 
    &ATSS $\times$ 3&$43.9$&$62.1$&$47.9$&$29.5$&$48.9$&$55.9$\\ 
    &PAA $\times$ 3&$44.1$&$61.7$&$47.9$&$28.6$&$48.6$&$58.9$\\ \cdashlinelr{2-8}
    &RS R-CNN $\times$ 3 with Ref. NMS&$44.2$&$63.0$&$48.5$&$28.2$&$48.4$&$58.5$\\ 
    &ATSS $\times$ 3 with Ref. NMS&$44.3$&$62.1$&$48.3$&$29.7$&$49.3$&$56.3$\\ 
    &PAA $\times$ 3 with Ref. NMS&$44.6$&$61.6$&$48.7$&$28.9$&$49.1$&\textbf{$59.5$}\\ \cdashlinelr{2-8}
    &RS R-CNN $\times$ 5&$43.4$&$63.0$&$47.7$&$28.0$&$47.5$&$57.0$\\ 
    &ATSS $\times$ 5&$44.1$&$62.3$&$48.4$&$29.4$&$49.0$&$56.3$\\ 
    &PAA $\times$ 5 &$44.4$&$62.0$&$48.4$&$28.9$&$49.0$&$59.2$\\  \cdashlinelr{2-8}
    &RS R-CNN $\times$ 5 with Ref. NMS&$44.5$&$63.0$&$49.1$&$28.8$&$48.6$&$58.4$\\ 
    &ATSS $\times$ 5 with Ref. NMS&$44.6$&$62.2$&$48.9$&$29.8$&$49.5$&$56.7$\\ 
    &PAA $\times$ 5 with Ref. NMS&$45.0$&$61.8$&$49.3$&$29.4$&$49.6$&$\textbf{59.9}$\\ \midrule
    \multirow{5}{*}{Mixtures of Experts}&Vanilla \gls{MoE} (RS R-CNN, ATSS, PAA)&$43.4$&$62.5$&$47.1$&$27.3$&$47.3$&$58.0$\\ 
    &Vanilla \gls{MoE} with Ref. NMS (RS R-CNN, ATSS, PAA)&$44.4$&$62.6$&$48.2$&$27.8$&$48.4$&$59.3$\\ \cdashlinelr{2-8}
    &\gls{CC} (ATSS and PAA) - Ours&$44.8$&$62.4$&$49.2$&$29.4$&$49.1$&$57.6$\\
    &\gls{CC} (RS R-CNN, ATSS, PAA) - Ours&$\textbf{45.5}$&$\textbf{63.2}$&$\textbf{50.0}$&$\textbf{29.7}$&$\textbf{49.7}$&$59.3$\\
    &&$\imp{2.3}$&$\imp{1.1}$&$\imp{2.9}$&$\imp{1.9}$&$\imp{2.2}$&$\imp{1.7}$\\
    \midrule
    \bottomrule
    \end{tabular}
     }
\end{table*}

\subsubsection{\glspl{DE} with Less Components and Refining \gls{NMS}}
In \cref{tab:ensemble_moe}, we compared \gls{CC} with \glspl{DE} with 5 components.
Note that in this case, the \glspl{DE} have more components, implying a higher number of parameters compared to our \gls{CC} with 2 or 3 components.
For the sake of completeness and provide a more fair comparison to our \gls{CC} with a maximum of 3 components , \cref{tab:app_ensemble_moe_} extends our comparison in \cref{tab:ensemble_moe} by including (i) the \glspl{DE} with 3 components and (ii) the \glspl{DE} with Refining \gls{NMS}.
%
%
%
\cref{tab:app_ensemble_moe_} presents that with equal number of components, our \gls{CC} outperforms the best \gls{DE} with 3 components by $1.4$ AP ($44.1$ of PAA $\times 3$ vs $45.5$ AP of \gls{CC}).
Furthermore, in the case that Refining \gls{NMS} is used for \glspl{CC}, their performance consistenty improves; showing the effectiveness of our aggregator.
Still, the performance of the best \gls{DE}, i.e., PAA with 5 components and Refining \gls{NMS}, is $0.5$ AP lower compared to our \gls{CC}; demonstrating the effectiveness of our approach.

\blockcomment{
\begin{table*}[t]
    \small
    \centering
    \caption{Comparison of our \gls{CC}  with \glspl{DE} and Vanilla \gls{MoE}. \glspl{MoE} obtained by our \gls{CC} outperforms \gls{DE}s significantly even with less detectors. Our gains in green are obtained compared to the best single model for each performance measure, represented as underlined. The gap between \glspl{DE} and our \gls{CC} increases when the number of components is 3 for \glspl{DE}.}
    \label{tab:app_ensemble_moe3}
    \scalebox{0.8}{
    \begin{tabular}{c|c||c|c|c||c|c|c} 
    \toprule
    \midrule
    Model Type&Detector&$\mathrm{AP}$&$\mathrm{AP_{50}}$&$\mathrm{AP_{75}}$&$\mathrm{AP_{S}}$&$\mathrm{AP_{M}}$&$\mathrm{AP_{L}}$\\ \midrule
    \multirow{3}{*}{Single Models}&RS R-CNN&$42.4$&$\underline{62.1}$&$46.2$&$26.8$&$46.3$&$56.9$\\
    &ATSS&$43.1$&$61.5$&$\underline{47.1}$&$\underline{27.8}$&$\underline{47.5}$&$54.2$\\ 
    &PAA&$\underline{43.2}$&$60.8$&$\underline{47.1}$&$27.0$&$47.0$&$\underline{57.6}$\\ \midrule
    \multirow{3}{*}{Deep Ensembles}
    &RS R-CNN $\times$ 3&$44.2$&$63.0$&$48.5$&$28.2$&$48.4$&$58.5$\\
    &RS R-CNN $\times$ 5&$44.5$&$63.0$&$49.1$&$28.8$&$48.6$&$58.4$\\ \cdashlinelr{2-8}
    &ATSS $\times$ 3&$43.9$&$62.4$&$47.5$&$29.2$&$48.4$&$55.9$\\
    &ATSS $\times$ 5&$44.1$&$62.4$&$48.0$&$29.7$&$48.6$&$56.2$\\ \cdashlinelr{2-8}
    &PAA $\times$ 3&$44.2$&$62.0$&$47.9$&$28.2$&$48.3$&$59.1$\\
    &PAA $\times$ 5
    &$45.0$&$61.8$&$49.3$&$29.4$&$49.6$&$\textbf{59.9}$\\
    \midrule
    \multirow{4}{*}{Mixtures of Experts}&Vanilla \gls{MoE} with Ref. NMS (RS R-CNN, ATSS, PAA)&$44.4$&$62.6$&$48.2$&$27.8$&$48.4$&$\textbf{59.3}$\\  \cdashlinelr{2-8}
    &\gls{CC} (ATSS and PAA) - Ours&$44.8$&$62.4$&$49.2$&$29.4$&$49.1$&$57.6$\\
    &\gls{CC} (RS R-CNN, ATSS, PAA) - Ours&$\textbf{45.5}$&$\textbf{63.2}$&$\textbf{50.0}$&$\textbf{29.7}$&$\textbf{49.7}$&$\textbf{59.3}$\\
    &&$\imp{2.3}$&$\imp{2.1}$&$\imp{2.9}$&$\imp{1.9}$&$\imp{2.2}$&$\imp{1.6}$\\
    \midrule
    \bottomrule
    \end{tabular}
     }
\end{table*}

\paragraph{Comparison with \glspl{DE} with Less Components}
For the sake of completeness, \cref{tab:app_ensemble_moe3} extends our comparison in \cref{tab:ensemble_moe} by including the \glspl{DE} with 3 components.
Considering that our best \gls{MoE} obtained via \gls{CC} also includes 3 components, these experiments compare \glspl{DE} and \gls{CC} with the same number of components.
To compare them in their best setting, we use Refining NMS in all of the models.
\cref{tab:app_ensemble_moe3} presents that with equal number of components, our \gls{CC} outperforms the best \gls{DE} with 3 components by $1.3$ AP; further demonstrating its effectiveness.
}

\blockcomment{
\begin{table}
    \small
    \setlength{\tabcolsep}{0.25em}
    \centering
    \caption{Effect of Hyperparameters.}
    \label{tab:app_ablation}
    \scalebox{1.0}{
    \begin{tabular}{c|c||c|c|c|c} 
    \toprule
    \midrule
    Method&Soft NMS&COCO&LVIS mini-test&LVIS&LVIS mini-val\\ \midrule
    \multirow{7}{*}{Uncalibrated MoE}& \xmark &$43.4$&$40.0$&$25.9$&$39.5/37.5$\\
     & Linear &$43.4$&$40.0$&&$39.5/37.5$\\
     & Gaussian, $\sigma=0.20$ &$41.6$&$40.5$&&$39.6/37.9$\\
     & Gaussian, $\sigma=0.40$ &$42.1$&$40.3$&&$39.5/37.8$\\
     & Gaussian, $\sigma=0.60$ &$42.4$&$40.4$&&$39.7/37.9$\\
     & Gaussian, $\sigma=0.80$ &$42.7$&$40.4$&&$39.6/37.8$\\
     & Gaussian, $\sigma=1.00$ &$42.9$&$40.4$&&$39.7/37.9$\\ \midrule
    \multirow{7}{*}{Calibrated MoE}& \xmark &$44.7$&$41.6$&$27.7$&$42.0/39.8$\\
     & Linear &$44.8$&$41.9$&$27.7$&$42.2/39.9$\\
     & Gaussian, $\sigma=0.20$ &$43.7$&$42.0$&$28.5$&$42.4/40.6$\\
     & Gaussian, $\sigma=0.40$ &$44.4$&$42.4$&$28.5$&$43.0/40.8$\\
     & Gaussian, $\sigma=0.60$ &$44.8$&$42.1$&$28.1$&$42.9/40.6$\\
     & Gaussian, $\sigma=0.80$ &$44.7$&$42.1$&$27.6$&$42.6/40.2$\\
     & Gaussian, $\sigma=1.00$ &$44.6$&$41.6$&$27.1$&$42.3/39.9$\\ \midrule
    \bottomrule
    \end{tabular}
     }
\end{table}

\begin{table*}
    \setlength{\tabcolsep}{0.25em}
    \centering
    \caption{An analysis of different calibration measures on raw probabilities and final probabilities. We would expect no accuracy drop but improvement in calibration. For raw probabilities, Class-agnostic IR is the only option for that. For final calibration, all methods except Class-Agnostic IR provides similar accuracy. Having looked at the calibration performance,  So, we choose Class-agnostic IR for raw probabilities.}
    \label{tab:ensemble}
    \scalebox{1.0}{
    \begin{tabular}{c|c|c|c|c||c|c|c} 
    \toprule
    \midrule
    Model&Calibration&AP&AP50&AP75&LaECE&LaACE&LaMCE \\ \midrule
    &None&$\mathbf{42.4}$&$62.1$&$46.2$&$13.79$&$\mathbf{5.18}$&$35.37$\\    
    &Pre-NMS Class-wise LR&$26.4$&$36.9$&$29.1$&$0.44$&$25.12$&$61.11$\\
    &Pre-NMS Class-wise IR&$41.9$&$61.5$&$46.0$&$\mathbf{0.03}$&$\underline{5.23}$&$\underline{25.10}$\\
    &Pre-NMS Class-agnostic LR&$\mathbf{42.4}$&$62.1$&$46.3$&$0.65$&$27.48$&$68.81$\\
    RS R-CNN&Pre-NMS Class-agnostic IR&$\mathbf{42.4}$&$62.2$&$46.4$&$\underline{0.14}$&$5.86$&$\mathbf{14.08}$\\ \cline{2-8}
    &None&$\mathbf{42.4}$&$62.1$&$46.2$&$36.45$&$29.30$&$45.42$\\    
    &Post-NMS Class-wise LR&$\mathbf{42.4}$&$62.1$&$46.2$&$4.36$&$14.19$&$40.20$\\
    &Post-NMS Class-wise IR&$41.8$&$61.3$&$45.6$&$\mathbf{1.56}$&$\underline{9.21}$&$38.43$\\
    &Post-NMS Class-agnostic LR&$\mathbf{42.4}$&$62.1$&$46.2$&$5.83$&$13.86$&$\underline{37.79}$\\
    &Post-NMS Class-agnostic IR&$\mathit{42.3}$&$62.1$&$46.2$&$\underline{3.15}$&$\mathbf{8.93}$&$\mathbf{35.72}$\\
    \midrule
    &None&$\textbf{43.1}$&$61.5$&$47.1$&$0.20$&$25.22$&$42.53$\\
    &Pre-NMS Class-wise LR&$42.0$&$60.4$&$45.9$&$0.12$&$10.13$&$24.79$\\
    &Pre-NMS Class-wise IR&$42.8$&$60.9$&$46.6$&$\mathbf{0.02}$&$\mathbf{5.40}$&$25.38$\\
    &Pre-NMS Class-agnostic LR&$42.8$&$61.5$&$46.8$&$0.17$&$7.86$&$\underline{23.07}$\\
    ATSS&Class-agnostic IR&$\textbf{43.1}$&$61.6$&$47.1$&$\underline{0.10}$&$\underline{6.43}$&$\mathbf{15.59}$\\\cline{2-8}
    &None&$\textbf{43.1}$&$61.5$&$47.1$&$5.01$&$17.48$&$40.00$\\    
    &Post-NMS Class-wise LR&$\textbf{43.1}$&$61.5$&$47.0$&$\underline{2.69}$&$\mathbf{9.03}$&$\mathbf{29.51}$\\
    &Post-NMS Class-wise IR&$42.5$&$60.6$&$46.5$&$\mathbf{2.35}$&$9.81$&$40.34$\\
    &Post-NMS Class-agnostic LR&$\mathit{43.0}$&$61.5$&$47.0$&$4.46$&$\underline{9.46}$&$\underline{29.59}$\\
    &Post-NMS Class-agnostic IR&$\textbf{43.1}$&$61.5$&$47.1$&$4.51$&$9.51$&$37.35$\\
    \midrule
    &None&$\textbf{43.2}$&$60.8$&$47.1$&$3.39$&$9.73$&$19.80$\\
    &Pre-NMS Class-wise LR&$37.1$&$50.9$&$40.6$&$0.22$&$28.95$&$60.05$\\
    &Pre-NMS Class-wise IR&$42.5$&$59.8$&$46.4$&$\textbf{0.03}$&$\textbf{5.26}$&$\underline{26.60}$\\
    &Pre-NMS Class-agnostic LR&$\textbf{43.2}$&$60.8$&$47.1$&$0.37$&$25.49$&$54.03$\\
    &Pre-NMS Class-agnostic IR&$\textbf{43.2}$&$60.8$&$47.1$&$\underline{0.14}$&$\underline{6.70}$&$\textbf{18.15}$\\\cline{2-8}
    PAA&None&$\textbf{43.2}$&$60.8$&$47.1$&$11.23$&$15.79$&$\underline{32.33}$\\    
    &Post-NMS Class-wise LR&$\textbf{43.2}$&$60.8$&$47.1$&$\underline{1.39}$&$12.01$&$33.76$\\
    &Post-NMS Class-wise IR&$42.4$&$59.8$&$46.2$&$\textbf{1.21}$&$\textbf{9.24}$&$37.84$\\
    &Post-NMS Class-agnostic LR&$\textbf{43.2}$&$60.8$&$47.1$&$1.63$&$11.88$&$\textbf{29.91}$\\
    &Post-NMS Class-agnostic IR&$\textbf{43.2}$&$60.8$&$47.0$&$1.62$&$\underline{9.61}$&$35.59$\\
    \midrule
    \bottomrule
    \end{tabular}
    }
\end{table*}

\begin{table*}
    \small
    \setlength{\tabcolsep}{0.25em}
    \centering
    \caption{How many images do we need? Observations: (1) GLaECE is not alone a discriminative metric. (2) 500 images are enough to improve the calibration of object detectors. This is also a benefit of class-agnostic IR.}
    \label{tab:num_img}
    \scalebox{1.0}{
    \begin{tabular}{c|c||c|c|c||c|c|c|} 
    \toprule
    \midrule
    Method&Number of Images&AP&AP50&AP75&LaECE&LaACE&LaMCE\\ \midrule
    &0&$42.4$&$62.1$&$46.2$&$13.79$&$5.18$&$35.37$\\
    &50&$42.3$&$62.1$&$46.1$&$0.15$&$6.71$&$18.88$\\
    &100&$42.3$&$62.1$&$46.3$&$0.15$&$6.53$&$19.88$\\
    RS R-CNN&250&$42.3$&$62.1$&$46.2$&$0.14$&$5.90$&$14.88$\\
    &500&$42.4$&$62.1$&$46.3$&$0.14$&$5.85$&$13.66$\\
    &2500&$42.3$&$62.1$&$46.3$&$0.14$&$5.86$&$14.08$\\
    \midrule
    &0&$43.1$&$61.5$&$47.1$&$0.20$&$25.22$&$42.53$\\
    &50&$43.0$&$61.5$&$47.0$& $0.11$&$7.01$&$18.32$\\
    &100&$43.1$&$61.5$&$47.1$&$0.11$&$6.58$&$16.77$\\
    ATSS&250&$43.1$&$61.5$&$47.1$&$0.10$&$6.49$&$16.34$\\
    &500&$43.1$&$61.6$&$47.1$&$0.10$&$6.39$&$16.08$\\
    &2500&$43.1$&$61.5$&$47.0$&$0.10$&$6.43$&$15.59$\\ 
    \midrule
    &0&$43.2$&$60.8$&$47.1$&$3.39$&$9.73$&$19.80$\\
    &50&$43.2$&$60.7$&$47.0$& $0.15$&$7.33$&$20.78$\\
    &100&$43.1$&$60.8$&$47.0$&$0.15$&$7.41$&$25.61$\\
    PAA&250&$43.2$&$60.8$&$47.1$&$0.14$&$6.71$&$18.78$\\
    &500&$43.2$&$60.8$&$47.0$&$0.14$&$6.61$&$18.38$\\
    &2500&$43.2$&$60.8$&$47.1$&$0.14$&$6.70$&$18.15$\\
    \midrule
    \bottomrule
    \end{tabular}
    }
\end{table*}

\begin{table}
    \small
    \setlength{\tabcolsep}{0.25em}
    \centering
    \caption{How many images do we need? Class agn isotonic reg.}
    \label{tab:num_img_post}
    \scalebox{1.0}{
    \begin{tabular}{c|c||c|c|c||c|c|c|} 
    \toprule
    \midrule
    Method&Number of Images&AP&AP50&AP75&LaECE&LaACE&LaMCE\\ \midrule
    &0&$42.4$&$62.1$&$46.2$&$36.45$&$29.30$&$45.42$\\
    &50&$42.3$&$62.1$&$46.1$&$3.37$&$8.71$&$36.49$\\
    &100&$42.3$&$62.1$&$46.2$&$3.39$&$8.89$&$33.93$\\
    RS R-CNN&250&$42.3$&$62.1$&$46.2$&$3.21$&$9.10$&$35.94$\\
    RS R-CNN&500&$42.4$&$62.1$&$46.2$&$3.19$&$8.95$&$37.24$\\
    &2500&$42.3$&$62.1$&$46.3$&$3.15$&$8.93$&$35.72$\\
    \midrule
    &0&$43.1$&$61.5$&$47.1$&$5.01$&$17.48$&$40.00$\\
    &50&$43.0$&$61.5$&$47.0$& $4.50$&$9.23$&$37.34$\\
    &100&$43.0$&$61.3$&$47.0$&$4.76$&$9.37$&$36.64$\\
    ATSS&250&$43.1$&$61.5$&$47.1$&$4.58$&$9.62$&$35.29$\\
    ATSS&500&$43.1$&$61.5$&$47.1$&$4.54$&$9.83$&$38.26$\\
    &2500&$43.1$&$61.5$&$47.0$&$4.51$&$9.51$&$37.35$\\ 
    \midrule
    &0&$43.2$&$60.8$&$47.1$&$11.23$&$15.79$&$32.33$\\
    &50&$43.2$&$60.7$&$47.0$& $1.33$&$10.79$&$44.55$\\
    &100&$43.0$&$60.6$&$46.8$&$1.27$&$9.25$&$37.88$\\
    PAA&250&$43.2$&$60.8$&$47.1$&$1.54$&$8.14$&$32.19$\\
    PAA&500&$43.2$&$60.7$&$47.0$&$1.09$&$9.15$&$36.33$\\
    &2500&$43.2$&$60.8$&$47.1$&$1.62$&$9.61$&$35.59$\\
    \midrule
    \bottomrule
    \end{tabular}
    }
\end{table}
}
\begin{figure*}[t]
        \captionsetup[subfigure]{}
        \centering
        \begin{subfigure}[b]{0.32\textwidth}
            \includegraphics[width=\textwidth]{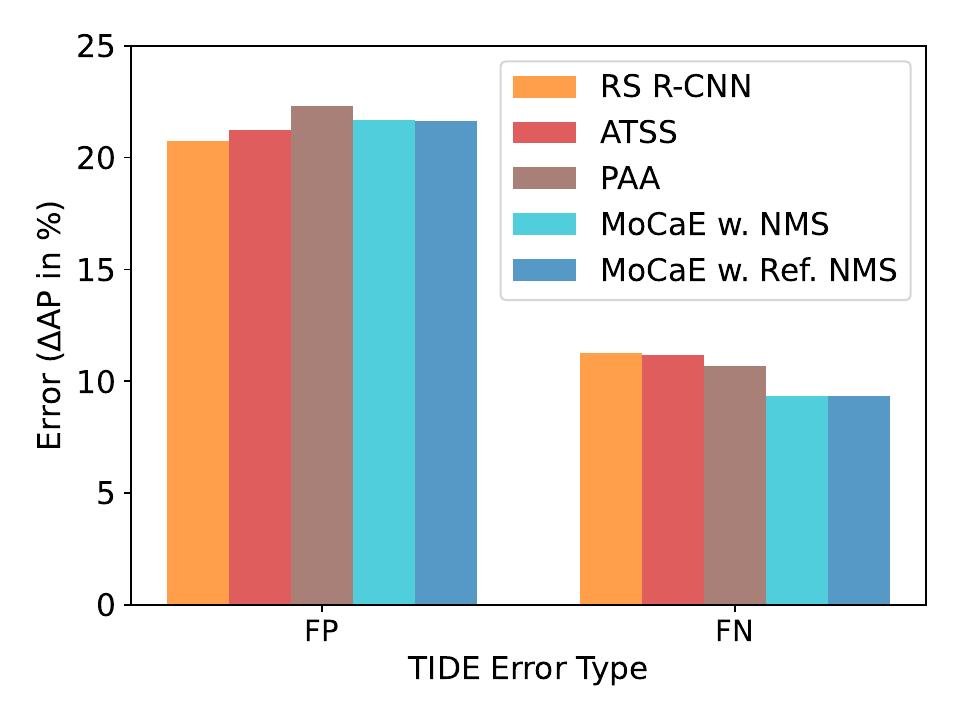}
            \caption{TIDE-style Analysis \cite{TIDE}}
        \end{subfigure}
        \begin{subfigure}[b]{0.32\textwidth}
            \includegraphics[width=\textwidth]{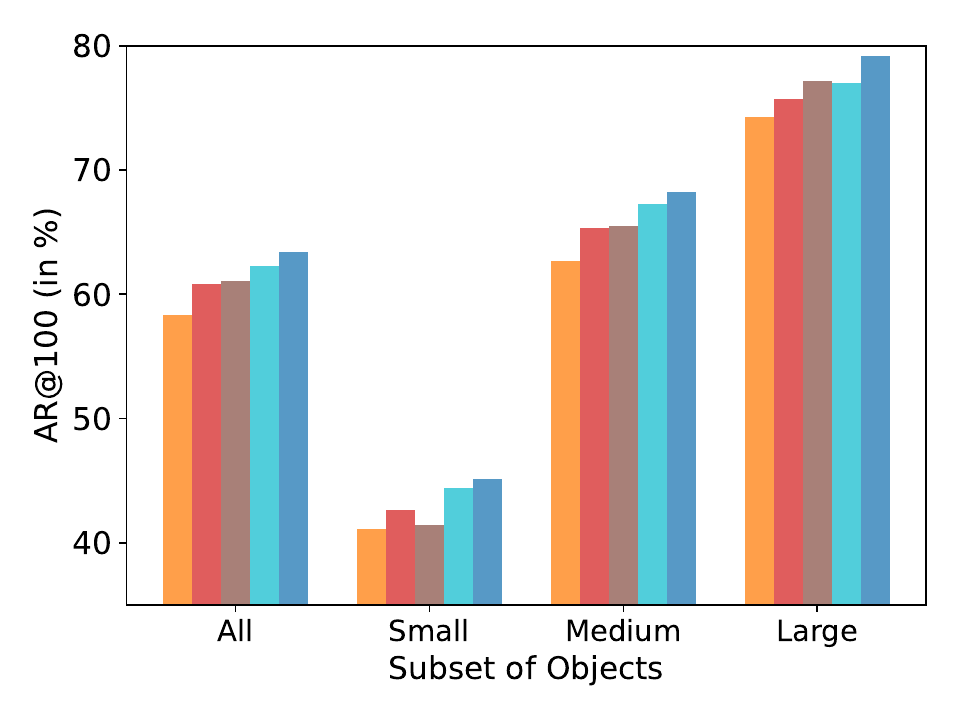}
            \caption{Average Recall}
        \end{subfigure}
        \begin{subfigure}[b]{0.32\textwidth}
            \includegraphics[width=\textwidth]{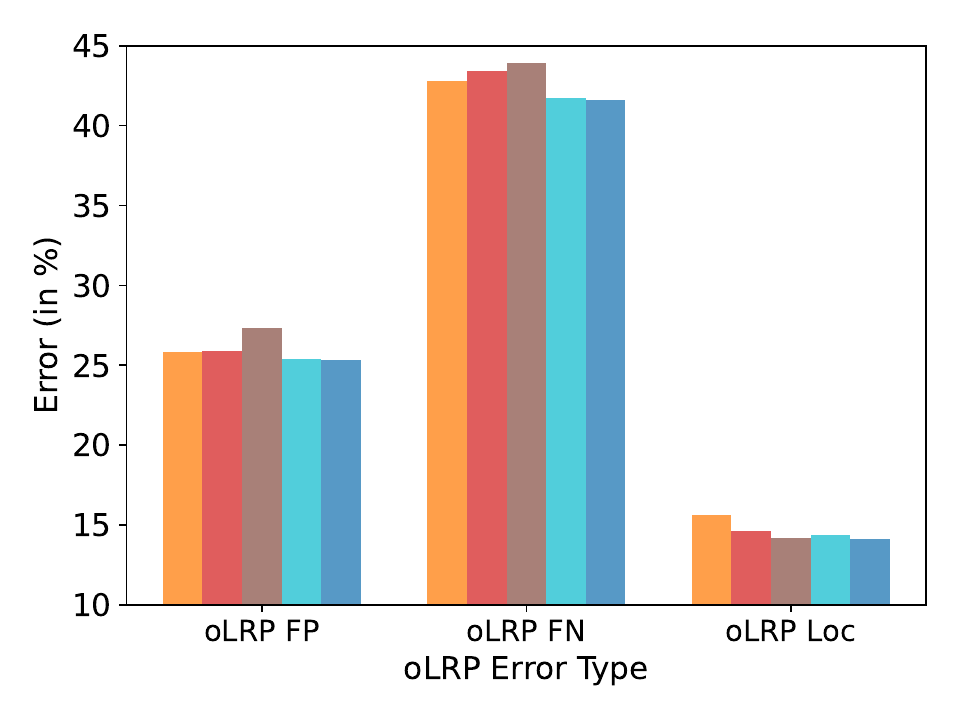}
            \caption{LRP Error Components \cite{LRPPAMI}}
        \end{subfigure}
        \caption{ The contribution of \gls{CC} on performance aspects
        }
        \label{fig:app_analysis}
\end{figure*}

\subsection{Why does \gls{CC} Improve Performance of Single Detectors?} 

Here, we investigate how \gls{CC} improves the performance of single detectors and note two main takeaways: First, \gls{CC} combines the detectors in a way that the resulting \gls{MoE} benefits from the complementary nature of the detectors; and as expected, second \gls{CC} mainly improves the recall of the single detectors.

\begin{figure}[]
\centering
\includegraphics[width=\textwidth]{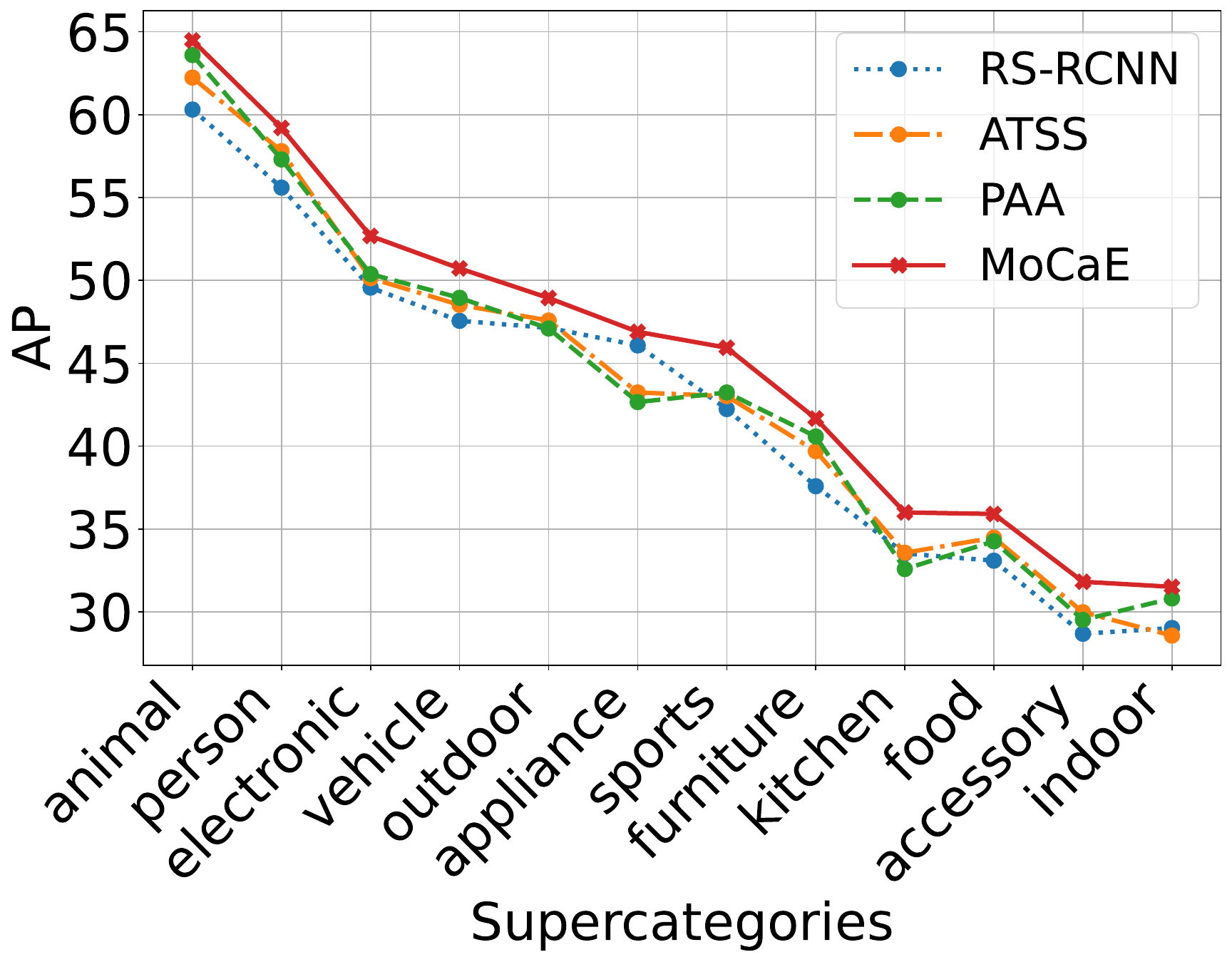}
\caption{$\mathrm{AP}$ scores of single models (RS-RCNN, ATSS, PAA) and \gls{CC} (consisting of RS-RCNN, ATSS, PAA) on different supercategories of the COCO \textit{mini-test} dataset. \gls{CC} performs consistently higher than any of the single models on every supercategory. \label{tab:classwise_ap_scores}}
\end{figure}

\subsubsection{Complementary Nature of \gls{CC}} 
Ideally, an \gls{MoE} aims to combine the individual components in such a way that they complement each other in different subsets of the input space to achieve a better performance.
Here, we show that this is, in fact, the case also for \gls{CC}.
%
Similar to our previous analyses, we combine ATSS, PAA and RS R-CNN and report the results on COCO \textit{mini-test}.
Differently, in order to be able to discuss the results, we utilize 12 supercategories such as \textit{animal}, \textit{food}, which are already present in COCO dataset.
\cref{tab:classwise_ap_scores} shows the results in which we can easily observe that different experts prevail on different supercategories.
To illustrate, RS R-CNN \cite{RSLoss} is the best single model in the \textit{appliance} supercategory, ATSS \cite{ATSS} is the best single model in the \textit{kitchen} supercategory and PAA \cite{paa} is the best in the \textit{indoor} supercategory.
A noteworthy point is that\gls{CC} performs better than \textit{any} of its components in \textit{all of the supercategories}, achieving notable improvements over the entirety of the input space.
This effectively shows the capability of \gls{CC} to leverage the complementary expertise of the single models to form a more accurate mixture.

\subsubsection{The Contribution of \gls{CC} on Different Performance Aspects} 
We now analyse what performance aspects, among localisation, recall and precision, are affected by \gls{CC} by expliting three different analyses tools from the detection literature.
First, we use TIDE that defines oracle APs as the \glspl{AP} obtained when \gls{FP} and \gls{FN} errors are completely mitigated.
Then, the difference between the oracle and actual AP correspond to the error of a detector for a specific performance aspect.
More specifically, while obtaining Oracle FP AP, the FP detections are simply removed from the final detection set. 
As for Oracle FN AP, the FN objects are removed from the dataset.
\cref{fig:app_analysis}(a) shows that \gls{CC} variants perform similar to the single detectors in terms of \gls{FP} Error while they clearly outperform them on \gls{FN} Error.
This indicates that one of the contributions of \gls{MoE} is to find the objects that are not detected by at least one of the individual detector as illustrated on \cref{fig:teaser} and \cref{tab:teaser_method}.
However, TIDE analysis does not provide insight on the localisation quality which is mainly targeted by Refining NMS.
Therefore, in our second analysis we exploit \gls{AR} defined as the average of the recall values over 10 IoUs from $0.50$ to $0.95$.
Aligned with the observation in TIDE analysis, \cref{fig:app_analysis}(b) presents that \gls{CC} improves \gls{AR} of the single detectors.
Furthermore, the performance gain mainly originates from the improvement in small and medium objects as \gls{CC}  does not improve the performance on large objects notably.
This indicates that the resulting \gls{MoE} is especially stronger than single detectors in more challenging object categories.
Using Refining NMS further boosts the AR performance as it improves the localisation performance, which is critical for recalls with higher \glspl{IoU}.
To investigate the benefit of \gls{MoE} in practical use-cases, we finally conduct an LRP analysis \cite{LRPPAMI,LRP} in \cref{fig:app_analysis}(c).
In this figure, oLRP FP, FN and Loc correspond to the components Optimal LRP Error defined as 1-Precision, 1-Recall and the average IoU Error of \gls{TP} detections.
Aligned with our previous analysis, \gls{MoE} mainly decreases the recall error and \gls{CC} with Refining NMS mainly contributes to the oLRP Loc component, outperforming all individual models in the end.
These three different analyses confirm that \gls{CC} with NMS mainly decreases the recall error of the detector and using Refining NMS contributes to the localisation error.

\blockcomment{
\begin{table}[t]
    \small
    \centering
    \caption{Ablation analysis using an \gls{MoE} of RS R-CNN, ATSS and PAA on COCO \textit{mini-test}.}
    \label{tab:app_ablation}
    \scalebox{0.8}{
    \begin{tabular}{c|c|c|c||c|c|c||c|c|c} 
    \toprule
    \midrule
    \multirow{2}{*}{Pipeline}&\multirow{2}{*}{Calibration}&\multicolumn{2}{c|}{Refining NMS}&\multicolumn{6}{c}{Detection Performance} \\
    & &Soft NMS&Score Voting&$\mathrm{AP}$&$\mathrm{AP_{50}}$&$\mathrm{AP_{75}}$&$\mathrm{AP_{S}}$&$\mathrm{AP_{M}}$&$\mathrm{AP_{L}}$ \\ \midrule
    Early &\xmark&\xmark&\xmark&$43.3$&$62.6$&$46.8$&$27.2$&$47.1$&$57.8$\\
    Early &\cmark&\xmark&\xmark&$44.5$&$\mathbf{63.7}$&$48.4$&$29.0$&$48.5$&$57.5$\\
    Late &\cmark&\xmark&\xmark&$44.7$&$63.1$&$48.9$&$29.2$&$49.0$&$58.2$\\
    Late &\cmark&\cmark&\xmark&$44.8$&$63.1$&$49.1$&$29.3$&$49.1$&$58.3$\\
    Late &\xmark&\cmark&\xmark&$43.4$&$62.5$&$47.1$&$27.3$&$47.3$&$58.0$\\
    Late &\xmark&\cmark&\cmark&$44.4$&$62.6$&$48.2$&$27.8$&$48.4$&$\mathbf{59.3}$\\
    Late &\cmark&\cmark&\cmark&$\mathbf{45.5}$&$63.2$&$\mathbf{50.0}$&$\mathbf{29.7}$&$\mathbf{49.7}$&$\mathbf{59.3}$\\
    \midrule
    \bottomrule
    \end{tabular}
     }
\end{table}
}
\begin{figure*}[t]
        \captionsetup[subfigure]{}
        \centering
        \begin{subfigure}[b]{0.4\textwidth}
            \includegraphics[width=\textwidth]{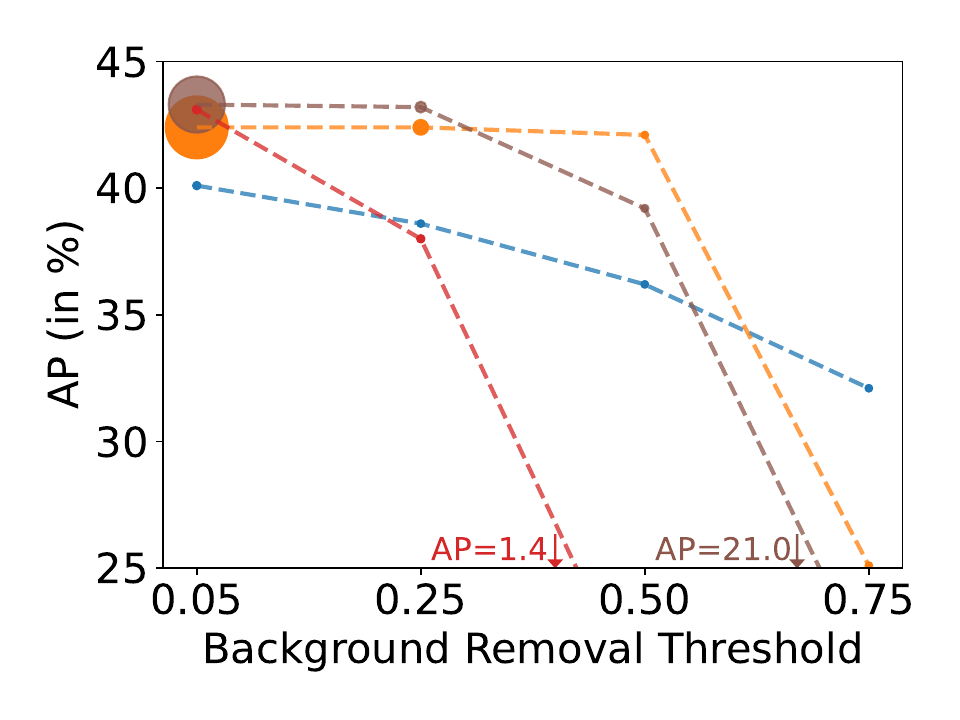}
            \caption{Uncalibrated}
        \end{subfigure}
        \begin{subfigure}[b]{0.4\textwidth}
            \includegraphics[width=\textwidth]{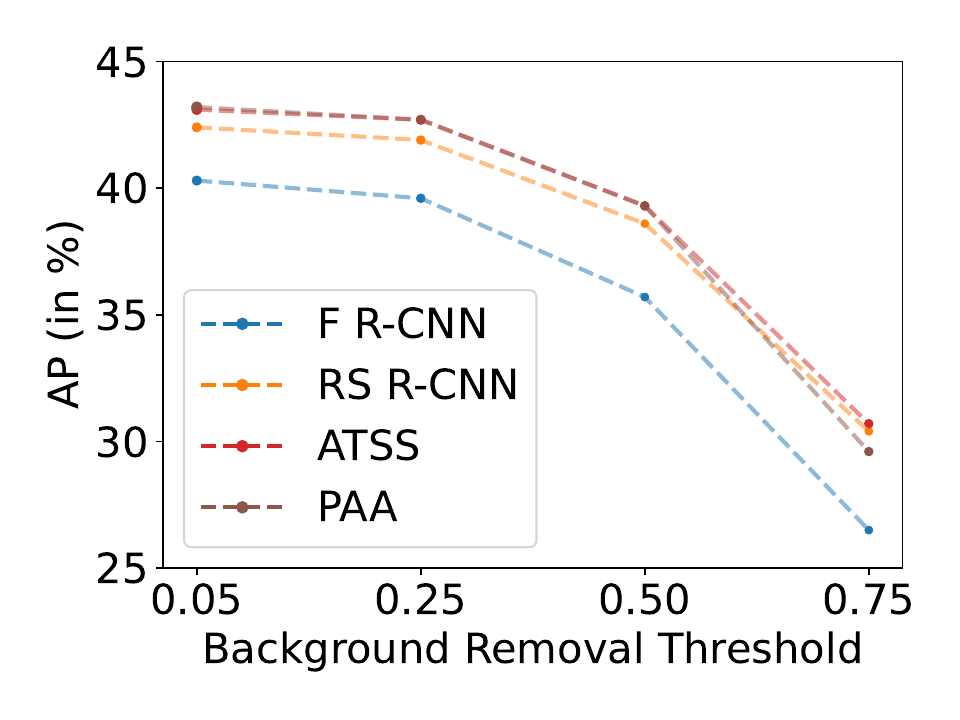}
            \caption{Early Calibrated}
        \end{subfigure}
        %
        \caption{The effect of background removal threshold on \gls{AP} and \gls{NMS} processing time for \textbf{(a)} uncalibrated and \textbf{(b)} early calibrated detectors on COCO. The area of the dots are proportional to the NMS processing time of the detectors. This threshold is typically set to $0.05$, in which case PAA and RS R-CNN have large \gls{NMS} processing time in \textbf{(a)}. Once uncalibrated, the detectors follow different trends and are sensitive to the threshold in terms of \gls{AP} and \gls{NMS} processing time. In \textbf{(b)}, early calibration (i) aligns the detector by reducing this sensitivity, and (ii) allows using $0.05$ by both maximizing the \gls{AP} and reducing \gls{NMS} processing time for the over-confident PAA and RS R-CNN. F R-CNN refers to Faster R-CNN.
        }
        \label{fig:prenms}
\end{figure*}

\begin{table}
    \small
    \setlength{\tabcolsep}{0.25em}
    \centering
    \caption{The values used in \cref{fig:prenms}. Early calibration regularizes the behaviour of the detectors by reducing their sensitivity to background removal threshold with respect to AP and NMS time. NMS time is measured in terms of ms using a single Nvidia 1080ti GPU.}
    \label{tab:nmsineff}
    \scalebox{0.85}{
    \begin{tabular}{c|c||c|c|c|c} 
    \toprule
    \midrule
    Background Removal&Detector&\multicolumn{2}{|c|}{Uncalibrated}&\multicolumn{2}{|c}{Early Calibrated}   \\ \cline{3-6}
    Threshold&&AP&NMS time&AP&NMS time   \\ \midrule
    &Faster R-CNN&$40.1$&$0.5$&$40.3$&$0.6$\\
    0.05&RS-RCNN &$42.4$&$35.4$&$$42.4$$&$0.6$\\
    &ATSS &$43.1$&$0.6$&$43.1$&$0.7$\\
    &PAA &$43.3$&$29.2$&$43.2$&$0.8$\\  \midrule
    &Faster R-CNN&$38.6$&$0.4$&$39.6$&$0.5$\\
    0.25&RS-RCNN &$42.4$&$2.0$&$41.9$&$0.5$\\
    &ATSS &$38.0$&$0.5$&$42.7$&$0.6$\\
    &PAA &$43.2$&$1.1$&$42.7$&$0.5$\\  \midrule
    &Faster R-CNN&$36.2$&$0.4$&$35.7$&$0.4$\\
    0.50&RS-RCNN &$42.1$&$0.4$&$38.6$&$0.4$\\
    &ATSS &$19.3$&$0.4$&$39.3$&$0.5$\\
    &PAA &$39.2$&$0.5$&$39.3$&$0.5$\\  \midrule
    &Faster R-CNN&$32.1$&$0.4$&$26.5$&$0.4$\\
    0.75&RS-RCNN &$25.1$&$0.4$&$30.4$&$0.4$\\
    &ATSS &$1.4$&$0.0$&$30.7$&$0.5$\\
    &PAA &$21.0$&$0.4$&$29.6$&$0.5$\\  \midrule
    \bottomrule
    \end{tabular}
    }
\end{table}

\subsection{A Use Case for Early Calibration: Reducing the Sensitivity to Background Removal Threshold} 
Here, we investigate an additional use-case of early calibration in which it reduces the sensitivity of the detectors to background removal threshold in terms of both \gls{AP} and efficiency.
As \gls{AP} provably benefits from more detections \cite{saod}, detectors prefer a small background removal threshold as the first step of post-processing (\cref{fig:cc}(b)).
\textit{To illustrate, $0.05$ is the common choice for COCO \cite{ATSS,paa,FasterRCNN} and it is as low as $10^{-4}$ for LVIS \cite{mmdetection,LVIS}.}
While this convention is preferred by AP, it can easily increase \gls{NMS} processing time especially for over-confident detectors.
This is because, for such detectors, the background removal step accepts redundant \glspl{TN}, which should have been rejected. 
Hence, due to this large number of redundant \glspl{TN} propagated to the \gls{NMS}, \gls{NMS} processing time significantly increases.
To illustrate on PAA, which uses a threshold of $0.05$ for COCO, NMS takes $29.2$ ms/image on a Nvidia 1080Ti GPU, while it only takes $\sim 0.6$ ms/image for ATSS and Faster R-CNN (F R-CNN).
This difference among the detectors can easily be noticed by comparing the areas of the dots at $0.05$ in \cref{fig:prenms}(a)\footnote{\cref{tab:nmsineff} shows the exact values used to obtain \cref{fig:prenms}}.
%
%
Specifically, we observed for PAA that $\sim 45$K detections are propagated to the NMS per image on average. 
After early calibration, this number of detections from the same threshold reduces to $\sim 2$K per image,  which now enables \gls{NMS} to take only $0.8$ ms/image as ideally expected. 
\cref{fig:prenms}(b) presents that NMS takes consistently between $0.6$ to $0.8$ ms/image for all detectors as the behaviour of the detectors are aligned.
%

%


\begin{table*}
    \centering
    \caption{Experiments with Self-aware Object Detectors. We use the General Object Detection setting in \cite{saod}. Please refer to the text for the details of the performance measures.}
    \label{tab:saod}
    \scalebox{0.75}{
    \begin{tabular}{c|c|c|c|c|c|c|c|c|c}
    \toprule
    \midrule
         Model Type&Self-aware&\multirow{2}{*}{$\mathrm{DAQ}\uparrow$}&OOD Detection Performance&\multicolumn{3}{c|}{Accuracy and Calibration for In-distribution}&\multicolumn{3}{c}{Accuracy and Calibration for Domain-shift} \\
         &Detector& &$\mathrm{BA}\uparrow$&$\mathrm{IDQ}\uparrow$&$\mathrm{LaECE}\downarrow$&$\mathrm{LRP}\downarrow$&$\mathrm{IDQ}\uparrow$&$\mathrm{LaECE}\downarrow$&$\mathrm{LRP}\downarrow$\\
    \midrule
    \multirow{3}{*}{Single Models}
    &SA-RS-RCNN&$\underline{40.9}$&$\underline{\mathbf{89.0}}$&$\underline{39.6}$&$18.0$&$\underline{73.9}$&$\underline{27.1}$&$\underline{19.2}$&$83.7$\\
    &SA-ATSS&$\underline{40.9}$&$87.9$&$\underline{39.6}$&$17.9$&$\underline{73.9}$&$27.3$&$20.6$&$\underline{83.5}$\\
    &SA-PAA&$40.6$&$87.3$&$39.0$&$\underline{17.8}$&$74.4$&$\underline{27.1}$&$21.0$&$83.6$\\
    \midrule
    \multirow{2}{*}{\glspl{MoE}} &Vanilla MoE&$42.5$&$88.7$&$39.9$&$18.1$&$73.6$&$29.2$&$19.6$&$82.1$\\
    & \gls{CC}&$\mathbf{42.9}$&$87.6$&$\mathbf{40.1}$&$\mathbf{17.1}$&$\mathbf{73.5}$&$\mathbf{29.8}$&$\textbf{19.1}$&$\mathbf{81.8}$\\
    \midrule
    \bottomrule
    \end{tabular}
    }
\end{table*}

\begin{table}
    \small
    \centering
    \caption{Object discovery on the out-of-distribution set (SinObj110K-OOD) from SAOD \cite{saod}. Our gains in \textcolor{forestgreen}{green} are highlighted with respect to the best single model (\underline{underlined}).}
    \label{tab:open_set_ood}
    \scalebox{1.0}{
    \begin{tabular}{c|c||c} 
    \toprule
    \midrule
    Model Type&Detector&$\mathrm{AR}$\\ \midrule
    \multirow{3}{*}{Single Models}&RS R-CNN&$41.3$\\
    &ATSS&$\underline{42.0}$\\ 
    &PAA&$39.7$ \\ \midrule
    \multirow{2}{*}{MoEs}
    &Vanilla \gls{MoE} &$42.7$\\ 
    &\gls{CC} - Ours&$\textbf{45.6}$\\
    && $\imp{3.6}$ \\
    \midrule
    \bottomrule
    \end{tabular}}
\end{table}

\subsection{Further Discussion on \gls{SAOD} Task}
%
\paragraph{Details of the \gls{SAOD} task}
Self-aware object detection task (SAOD) \cite{saod} provides a comprehensive framework to evaluate the robustness of object detectors.
Specifically, the performance aspects that are jointly considered in this task are out of distribution detection, calibration, domain shift as well as the accuracy.
Specifically, \gls{SAOD} evaluates an object detector in terms of the following criteria:
\begin{compactitem}
    \item Rejecting OOD images utilizing reliable image-level uncertainty estimates
    \item Yielding accurate and calibrated detections
    \item Being robust to domain shift under varying severities.
\end{compactitem}

The evaluation is conducted using more than $150K$ single images, which is a large-scale test set enabling thorough evaluation.
Specifically, the \gls{SAOD} task utilizes the following split of the datasets for evaluating a detector in terms of the aforementioned performance aspects:
\begin{compactitem}
        \item $D_{ID}$ : The in-distribution dataset consisting of images containing the same set of foreground objects as the training set.
    \item $T(D_{ID})$ : Domain-shift in-distribution dataset obtained through applying transformations from \cite{hendrycks2019robustness} with severities $1$, $3$ and $5$ on the in-distribution set.
    \item $D_{OOD}$ : The out-of-distribution dataset that contains only the objects of different foreground classes than that of the $D_{ID}$.
\end{compactitem}
%

%
%

On this dataset, the ideal behavior expected from a robust object detector for a given input $X$ is:

\begin{compactitem}
        \item If $X \in D_{ID}$, ``accept'' the input and provide accurate and calibrated detections, any rejection is penalized.
    \item If $X \in T(D_{ID})$, with severities $1$ and $3$, ``accept'' the input and provide accurate and calibrated detections, any rejection is penalized.
    \item If $X \in T(D_{ID})$, with severity $5$, provide the choice to ``accept'' the input though no penalty for rejections as transformed images might have severe deformities with respect to their original versions.
    \item If $X \in D_{OOD}$, ``reject'' the input and refrain from providing any detections, any accept is be penalized.
\end{compactitem}
%

%
%
%


%

In association with these datasets, the authors also propose the following evaluation measures:
\begin{compactitem}
    \item Balanced Accuracy (BA) measures the OOD performance as the harmonic mean of TPR and TNR in this binary classification problem.
    \item \gls{LaECE} measures the calibration performance, as discussed in App \ref{app:relatedwork}.
    \item In-Distribution Quality (IDQ) combines accuracy and calibration performance as the harmonic mean of 1-LRP \cite{LRP} and 1-LaECE on in-distribution data $D_{ID}$. Analogously, for domain-shifted data, $T(D_{ID})$, $\text{IDQ}_{\text{T}}$ is used.
    \item Finally, Distribution-Awareness Quality (DAQ) as the main performance measure of the task, unifies these measures. Specifically, DAQ is a higher the better measure, defined as the harmonic mean of BA, IDQ and $\text{IDQ}_{\text{T}}$. 
\end{compactitem}

\paragraph{Implementation Details}
Based on the definition of this task, the authors also propose an algorithm to convert any detector to a self-aware one\footnote{We refer the reader to the Algorithm A.1 and Algorithm A.2 in \cite{saod} for the details.}.
We follow the proposed algorithm to make RS R-CNN, ATSS and PAA, as well as Vanilla \gls{MoE} and \gls{CC} self-aware.
We use the General Object Detection setting in \cite{saod} as our models are trained on COCO, aligned with this dataset.
We utilise the official SAOD code throughout our experiments by keeping all the settings.

\paragraph{Discussion of the Results}
The results are presented in \cref{tab:saod} in which \gls{CC} improves DAQ measure, the main performance measures of this task, up to $+2$ points compared to the best single model while also showing notable improvements in terms of the rest of the measures.
Furthermore, \gls{CC} also outperforms Vanilla \gls{MoE}.
This highlights that the \gls{CC} is more reliable than its counterparts in terms of the reliablity aspects considered within the SAOD \cite{saod} framework.
When we examine the individual robustness aspects, we can easily see that \gls{CC} outperforms all of the single detectors and Vanilla \gls{MoE} in terms of calibration and accuracy both on in-distribution and domain-shifted data.
On the other hand, we observe that the OOD performance drops from $\sim 1$ BA compared to the best single model.
In the following, we further discuss why this is the case and provide more insight.

\paragraph{Is Lower OOD Performance a Pitfall of \gls{CC} or its Strength?}
\gls{CC} combines individual detectors to ideally benefit from the strength of each detector in the mixture.
This commonly manifests itself to increase the recall performance of the individual components, i.e. Average Recall (AR), as we presented in \cref{tab:teaser_method} while motivating \gls{CC} as well as in \cref{fig:app_analysis}.
AR consistently increases in such cases due to the fact that the detections of the individual experts are combined properly in in-distribution images.

Now, consider an alternative case, in which the input image is out-of-distribution (OOD), on which each detection is a \gls{FP} as an OOD image does not contain an in-distribution object (as defined by \cite{saod}).
In this specific case, combining the detectors properly might result in more \glspl{FP} easily.
To illustrate on a toy example, assume there are two images $X_1$ and $X_2$ and $k$ represents the number of maximum detections in an image, which is typically $k=100$.
Also assume that, an overconfident detector yields a high-confident \gls{FP} on an \gls{OOD} image $X_1$ with $k-1$ lower confident detections. As for $X_2$ this detector does not have a high-confident detection.
This detector accepts the OOD image $X_1$ and rejects $X_2$.
A second but an underconfident detector yields very low confidence detections on $X_1$ and one relatively confident detection on $X_2$.
In contrary to $X_1$, this detector accepts the OOD image $X_2$ and rejects $X_1$.
When we combine these two detectors without calibration, the resulting Vanilla \gls{MoE} will mimic the overconfident detector, potentially rejecting $X_2$ and accepting $X_1$.
Unfortunately, calibrating these two detectors in the form of \gls{CC} will have a confident detection in each image, potentially resulting in the acceptance of both OOD images.
This is why Vanilla \gls{MoE} mimics the most confident detector in \cref{tab:saod}.
In this specific case, the most confident detector, RS R-CNN has the largest BA, and consequently Vanilla \gls{MoE} outperforms \gls{CC} in terms of \gls{OOD} performance.

Following from the comparison of \gls{OOD} performances of \glspl{MoE}, the question arises why detectors generate high confidence detections on \gls{OOD} images and whether there is a benefit of having them.
More specifically, we conjecture that this is mainly a result of object discovery as the $D_{OOD}$ merely contains the classes that are not exactly present in $D_{ID}$, in which objects of semantically similar classes can still exist across the two.
To support this claim, we design an experiment on the same \gls{OOD} split.
In this experiment, we use the bounding box annotations already present in the \gls{OOD} split and check whether which of the models find the most number of objects.
We do not distinguish among the classes as all the classes in \gls{OOD} split (SinObj110K-OOD) pertains to the unknown class for all of the models we use.
We report \gls{AR} in \cref{tab:open_set_ood} to evaluate the models with respect to their object discovery characteristics.
The results show that \gls{CC} outperforms all individual models and Vanilla \gls{MoE} with a significant margin in terms of \gls{AR}; demonstrating that it finds more objects compared to the other models.
This is because, unsurprisingly, the confident detections of individual models correspond to the objects and combining them after calibration in \gls{CC} results in finding more unknown objects.
As a result, while \gls{CC} does not perform the best in terms of image-level \gls{OOD} detection, it is, in fact, a better alternative for object discovery.
%

%

%

\begin{table*}[t]
\small
\centering
\caption{Oracle \gls{MoE} and pitfalls of \gls{CC}. Object detection performance on COCO \textit{mini-test}.  When the single methods have significant performance gap, then \gls{CC} might not perform well as the calibrators are imperfect.  Oracle \gls{MoE} is an \gls{MoE} in which each expert is perfectly calibrated. As a result, all experts have $0$ \gls{LaECE} in Oracle \gls{MoE}. Following \cref{theorem:mocae}, Oracle \gls{MoE} outperforms all single detectors. }
\label{tab:pitfall1}
\scalebox{0.95}{
\begin{tabular}{c|c||c|c|c||c|c|c||c|c} 
\toprule
\midrule
Model Type&Method&$\mathrm{AP}$&$\mathrm{AP_{50}}$&$\mathrm{AP_{75}}$&$\mathrm{AP_{S}}$&$\mathrm{AP_{M}}$&$\mathrm{AP_{L}}$&Unc. \gls{LaECE}&Cal. \gls{LaECE} \\ \midrule
\multirow{8}{*}{Single Models}&RS R-CNN &$42.4$&$62.1$&$46.2$&$26.8$&$46.3$&$56.9$&$36.45$&$3.19$\\
&ATSS &$43.1$&$61.5$&$47.1$&$27.8$&$47.5$&$54.2$&$5.01$&$4.54$\\
&PAA &$43.2$&$60.8$&$47.1$&$27.0$&$47.0$&$57.6$&$11.23$&$1.09$\\
&YOLOv7&$55.6$&$73.1$&$60.6$&$41.2$&$60.4$&$69.5$&$9.23$&$4.60$\\
&QueryInst&$55.9$&$75.4$&$61.3$&$38.5$&$60.8$&$73.2$&$4.54$&$3.19$\\
&DyHead&$56.8$&$75.6$&$62.2$&$42.8$&$60.6$&$71.0$&$10.04$&$6.26$\\
&EVA&$\underline{64.5}$&$\underline{82.3}$&$\underline{71.0}$&$50.6$&$\underline{68.9}$&$78.1$&$13.39$&$7.13$\\
&Co-DETR&$\underline{64.5}$&$81.7$&$70.8$&$\underline{51.0}$&$68.6$&$\underline{79.3}$&$5.81$&$4.82$\\
\midrule \midrule
\glspl{MoE} of&Vanilla \gls{MoE} &$64.6$&$82.3$&$71.3$&$50.7$&$68.8$&$79.0$&N/A&N/A\\
EVA and Co-DETR&\gls{CC}&$65.0$&$82.6$&$71.5$&$51.0$&$69.0$&$79.6$&N/A&N/A\\ 
&Oracle \gls{MoE} &$\mathbf{81.9}$&$\mathbf{96.6}$&$\mathbf{91.1}$&$\mathbf{72.7}$&$\mathbf{85.2}$&$\mathbf{92.1}$&N/A&N/A \\ \midrule \midrule
\glspl{MoE} of&Vanilla \gls{MoE} &$60.0$&$76.3$&$65.8$&$43.8$&$64.9$&$76.5$&N/A&N/A\\
RS R-CNN, ATSS, PAA&\gls{CC}&$61.9$&$78.8$&$69.0$&$49.0$&$66.6$&$76.4$ &N/A&N/A\\ 
EVA, Co-DETR&Oracle \gls{MoE} &$\mathbf{85.3}$&$\mathbf{97.2}$&$\mathbf{93.2}$&$\mathbf{76.3}$&$\mathbf{88.5}$&$\mathbf{94.5}$ &N/A&N/A\\ \midrule \midrule
\glspl{MoE} of&Vanilla \gls{MoE} &$64.1$&$81.4$&$70.6$&$50.2$&$68.5$&$78.4$&N/A&N/A\\
YOLOv7, QueryInst, DyHead&\gls{CC}&$63.1$&$80.1$&$69.5$&$49.9$&$68.0$&$77.5$ &N/A&N/A\\ 
EVA, Co-DETR&Oracle \gls{MoE} &$\mathbf{86.1}$&$\mathbf{97.4}$&$\mathbf{93.8}$&$\mathbf{77.4}$&$\mathbf{89.1}$&$\mathbf{95.0}$ &N/A&N/A\\ \midrule \midrule
\glspl{MoE} of&Vanilla \gls{MoE}&$60.1$&$76.3$&$66.0$&$44.1$&$65.1$&$76.6$&N/A&N/A\\
All Single&\gls{CC} &$61.7$&$78.3$&$68.1$&$48.7$&$66.9$&$76.3$ &N/A&N/A\\ 
Models&Oracle \gls{MoE}&$\mathbf{86.7}$&$\mathbf{97.4}$&$\mathbf{94.1}$&$\mathbf{78.0}$&$\mathbf{89.8}$&$\mathbf{95.5}$ &N/A&N/A\\
\bottomrule
\end{tabular}
}
\end{table*}

\blockcomment{
\begin{table}[t]
\small
\setlength{\tabcolsep}{0.7em}
\centering
\caption{Object detection performance on COCO \textit{mini-test} using SOTA object detectors. The gains are reported compared to the best single model as underlined. \gls{CC} reaches \gls{SOTA}.  }
\label{tab:found_model_minitest}
\scalebox{0.85}{
\begin{tabular}{c||c|c|c||c|c|c} 
\toprule
\midrule
Method&$\mathrm{AP}$&$\mathrm{AP_{50}}$&$\mathrm{AP_{75}}$&$\mathrm{AP_{S}}$&$\mathrm{AP_{M}}$&$\mathrm{AP_{L}}$ \\ \midrule
EVA \cite{EVA}&$\underline{64.5}$&$\underline{82.3}$&$\underline{71.0}$&$50.6$&$\underline{68.9}$&$78.1$\\
Co-DETR \cite{codetr}&$\underline{64.5}$&$81.7$&$70.8$&$\underline{51.0}$&$68.6$&$\underline{79.3}$\\
\midrule 
Vanilla \gls{MoE} &$64.6$&$82.3$&$71.3$&$50.7$&$68.8$&$79.0$\\
&$\imp{0.1}$&$\textcolor{blue}{0.0}$&$\imp{0.3}$&$\nimp{0.3}$&$\nimp{0.1}$ &$\nimp{0.3}$\\ \midrule 
\gls{CC} &$\mathbf{65.0}$&$\mathbf{82.6}$&$\mathbf{71.5}$&$\mathbf{51.0}$&$\mathbf{69.0}$&$\mathbf{79.6}$ \\ 
(Ours)&$\imp{0.5}$&$\imp{0.3}$&$\imp{0.5}$&$\textcolor{blue}{0.0}$&$\imp{0.1}$&$\imp{0.3}$ \\ 
\midrule
Oracle \gls{MoE} &$86.7$&$97.4$&$94.1$&$78.0$&$89.8$&$95.5$ \\
\bottomrule
\end{tabular}
}
\end{table}
}

\subsection{The Effect of \cref{theorem:mocae} using an Oracle \gls{MoE}}
\label{subsec:practicalimplications}
We proved in \cref{theorem:mocae} that the calibration target that we designed in Eq. \eqref{eq:gen_calibration_} provides the optimal AP under certain assumptions on post-processing.
Here, we show that the assumptions are, in fact, not very strict and significantly higher AP values can easily be observed once the individual detectors in the mixture are perfectly calibrated as suggested in \cref{theorem:mocae}.
To do so, we design an Oracle \gls{MoE}, in which we replace the confidence score of the detections by the corresponding calibration target in Eq. \eqref{eq:gen_calibration_} such that the detectors are now perfectly calibrated.
In Oracle \gls{MoE}, we use the standard \gls{NMS} to keep its design simple.
\cref{tab:pitfall1} presents the results of this Oracle \gls{MoE} combining different object detectors on COCO \textit{mini-test}.
When only two detectors, EVA and Co-DETR, are combined using Oracle \gls{MoE}, we observe that the gain of the Oracle \gls{MoE} compared to the best single model is more than $15$AP, which is very significant.
Furthermore, as the number of components increases, the performance of Oracle \gls{MoE} consistently increases up to $86.7$ AP and $97.4$ $\mathrm{AP_{50}}$ when all of the eight detectors (with very different performances) are combined.
The significance of these AP values validates \cref{theorem:mocae} from the practical perspective. 
On the other hand, once the performance gap among the combined detectors increases, we observe that Vanilla \gls{MoE} and \gls{CC} do not perform well, which we discuss in the following section.

\subsection{Limitations of \gls{CC}}
\label{subsec:pitfalls}

\paragraph{On the importance of performance difference among the detectors}
As discussed in \cref{subsec:ablation} in the paper, the main limitation of \gls{CC} (as well as Vanilla \gls{MoE}) is that it benefits from combining similarly performing detectors.
This is presented in \cref{tab:pitfall1} in which we can see that combining low-performing detectors (e.g., RS R-CNN or YOLOv7) with EVA and Co-DETR decreases the performance of \gls{MoE} to an AP less than the best single detector.
On the other hand, as just presented in App. \ref{subsec:practicalimplications}, this is not the case for Oracle \gls{MoE}.
This suggests that while the calibration decreases the \gls{LaECE} of the single detectors significantly (\cref{tab:pitfall1}), they are still far from being well-calibrated to construct an accurate \gls{MoE}.
This is because, we keep the capacity of the calibrators low, that is Class-agnostic \gls{LR} and \gls{IR} are monotonically increasing function of the predicted confidence; thereby preserving the ranking among the detections and the AP of each detector.
On the other hand, this capacity limitation might not result in better  \glspl{MoE} than the individual components once their performance gap is high.
Furthermore, the Oracle \gls{MoE} consistently improves upon the individual detectors even the performance gap is significantly high as shown in \cref{tab:pitfall1}.
This demonstrates the importance of the calibration quality in obtaining an accurate \gls{MoE} as well as the potential to improve the object detectors using \glspl{MoE}, which is an open problem.

\paragraph{\gls{CC} requires more resource than single models}
Another limitation of \gls{CC} stems from using multiple models during inference, which is also the case for \glspl{DE}. 
Still, as each detector can process the input in parallel, the overhead introduced by \gls{CC} would be negligible when a separate GPU is allocated for each detector.
Besides, comparing \cref{fig:teaser_method}(b) with \cref{fig:teaser_method}(a), we showed that calibration balances the contribution of the detectors to \gls{MoE}.
However, PAA, as the most accurate detector. has still the lowest contribution with $\% 25.61$ in \cref{fig:teaser_method}(b).
This is because \gls{LaECE} after calibration in \cref{tab:teaser_method}(Top) is still non-zero for the detectors.
Therefore, better calibration methods could give rise to more accurate \glspl{MoE}.


\subsection{Full Version of the Tables Pruned in the Paper}
\cref{tab:app_lvis} presents the detailed results on LVIS dataset for instance segmentation; \cref{tab:app_rotbb} includes the performance of all classes for DOTA dataset; \cref{tab:ensembles_cal_app} shows the performance over different object scales; and \cref{tab:testdev_app} includes COCO minitest results for common object detectors as well as their main differences in terms of pretraining data and the backbone.
These tables are excluded from the main paper due to the space limitation.

\begin{table*}[t]
\small
\setlength{\tabcolsep}{0.15em}
\centering
\caption{Detailed results on LVIS val set. The detectors have different characteristics in terms of exploiting Repeat Factor Sampling (RFS), backbone, the number of training epochs and the loss function. While Vanilla \gls{MoE} does not yield gain, \gls{CC} enables a stronger \gls{MoE} than all single detectors.}
\label{tab:app_lvis}
\scalebox{0.95}{
\begin{tabular}{c|c|c|c||c|c|c||c|c|c||c|c|c||c|c|c} 
\toprule
\midrule
Method&RFS&Backbone&Epoch&\multicolumn{6}{c||}{\underline{Instance Segmentation Performance}}&\multicolumn{6}{c}{\underline{Object Detection Performance}} \\ 
&&&&$\mathrm{AP}$&$\mathrm{AP_{50}}$&$\mathrm{AP_{75}}$&$\mathrm{AP_{r}}$&$\mathrm{AP_{c}}$&$\mathrm{AP_{f}}$&$\mathrm{AP}$&$\mathrm{AP_{50}}$&$\mathrm{AP_{75}}$&$\mathrm{AP_{r}}$&$\mathrm{AP_{c}}$&$\mathrm{AP_{f}}$ \\ \midrule
Seesaw Mask R-CNN \cite{seesawloss}&\xmark&ResNet-50&24&$\underline{25.4}$&$\underline{39.5}$&$26.9$&$15.8$&$\underline{24.7}$&$\underline{30.4}$&$25.6$&$41.6$&$26.6$&$14.0$&$24.0$&$32.3$\\
RS Mask R-CNN \cite{RSLoss}&\cmark&ResNet-50&12&$25.1$&$38.2$&$26.8$&$\underline{16.5}$&$24.3$&$29.9$&$25.8$&$39.7$&$27.8$&$15.1$&$24.5$&$32.0$\\
Mask R-CNN \cite{MaskRCNN}&\cmark&ResNeXt-101&12&$\underline{25.4}$&$39.2$&$\underline{27.3}$&$15.7$&$\underline{24.7}$&$\underline{30.4}$&$\underline{26.6}$&$\underline{42.1}$&$\underline{28.5}$&$\underline{15.4}$&$\underline{25.2}$&$\underline{33.1}$\\\midrule 
Vanilla \gls{MoE}&N/A&N/A&N/A&$25.2$&$38.3$&$26.8$&$16.5$&$24.3$&$29.9$&$25.9$&$39.8$&$27.9$&$15.1$&$24.5$&$32.2$ \\
&&&&$\nimp{0.2}$&$\nimp{1.2}$&$\nimp{0.5}$&$\textcolor{red}{0.0}$&$\nimp{0.4}$&$\nimp{0.5}$&$\nimp{0.7}$&$\nimp{2.3}$&$\nimp{0.6}$&$\nimp{0.3}$&$\nimp{0.7}$&$\nimp{0.9}$ \\
\midrule
\gls{CC}&N/A&N/A&N/A&$\mathbf{27.7}$&$\mathbf{42.8}$&$\mathbf{29.4}$&$\mathbf{18.2}$&$\mathbf{27.3}$&$\mathbf{32.4}$&$\mathbf{29.1}$&$\mathbf{44.8}$&$\mathbf{31.4}$&$\mathbf{17.0}$&$\mathbf{27.9}$&$\mathbf{35.8}$\\
(Ours)&&&&$\imp{2.3}$&$\imp{3.3}$&$\imp{2.1}$&$\imp{1.7}$&$\imp{2.4}$&$\imp{2.0}$&$\imp{2.5}$&$\imp{1.6}$&$\imp{2.9}$&$\imp{1.6}$&$\imp{2.7}$&$\imp{2.7}$ \\
\bottomrule
\end{tabular}
}
\end{table*}

\begin{table*}[t]
    \small
    \setlength{\tabcolsep}{0.08em}
    \centering
    \caption{The performance of all classes on DOTA v1.0. $\mathrm{AP_{50}}$ is reported as the performance measure of DOTA.}
    \label{tab:app_rotbb}
    \scalebox{0.85}{
    \begin{tabular}{c||c||c|c|c|c|c|c|c|c|c|c|c|c|c|c|c} 
    \toprule
    \midrule
    \multirow{2}{*}{Detector}&\multirow{2}{*}{All}&\multicolumn{15}{c}{\underline{Dataset Classes}} \\
    & &Plane&Baseba.&Bridge&Ground-t.&Small-veh.&Large-veh.&Ship&Tennis&Basket.&Storag.&Soccer&Roundab.&Harbor&Swimm.&Helico.\\ \midrule
    RTMDet&$81.32$&$88.04$&$\underline{86.20}$&$58.50$&$82.43$&$81.21$&$84.87$&$\underline{88.70}$&$\underline{\mathbf{90.89}}$&$\underline{\mathbf{88.75}}$&$87.33$&$\underline{72.12}$&$70.85$&$\underline{81.16}$&$81.49$&$77.24$\\
    LSKN (prev. SOTA)&$\underline{81.85}$&$\underline{\mathbf{89.69}}$&$85.70$&$\underline{61.47}$&$\underline{83.23}$&$\underline{81.37}$&$\underline{\textbf{86.05}}$&$88.64$&$90.88$&$88.49$&$\underline{\mathbf{87.40}}$&$71.67$&$\underline{71.35}$&$79.19$&$\underline{81.77}$&$\underline{80.85}$\\ \midrule 
    Vanilla \gls{MoE}&$80.60$&$87.76$&$83.27$&$\mathbf{61.77}$&$78.25$&$81.26$&$85.33$&$88..34$&$89.93$&$85.54$&$86.35$&$70.98$&$65.92$&$84.28$&$\mathbf{82.45}$&$77.57$ \\
    &$\nimp{1.25}$&$\nimp{1.93}$&$\nimp{2.93}$&$\imp{0.30}$&$\nimp{4.98}$&$\nimp{0.11}$&$\nimp{0.72}$&$\nimp{0.36}$&$\nimp{0.11}$&$\nimp{0.96}$&$\nimp{1.05}$&$\nimp{1.14}$&$\nimp{5.43}$&$\imp{3.12}$&$\imp{0.68}$&$\nimp{3.28}$\\ \midrule
    \gls{CC}&$\mathbf{82.62}$&$89.09$&$\mathbf{86.47}$&$61.38$&$\mathbf{83.28}$&$\mathbf{81.43}$&$85.03$&$\mathbf{88.72}$&$90.86$&$88.31$&$87.11$&$\mathbf{75.50}$&$\mathbf{74.12}$&$\mathbf{84.49}$&$81.63$&$\mathbf{81.93}$\\
    (Ours)&$\imp{0.77}$&$\nimp{0.60}$&$\imp{0.27}$&$\nimp{0.09}$&$\imp{0.05}$&$\imp{0.06}$&$\nimp{1.02}$&$\imp{0.02}$&$\nimp{0.03}$&$\nimp{0.44}$&$\nimp{0.29}$&$\imp{3.38}$&$\imp{2.77}$&$\imp{3.33}$&$\nimp{0.14}$&$\imp{1.08}$\\ \midrule
    \bottomrule
    \end{tabular}
    }
\end{table*}

\begin{table*}[t]
    \small
    \setlength{\tabcolsep}{0.6em}
    \centering
    \caption{Effect of calibration on \gls{MoE} performance. All \gls{MoE}s use Late calibration with standard NMS. While combining uncalibrated detectors do not provide notable gain over the single detectors, calibration is essential for a strong \gls{MoE} resulting in up to $\sim 1.5$ AP gain over single detectors.}
    \label{tab:ensembles_cal_app}
    \scalebox{0.9}{
    \begin{tabular}{c|c||c|c|c||c|c|c||c|c|c} 
    \toprule
    \midrule
    \multirow{2}{*}{Model Type}&\multirow{2}{*}{Calibration}&\multicolumn{3}{c|}{\underline{Combined Detectors}}&\multicolumn{6}{c}{\underline{Object Detection Performance}} \\  
    &&RS R-CNN&ATSS&PAA&$\mathrm{AP}$&$\mathrm{AP_{50}}$&$\mathrm{AP_{75}}$&$\mathrm{AP_{S}}$&$\mathrm{AP_{M}}$&$\mathrm{AP_{L}}$ \\ \midrule
    \multirow{3}{*}{Single Models}&N/A&\cmark& & &$42.4$&$62.1$&$46.2$&$26.8$&$46.3$&$56.9$\\
    &N/A& &\cmark& &$43.1$&$61.5$&$47.1$&$27.8$&$47.5$&$54.2$\\
    &N/A& & &\cmark&$43.2$&$60.8$&$47.1$&$27.0$&$47.0$&$57.6$\\ \midrule \midrule
    \multirow{8}{*}{Mixtures of Experts}&\xmark&\cmark&\cmark& &$42.4$&$62.1$&$46.3$&$26.8$&$46.3$&$\mathbf{56.9}$\\
    &\cmark&\cmark&\cmark& &$\mathbf{44.1}$&$\mathbf{63.0}$&$\mathbf{48.4}$&$\mathbf{28.5}$&$\mathbf{48.4}$&$56.8$\\ \cline{2-11}
    &\xmark&\cmark& &\cmark&$43.4$&$62.5$&$47.1$&$27.3$&$47.3$&$58.0$ \\
    &\cmark&\cmark& &\cmark&$\mathbf{44.0}$&$\mathbf{62.7}$&$\mathbf{47.9}$&$\mathbf{28.2}$&$\mathbf{48.2}$&$\mathbf{58.1}$\\ \cline{2-11}
    &\xmark& &\cmark&\cmark&$43.3$&$60.9$&$47.2$&$27.1$&$47.2$&$\mathbf{57.6}$ \\
    &\cmark& &\cmark&\cmark&$\mathbf{44.4}$&$\mathbf{62.5}$&$\mathbf{48.5}$&$\mathbf{29.2}$&$\mathbf{48.5}$&$57.3$\\ \cline{2-11}    
    &\xmark&\cmark&\cmark&\cmark&$43.4$&$62.5$&$47.1$&$27.3$&$47.3$&$58.0$\\ 
    &\cmark&\cmark&\cmark&\cmark&$\mathbf{44.7}$&$\mathbf{63.1}$&$\mathbf{48.9}$&$\mathbf{29.2}$&$\mathbf{49.0}$&$\mathbf{58.2}$\\ 
    \midrule
    \bottomrule
    \end{tabular}
     }
\end{table*}

\begin{table*}[t]
\small
\setlength{\tabcolsep}{0.5em}
\centering
\caption{Object detection performance on COCO \textit{test-dev} and \textit{mini-test} using strong object detectors. The gains are reported compared to the best single model as underlined. \gls{CC} maintains the significant \gls{AP} boost also for this challenging setting as well.}
\label{tab:testdev_app}
\scalebox{0.9}{
\begin{tabular}{c|c|c||c|c|c||c|c|c||c|c|c||c|c|c} 
\toprule
\midrule
Method&Pretraining&Backbone&\multicolumn{6}{c||}{COCO test-dev}&\multicolumn{6}{c}{COCO minitest}\\ 
&Data&&$\mathrm{AP}$&$\mathrm{AP_{50}}$&$\mathrm{AP_{75}}$&$\mathrm{AP_{S}}$&$\mathrm{AP_{M}}$&$\mathrm{AP_{L}}$&$\mathrm{AP}$&$\mathrm{AP_{50}}$&$\mathrm{AP_{75}}$&$\mathrm{AP_{S}}$&$\mathrm{AP_{M}}$&$\mathrm{AP_{L}}$ \\ \midrule
YOLOv7 \cite{yolov7}&None&L-size conv.&$55.5$&$73.0$&$60.6$&$37.9$&$58.8$&$67.7$&$55.6$&$73.1$&$60.6$&$41.2$&$60.4$&$69.5$\\
QueryInst \cite{queryinst}&None&Swin-L&$55.7$&$\underline{75.7}$&$61.4$&$36.2$&$58.4$&$\underline{70.9}$&$55.9$&$75.4$&$61.3$&$38.5$&$\underline{60.8}$&$\underline{73.2}$\\
DyHead \cite{dyhead}&ImageNet22K&Swin-L&$\underline{56.6}$&$75.5$&$\underline{61.8}$&$\underline{39.4}$&$\underline{59.8}$&$68.7$&$\underline{56.8}$&$\underline{75.6}$&$\underline{62.2}$&$\underline{42.8}$&$60.6$&$71.0$\\ \midrule 
Vanilla \gls{MoE} &N/A&N/A&$57.6$&$76.6$&$63.2$&$40.0$&$60.9$&$70.8$&$57.7$&$76.3$&$62.9$&$42.6$&$62.7$&$72.8$\\
&&&$\imp{1.0}$&$\imp{0.9}$&$\imp{1.4}$&$\imp{0.6}$&$\imp{1.1}$&$\nimp{0.1}$&$\imp{0.9}$&$\imp{0.7}$&$\imp{0.7}$&$\nimp{0.2}$&$\imp{1.9}$&$\nimp{0.4}$ \\ \midrule 
\gls{CC} &N/A&N/A&$\mathbf{59.0}$&$\mathbf{77.2}$&$\mathbf{64.7}$&$\mathbf{41.1}$&$\mathbf{62.6}$&$\mathbf{72.4}$&$\mathbf{58.9}$&$\mathbf{76.8}$&$\mathbf{64.3}$&$\mathbf{44.7}$&$\mathbf{63.6}$&$\mathbf{74.1}$ \\
(Ours)&&&$\imp{2.4}$&$\imp{1.5}$&$\imp{2.9}$&$\imp{1.7}$&$\imp{2.8}$&$\imp{1.5}$&$\imp{2.1}$&$\imp{1.1}$&$\imp{2.1}$&$\imp{1.9}$&$\imp{2.8}$&$\imp{1.1}$ \\ \midrule
\bottomrule
\end{tabular}
}
\end{table*}

\end{document}